\documentclass[twoside,11pt]{article}
\usepackage{jair, theapa, rawfonts}

\usepackage[noend]{algpseudocode}

\usepackage{soul} %
\usepackage{amsfonts} %
\usepackage[hidelinks]{hyperref} %
\usepackage{siunitx} %
\usepackage{afterpage} %
\usepackage{graphbox} %
\usepackage[export]{adjustbox} %
\usepackage{natbib}
\usepackage{dirtytalk}
\usepackage[ruled]{algorithm}
\usepackage{amsmath}
\usepackage[mark=***]{sectionbreak}
\usepackage{tcolorbox}
\usepackage{tikz}
\usetikzlibrary{fadings,positioning}

\jairheading{}{2022}{1-69}{11/22}{?/23}
\ShortHeadings{Five Properties of Specific Curiosity}
{Ady, Shariff,  G\"{u}nther, Pilarski}
\firstpageno{1}

\begin{document}

\title{Five Properties of Specific Curiosity You Didn't Know Curious Machines Should Have}

\author{\name Nadia M. Ady \email nmady@ualberta.ca \\
       \addr  Dept. of Computing Science, University of Alberta \&\\
       Alberta Machine Intelligence Institute (Amii)\\
       Edmonton, Alberta, Canada \\ 
       \AND
       \name Roshan Shariff \email rshariff@ualberta.ca \\
       \addr  Dept. of Computing Science, University of Alberta \&\\
       Alberta Machine Intelligence Institute (Amii)\\
       Edmonton, Alberta, Canada \\ 
       \AND
       \name Johannes G\"{u}nther \email gunther@ualberta.ca\\
       \addr  Dept. of Computing Science, University of Alberta\\
       Edmonton, Alberta, Canada \&\\ Sony AI\\ 
       \AND       
       \name Patrick M. Pilarski \email pilarski@ualberta.ca \\
       \addr  Depts. of Medicine and Computing Science, University of Alberta \&\\
       Alberta Machine Intelligence Institute (Amii) \&\\
       DeepMind\\Edmonton, Alberta, Canada}

\maketitle

\begin{abstract}
Curiosity for machine agents has been a focus of lively research activity. The study of human and animal curiosity, particularly \emph{specific curiosity}, has unearthed several properties that would offer important benefits for machine learners, but that have not yet been well-explored in machine intelligence. In this work, we conduct a comprehensive, multidisciplinary survey of the field of animal and machine curiosity. As a principal contribution of this work, we use this survey as a foundation to introduce and define what we consider to be five of the most important properties of specific curiosity: 1) directedness towards inostensible referents, 2) cessation when satisfied, 3) voluntary exposure, 4) transience, and 5) coherent long-term learning. As a second main contribution of this work, we show how these properties may be implemented together in a proof-of-concept reinforcement learning agent: we demonstrate how the properties manifest in the behaviour of this agent in a simple non-episodic grid-world environment that includes curiosity-inducing locations and induced targets of curiosity. As we would hope, our example of a computational specific curiosity agent exhibits short-term directed behaviour while updating long-term preferences to adaptively seek out curiosity-inducing situations. This work, therefore, presents a landmark synthesis and translation of specific curiosity to the domain of machine learning and reinforcement learning and provides a novel view into how specific curiosity operates and in the future might be integrated into the behaviour of goal-seeking, decision-making computational agents in complex environments.
\vfil
\end{abstract}

\section{Books, Bookstores, and Machine Curiosity} \label{sec:intro}

Imagine you have a favourite corner bookstore near your home. You walk into the shop, browse the shelves for something new, take it home and read it end-to-end in less than a week---perhaps at the expense of sleep. The reading feels good; the unfolding plot makes you almost unable to put down the book. You read the last page and want more. There isn't any more book left, so you walk directly back to the bookstore for a chance at another great read. You don't buy and read the same book (that would be silly and you know how it ends); instead, you know that the bookstore can give you a new engaging reading experience. The more you read, the more you like reading---and that corner bookstore too! %
This example is rooted in the properties of human curiosity. In this paper, we focus on improving the specificity of how we think about curiosity with the goal of facilitating the implementation of key properties of human curiosity in machines. 

Humans have thought about their own curiosity for thousands of years, dating back at least to Aristotle in 350 BCE \cite[p.~76, in reference to \textit{Metaphysics}, Bk.~1, Ch.~2]{loewenstein1994psychology}. The study of human curiosity remains an active area of research with many diverse interpretations; recent psychological, neuroscientific and philosophical accounts by \citet{kidd2015psychology} and \citet[pp.~1--28]{zurn2015curiosity} review some of this diversity of thought. Over the last three decades, curiosity has started to also catch the focused attention of researchers seeking to create increasingly intelligent non-human machines. Select ideas from the study of human curiosity inspired fantastic breakthroughs in machine intelligence, from a robot dog shifting its own learning focus across progressively more difficult situations \citep{oudeyer2007intrinsic} 
to a simulated agent achieving higher-than-ever-before scores in Montezuma's Revenge, one of the so-called ``hard exploration'' games in the Atari suite (\citealp[pp.~4, 7]{bellemare2016unifying}; \citealp{dvorsky2016artificial}; \citealp[pp.~1,4]{burda2019exploration}; \citealp{cobbe2018reinforcement}). Work on machine curiosity is expected to continue to play an influential role in machine intelligence research. Advancing curiosity in the domain of machine intelligence is also expected to have substantial reciprocal benefits for research domains focused on human and animal curiosity, like those in psychology, education, philosophy, neuroscience, and behavioural economics.\footnote{Advances in machine intelligence have long supported the development of new theories of biological intelligence (%
\citealp[pp.~363--370]{newell1970remarks}; \citealp[Chs.~14--15]{sutton2018reinforcement}). We propose that the understanding of curiosity, as a facet of intelligence, can be similarly bolstered through the development of models of curiosity for machine intelligence. Given curiosity's political role ``equip[ping] us to pursue a more intellectually vibrant and equitable world'' (p.~xi--xii), scholars like \cite{zurn2020introduction} have emphasized the urgency of transdisciplinary conversation on curiosity.} 

To readers focused on curiosity in those domains: this paper is in large part addressed to you. One goal of this paper is to provide a new perspective on curiosity applicable to any learner, whether human, animal, or machine. Implementing any concept as an algorithm requires a different way of thinking. The abstractions that have been used thus far to improve our understanding of biological curiosity are different from those needed to build a curious machine. This approach and this work provide a unique perspective that may help researchers from multiple disciplines understand curiosity more deeply. This paper is meant to contribute as much to the field of curiosity studies \cite[see][p.~xii--xiii]{zurn2020introduction} as to that of machine intelligence.

The synthesis completed in this work has led us to define and explore five key properties needed to capture the full value of human curiosity for machine curiosity. While existing frameworks for curiosity in machine intelligence have offered clear successes, we suggest that learners implementing those frameworks would not exhibit the full range of curious behaviours exhibited in our bookstore example\footnote{This argument can be found in Section \ref{sec:intrinsic:limitations}, 
but we recommend understanding the five key properties in Section \ref{sec:properties} first.}%
---and we \textit{do} want to see that full range. In contrast, with our five properties, we posit that a learner would exhibit recognizably curious behaviour.  

Our five properties are all drawn from the study of \textit{specific curiosity}. Specific curiosity has been described by  \citet[p.~77]{loewenstein1994psychology} as ``the desire for a particular piece of information.'' %
In this paper, the term specific curiosity refers to one of the most intuitive uses of the unadorned term \textit{curiosity}, as it is the cognitive and emotional condition humans imply by saying, ``I am curious to know $X$.'' However, we aim not to consider specific curiosity through the lens of any particular definition in this work. Instead, we focus on detailed descriptions of its properties, allowing for future proposals of aspects of specific curiosity that this list does not include. 

To understand why we refer to \textit{specific} curiosity in particular, it helps to know some history of curiosity research. The term curiosity has been used as an umbrella term to describe a number of phenomena, generally information-seeking and knowledge-seeking behaviours in humans and other animals \citep[p.~449]{kidd2015psychology}. When a particular subset of these phenomena is studied, it often acquires a distinct name to define the scope of the study and clarify that the behaviours of interest may or may not represent a ``different'' phenomenon than other phenomena under the curiosity umbrella \citep[e.g.,][p.~180]{berlyne1954theory}. This choice allows authors to leave open the possibility that the phenomenon may be ``different'' in any of a number of ways, such as having different underlying mechanisms.  Specific curiosity is one such subset of the curiosity umbrella. 

In particular, the identifier `specific' is typically used in contrast with `diversive' \cite[p.~77]{loewenstein1994psychology}. While only later used with the word `curiosity,' the specific--diversive division derives from \cite{berlyne1960conflict}, who differentiated taking exploratory action for the purpose of learning something specific (specific exploration) versus for the purpose of relieving boredom or increasing stimulation (diversive
exploration) (\citeauthor{berlyne1960conflict}, \citeyear{berlyne1960conflict}, p.~80; \citeyear{berlyne1966curiosity}, p.~26). While \cite{berlyne1966curiosity} felt that diversive exploration seemed ``to be motivated by factors quite different from curiosity'' (p.~26), the terms ``specific curiosity'' and ``diversive curiosity'' have been used by other authors over the intervening years. We have chosen to adopt the term specific curiosity not only to emphasize that we are interested in a motivation to learn something specific,
but also to differentiate our goals from those of works on machine curiosity typical today (see Section \ref{sec:intrinsic} for an overview).

In Section \ref{sec:properties}, we describe the five key properties of specific curiosity in detail, specifically considering in Section \ref{sec:compu} their translation to the domain of reinforcement learning and related curiosity methodologies therein. In Sec. \ref{sec:casestudy}, we then offer an experimental demonstration of a computational specific curiosity agent inspired by those properties, along with detailed analysis of the resulting behaviour both with the properties intact and when each individual property is ablated in turn. Would a machine learner exhibit behaviour similar to that of the curious reader (you!) if placed in a similar setting? In our experiment, we will show how including just three of the key properties already helps a machine learner exhibit behaviour analogous to yours in the bookstore example. A machine learner might indeed return to the bookstore with the addition of a few specific and possibly easy to implement computational properties of specific curiosity. 

\section{Understanding Specific Curiosity} \label{sec:properties}

As you were reading your book, why did you have trouble putting it down? A clever author can walk the reader from question to question along the narrative. Each individual question seems to be a variation on: ``What's going to happen next?'' but each question is new and specific to the moment (How did they get out of the locked room? What is that character's motivation? Did the butler do it?) You know how to find each answer and in doing so, satisfy your curiosity---keep reading!

\setcounter{subsection}{-1}
\subsection{A Framework for Expressing Specific Curiosity}

Our first major act of synthesis in this manuscript will be to conceptually separate the moment where curiosity is induced from the moment where curiosity is satisfied. A curious learner cycles between these two types of situations. While \citet{gruber2019how} proposed a similar cycle---the Prediction, Appraisal, Curiosity, and Exploration (PACE) cycle (p.~1015)---their focus was on the development of a neuroscientific framework. Their neuroscientific focus does not emphasize the conceptual options and multiplicity of theoretical positions that we believe will best support the machine intelligence research community. In contrast, our synthesis is designed to support exploration of the range of possibilities for effective machine curiosity. %

Two key ideas are needed for understanding specific curiosity: (1) specific curiosity involves the consideration and manipulation of something the learner does not know: an inostensible concept; (2) inducing and satisfying curiosity require substantially different cognitive (and often physical) activities from a learner. Neither of these ideas are commonplace in the machine curiosity literature to date. Within this section, we set the stage by providing detail to develop the reader's intuition of these ideas, as this intuition will be needed to understand the five key properties that follow. From a computing perspective, where we aim to implement these ideas, some of the language we will use to describe our framework will be uncomfortably vague. This language include abstractions like \textit{knowledge}, \textit{concept}, and \textit{object}. However, given the research community's current understanding of minds---both biological and machine---these abstractions are still necessary.\footnote{Our view on our inability to define these abstractions mirrors \citeauthor{frege1951concept}'s defense of using the word \textit{concept} without definition (as translated by Geach; \citeyear{frege1951concept}, pp.~42-43): ``If something has been discovered that is simple, or at least must count as simple for the time being, we shall have to coin a term for it, since language will not contain an expression that exactly answers."} The challenging work of understanding mechanisms of mind is ongoing, and with progress towards solidifying these abstractions, our understanding and implementations of curiosity will improve as well.

\subsubsection{Inostensible Concepts} \label{sec:prelim:inostensible}

As you asked each question about the narrative of your book, you were able to think about \textit{something you wanted to know}. This experience follows the perspective put forward by \citet[p.~87]{loewenstein1994psychology} where specific curiosity\footnote{\citet[p.~92]{loewenstein1994psychology} actually expresses the information-gap perspective as a description of \textit{specific epistemic state curiosity}, a term which delineates his concept of interest on traditional axes of types of curiosity: specific vs. diversive, perceptual vs. epistemic, and state vs. trait. As we noted in Section \ref{sec:intro}, the specific-diversive axis stems from the difference between taking exploratory action for the purpose of learning something specific (specific exploration) versus the purpose of relieving boredom or increasing stimulation (diversive exploration) (\citealp[p.~80]{berlyne1960conflict}; \citeyear[p.~26]{berlyne1966curiosity}). See Footnote \ref{foot:state} for more description of the state-trait distinction. In this paper, we focus on \textit{specific state curiosity}, but primarily use the simplified term \textit{specific curiosity} with `state' implied throughout. Finally, the perceptual-epistemic axis is meant to provisionally differentiate motivation relieved by perception from motivation relieved ``by the acquisition of knowledge" (\citealp[p.~180]{berlyne1954theory}; \citeyear[pp.~399--400]{berlyne1957conflict}; \citeyear[p.~274]{berlyne1960conflict}). For the purposes of this paper, we need not make a distinction along this axis, as the properties effectively describe either perceptual or epistemic forms. Even \citeauthor{berlyne1960conflict}, who made the original distinction, suggested that epistemic and perceptual curiosity seem to be closely related (\citeyear[p.~280]{berlyne1960conflict}).} arises when a learner becomes focused on an information gap\footnote{
While \cite{loewenstein1994psychology} appears to have popularized the information gap as a theory of curiosity, the connection between curiosity and ``gaps in information'' goes back at least to \cite{berlyne1960conflict}, who extended \citeauthor{bartlett1958thinking}'s (\citeyear{bartlett1958thinking}) suggestion that thinking arises as a ``reaction to a gap" \citep[p.~280]{berlyne1960conflict} to suggest that such ``gaps in information" (\citealp[p.~280]{berlyne1960conflict}; cf. \citealp[pp.~22,~24]{bartlett1958thinking}) are similar to his own idea of \textit{conflict}, and not only evoke thinking, but other knowledge-seeking behaviours \cite[p.~280]{berlyne1960conflict}.
}---a gap between what they know and what they want to know. However, the term information gap is not well-specified\footnote{The prevalence of the term `information gap' in the study of curiosity, and the breadth of definitions and posited types of curiosity have led to the term occasionally being stretched beyond our area of interest in this paper. For example, \cite{pekrun2019murky} has recently extended the term to include ``a gap between current knowledge and the as yet unknown, expanded knowledge that could be gained by unspecific exploration" (p.~908) to account for \textit{diversive curiosity}. We do not include this non-specific viewpoint in our treatment of the idea, a choice which we believe is appropriate, as multiple authors have called into question whether diversive `curiosity' should be considered a form of curiosity at all (\citealp[p.~229--230]{markey2014curiosity}; \citealp[pp.~77-78]{loewenstein1994psychology}).}
and needs to be clarified before we can implement it algorithmically. 

We can partially clarify the meaning of \textit{information gap} via the term \textit{inostensible concept}, as coined by \citeauthor{inan2012philosophy} (\citeyear[p.~30]{inan2010inostensible}; \citeyear[p.~34]{inan2012philosophy}).\footnote{Beyond \cite{loewenstein1994psychology}'s idea of an information gap, Inan's idea of the inostensible concept follows earlier work that describes ideas similar to the inostensible concept. \cite{berlyne1954theory} described questions evoking ``mediating `concepts' or `meaning' responses'' (p.~182).
}
An inostensible concept can be simplified as a ``known unknown": something you know you do not know. If you are experiencing specific curiosity, you have an inostensible concept at the focus of that specific curiosity. In thinking about something you don't know, you are manipulating a \textit{concept} of that something you don't know.

To make this more concrete, let's think through an example. 
As you were reading the book you acquired at the bookstore, perhaps you stumbled upon the following description:
\begin{quote}{\em ``Except for an odd splash of some dark fluid on one of the white-papered walls, the whole place appeared neat, cheerful and ordinary.''}\end{quote} 
If you're anything like me, you might ask yourself, ``Why is there dark fluid on the wall?" An inostensible concept is implicit to this question.\footnote{Here, we use the word \textit{question} loosely, to refer to a feeling of recognizing an inostensible concept, because these can often be reasonably approximated as questions in the linguistic sense. We do not assume that curiosity requires linguistic abilities, and suggest that curiosity can arise prior to, or without, putting such a feeling into words. This view may be in contrast with \cite{inan2012philosophy}, who argues that pre-language children and animals cannot experience curiosity beyond instinctual ``novelty seeking, sensation seeking, or exploratory behavior" (p.~125). Our understanding of concepts in the minds of pre-language children is still extremely limited \citep{alessandroni2020development} and it seems hasty to assume that if concepts aren't communicated to us, they don't exist.} The inostensible concept could be approximated as ``how dark fluid ended up on the wall." If we knew the story of the dark fluid, our question would be answered: the concept would be \textit{ostensible}, rather than inostensible.

``\emph{How dark fluid ended up on the wall}" can be manipulated like any other concept in the mind. Note that you don't need to know how dark fluid ended up on the wall to be able to think about the inostensible concept ``\emph{how dark fluid ended up on the wall}."  Your inostensible concept (your known unknown) is composed of other concepts you are already familiar with: you have enough of an idea of what ``fluids" and ``walls" are and what ``dark" and ``ended up" mean to roughly conceptualize what it would mean to know ``how dark fluid ended up on the wall." This rough concept cobbled together from concepts you already know is the inostensible concept. It is in this sense that \citet{whitcomb2010curiosity} indicates that curiosity does not require you ``to conceive of its satisfier," rather, ``curiosity requires you to conceive only of everything your questions are about" (p.~671). The inostensible concept is the concept formed from everything your question is about. 

Each inostensible concept has an \textit{object} that it refers to, also called an \textit{inostensible referent}. For our example concept, the inostensible referent is the story of how dark fluid ended up on the wall. The term \textit{object} is not well-defined from a computational perspective, but we can think instead about what we are trying to achieve. While we might talk about trying to acquire this object, this story, we're really thinking of a particular objective: we want to incorporate the story of how the dark fluid ended up on the wall into our knowledge base.
This incorporation is the act of making an inostensible concept ostensible. There may be multiple approaches to make the inostensible concept ostensible: while we could read the next few pages of the book, we could also ask someone who has read this book before. It is computationally relevant that there are likely many different possible sets of observations one could make through different sensory apparatuses (e.g., eyes or ears) to make a given inostensible concept ostensible. 

The term \textit{inostensible concept} gives us additional power over the information gap perspective alone, as it gives us a foundation for satisfying our curiosity, for closing the gap. Our inostensible concept is defined by properties of the object of our curiosity (for example, the story must involve dark fluid ending up on the white-papered walls) that help us differentiate our particular object of interest from others.
This foundation allows us to use mental simulation---``the capacity to imagine what will or what could be" \citep[p.~8]{hamrick2019analogues}---to plan out sequences of actions we could take to make an inostensible concept ostensible. But before focusing on satisfying curiosity, we should talk about \textit{inducing curiosity}.

\subsubsection{Inducing and Satisfying Curiosity} \label{sec:prelim:inducingsatisfying}

Within the context of this work, we are focused on specific curiosity as \textit{temporary}\footnote{\label{foot:state}
Curiosity has been studied both as occurring temporarily, as is our focus, and as a persistent personality characteristic (\citealp[p.~76]{litman2003measuring}; \citealp[p.~908]{pekrun2019murky}). In the literature, the former is termed \textit{state curiosity} while the latter is called \textit{trait curiosity}. In the study of reinforcement learning---a field central to the computational components of this text---the term \textit{state} has a formal meaning (see Section \ref{sec:compu:rl}) to which we will want to refer. For this reason, we avoid using the term state curiosity in this work, despite recognizing it as the accepted term. Following the \cite{2014websters} definition of `state' as ``a particular mental or emotional condition" we will occasionally use \textit{condition} in places the word \textit{state} might be used in other works on curiosity.} and emphasize that specific curiosity is largely considered to be binary (it can be `on' or `off'). In particular, each time specific curiosity is induced, it is associated with exactly one inostensible concept at its focus.
When curiosity associated with a different inostensible concept arises, it is not a continuation of the same instance of specific curiosity.

For this reason, the recognition of an inostensible concept is key to an instance of specific curiosity being induced. Specific curiosity primarily arises after processing new observations, where we allow for both observations we might consider external, like your eyes falling across the phrase \say{Except for an odd splash of some dark fluid on one of the white-papered walls,} and observations we might consider internal, like a thought. We refer to a set of such observations as \textit{curiosity-inducing observations} and the situation where we make such observations as a \textit{curiosity-inducing situation}.

\hypertarget{tar:whencurious}{There are multiple theories about what kinds of situations induce curiosity.} \citeauthor{berlyne1960conflict} theorized that curiosity was induced by observations resulting in his concept of \textit{conflict}, where two or more incompatible responses to an observation are evoked and the brain lacks the information to reconcile which is more appropriate (\citeyear[p.~10]{berlyne1960conflict}; \citeyear[p.~26]{berlyne1966curiosity}). %
\cite{inan2012philosophy} contended that ``a certain kind of interest" (p.~126) is needed for awareness of an inostensible concept to result in curiosity.
\cite{chater2016underappreciated} suggested that curiosity arises when a learner either obtains new information they can't make sense of or becomes aware of a potential way of obtaining ``information that could help make sense of existing, stored, information" (p.~145).
\citet{pekrun2019murky} and \citet[pp.~812, 815]{peterson2019case} theorize that a ``\emph{sense of control} that it will be possible to close the gap" \cite[p.~909]{pekrun2019murky} is necessary to experience the condition of curiosity. However, while \cite{pekrun2019murky} considers a sense of control to be necessary, it isn't sufficient: the theory also includes ``an urge to close the gap" (p.~906) as a separate necessary component of curiosity, leaving open the question of in which situations such an urge will occur. %
This diversity of theories parallels the diversity of mechanisms suggested for machine `curiosity' that we will describe in Section \ref{sec:intrinsic}.
Despite this question being a longstanding area of study, we still don't precisely understand the situational determinants of curiosity.

Once induced, specific curiosity is thought to be able to end in two different ways: either attention is distracted (a possibility we discuss further in Section \ref{sec:transience}) or curiosity is \textit{satisfied} \citep[p.~183]{berlyne1954theory}. 
Drawing from the terminology of inostensible concepts used by \citet{inan2012philosophy}, curiosity is satisfied when the inostensible concept at the focus of that instance of curiosity is made ostensible (pp.~35--36). While it may seem obvious to some readers, we wish to draw attention to the point that the curiosity-inducing situation and the curiosity-satisfying situation \textit{must} be different.\footnote{Why is this distinction between curiosity-inducing situations and curiosity-satisfying situations so critical to us, the authors? In both the psychological literature on curiosity and the literature on intrinsic-reward-based computational curiosity, the curiosity-inducing situation is sometimes not differentiated from the curiosity-satisfying situation, limiting our understanding of how learning occurs through curiosity. We speak more to these limitations in Section \ref{sec:intrinsic:limitations}.} 

As an example where this requirement may not seem to hold, imagine you're moving through the bookstore and notice a peculiar noise from the floorboard as you transfer your weight onto it. If you experienced curiosity focused on whether your weight transfer caused the noise, you might find yourself satisfying your curiosity by repeating the same action that seemed to generate the noise the first time, transferring your weight back onto the same spot. In this case, it might seem that the curiosity-satisfying observation is the same as the curiosity-inducing observation. 
We see this kind of ``repeated trial'' action for many scientific curiosity questions \cite[pp.~105--106]{bonawitz2010just}.
For curiosity to be induced, however, the learner needs an inostensible concept. Critically, this means the learner knows there is something they do not know. If the curiosity-inducing situation provided the right information to satisfy this instance of curiosity, specific curiosity would not have been entered to begin with, because the known unknown would not be unknown after all. An observation of the peculiar noise as you transferred your weight does not tell you that your weight transfer caused the noise. Rather, it is the intervention and \textit{set} of repeated, consistent observations that when you transfer your weight onto that spot, the peculiar noise reoccurs that brings you enough confidence in your understanding for your curiosity to be satisfied.

But what does it mean for curiosity to be satisfied? Turning back to our example inostensible concept of \emph{how dark fluid ended up on the wall,} if I were to become curious about the content of this inostensible concept,\footnote{Yes, a learner can think about an inostensible concept without experiencing curiosity to resolve it \cite[pp.~42, 125--126]{inan2012philosophy}. The question of whether curiosity will occur or not leads us back to the open question of \hyperlink{tar:whencurious}{what kinds of situations induce curiosity}.} my curiosity might be satisfied when I read the phrase \say{\say{The two clergymen,} said the waiter, \say{that threw soup at the wall,}} printed on the following page of the book; my observation of this phrase upon turning the page constitutes a curiosity-satisfying situation. Assuming I considered the waiter sufficiently trustworthy, I may be satisfied that I now know that \emph{two clergymen threw soup at the wall, leaving a dark stain} is \emph{how a splash of dark fluid came to be on the wall}. My initially inostensible concept is now ostensible, and my curiosity is satisfied.\footnote{An illuminating description of the satisfaction of curiosity has been put forth by \citet[p.~135]{inan2012philosophy}, where curiosity is satisfied ``only when the curious being gains some new experience that [they believe] to be sufficient to come to know a certain object as being the object of [their] inostensible concept," and the interested reader might look to \citeauthor{inan2012philosophy}'s Chapter 9 for more detail. The curious reader, on the other hand, who simply wants to know where our example inostensible concept was lifted from can instead be directed to \textit{The Innocence of Father Brown} by G. K. \cite{chesterton1911innocence}.}

We use both the term \textit{observation} and the term \textit{situation} with the descriptors \textit{curiosity-inducing} and \textit{curiosity-satisfying} because multiple observations may be needed to enter or exit specific curiosity. For curiosity to be induced by reading the phrase \say{Except for an odd splash of some dark fluid on one of the white-papered walls,} you likely require multiple placements of gaze on the text. Similarly, you had to transfer your weight over that peculiar-sounding floorboard multiple times to be satisfied about the causal relationship. Without a sufficient set of the right observations, curiosity won't be induced or satisfied, respectively. Using the term \textit{situation} allows us to refer to the moment of the final observation while recognizing that more observations beyond the final one may have been needed.\footnote{We considered following \cite{isikman2016effects} in their use of the term \textit{curiosity-evoking events} rather than \textit{curiosity-inducing situations}. However, we felt that the connotations of the word \textit{event}, while allowing for the inclusion of multiple observations, suggests that the observations happen ``all at once"---temporally close together---while we mean for \textit{situation} to imply that a complete set of curiosity-inducing observations might occur across more time than might be considered a single event.}

The primary goal of this preliminary section was to clarify specific curiosity as a short-term condition and to differentiate the terms \textit{inostensible concept}, \textit{curiosity-inducing situation} and \textit{curiosity-satisfying situation}. While these terms are not all broadly used, in this work they are meant to help us be specific, as the concepts they refer to have sometimes been described interchangeably in the literature.
For example, \citet{dember1957analysis} used the term \textit{goal stimuli} as both curiosity-inducing (p.~92) and curiosity-satisfying (p.~91).
Similarly, \citet[p.~463]{dubey2020reconciling} use the term \textit{stimulus} as offering both the curiosity-inducing and curiosity-satisfying observations (e.g. a trivia question and its answer considered as one stimulus without differentiating which of the two the learner is seeking). 
In this preliminary section, we have contributed an argument for the importance of separating what occurs when curiosity is induced from what happens when curiosity is satisfied. We believe that future work---both the study of biological curiosity and the design of machine curiosity---can proceed with improved clarity with this separation recognized. \\

\subsection{Directedness Towards Inostensible Referents} \label{sec:directedness}

\begin{tcolorbox}%
Our first key property is \textit{directedness towards inostensible referents}. When specific curiosity is induced, the learner is motivated to take actions directed towards satisfying their curiosity.
\end{tcolorbox}

Directedness towards inostensible referents is inherent to many of the experiments used for studying curiosity. One of the most common experimental paradigms for this purpose is the \emph{trivia task}. In trivia tasks, experimenters attempt to induce curiosity using a trivia question, which can theoretically be satisfied by showing the associated answer to the question. Many trivia task experiments require participants to take specific actions to gain access to a curiosity-satisfying situation, like paying a token \cite[p.~970]{kang2009wick}, breaking a seal to open an envelope \cite[pp.~565, 567]{litman2005epistemic}, or pressing a key to indicate they would like to wait a short period to see the answer rather than skip ahead to another question immediately (\citealp[p.~268]{marvin2016curiosity}; \citealp[p.~463]{dubey2020reconciling}). When curious, participants generally took the specified actions to gain access to the inostensible referent. Even using an experimental setup that simply displayed the answer after a delay, \citet[p.~81]{baranes2015eye} showed that, when curious, participants' behaviour was directed in anticipation of receiving the answer, as they moved their gaze to where the answer would be shown.

Another common experimental paradigm for studying curiosity requires participants to take an action to ``uncover" a picture. For example, \citet[p.~390]{nicki1970reinforcing} required participants to press a key if they wanted to see an in-focus version of a just-seen blurred picture, while participants studied by \citet[as described by \citealp{loewenstein1994psychology}, p.~89]{loewenstein1992why} and \citet[p.~663]{hsee2016pandora} needed to click a computer mouse if they wanted to remove boxes occluding pictures of animals.%

While the above paradigms elicit simple actions to satisfy curiosity, experimenters have also used more complex situations requiring participants to take more extended sequences of directed action to acquire curiosity-satisfying information. 
\citet[pp.~819--820]{polman2017curiosity} observed an increase in stairwell traffic when they placed a curiosity-inducing situation (a placard with a trivia question) by an elevator along with the explanation that the answer could be found in the nearby stairwell. 

However, all of these experimental paradigms largely make use of what \citet[p.~818]{polman2017curiosity} call a ``curiosity appeal,'' where the experimenter or the context induces curiosity and offers a promise that a particular sequence of actions will lead to a curiosity-satisfying situation. However, outside of experiments, there isn't always an obvious plan to follow to satisfy one's curiosity. There have been multiple suggestions of a theoretical connection with creativity \citep[pp.~77--78]{gross2020cultivating}, at least in part because curiosity appears to often require the creation of non-obvious plans of actions to acquire appropriate curiosity-satisfying observations \citep[p.~2]{hagtvedt2019curiosity}. While the theory that curious learners can generate complex, adaptable plans of action to satisfy their curiosity remains understudied, this idea remains a strong starting point for thinking about how machine learners might demonstrate the directedness characteristic of specific curiosity.

\subsection{Cessation When Satisfied} \label{sec:ceases}

\begin{tcolorbox}
Our second key property is \textit{cessation when satisfied}. This property refers to the instance of specific curiosity ending immediately once curiosity has been satisfied, so the learner's motivation is no longer directed towards the same kind of observations that were or would have been curiosity-satisfying when the learner was still curious.
\end{tcolorbox}

Once a learner has achieved the goal of transforming an inostensible concept into an ostensible one, they do not need to seek the same curiosity-satisfying situation again. You didn't repeatedly read the page describing how the protagonist escaped their brush with death; once you knew the answer, curiosity did not drive you to experience it again. Instead, in the process of transforming that particular inostensible concept, you found yourself with a new question as to the relationship of the protagonist with their mysterious saviour, and while curiosity motivates you to investigate the same book, you're no longer interested in the preceding pages, only the following ones.

Theories of specific curiosity regularly reference the satisfaction of curiosity (\citealp[p.~92]{loewenstein1994psychology}; \citealp[p.~129]{schmitt2008epistemic}; \citealp[p.~1014]{gruber2019how}), and some authors consider the cessation of curiosity when ``the information gap is closed or the conflict is resolved" inherent to curiosity's definition \cite[p.~45]{renninger2016power}. 

The idea that specific curiosity ceases when satisfied has influenced the empirical study of curiosity. A number of studies have explored differences in behaviour or physiological changes when curiosity is satisfied. On the behavioural side, results shared by \cite[p.~1194]{wiggin2019curiosity} suggest that when curiosity is left unsatisfied, humans are more likely to make indulgent choices---choices that provide short-term pleasure but are not in the chooser's long-term interest, like ``the consumption of luxuries, hedonics, and other temptations" \cite[p.~1195]{wiggin2019curiosity}. Another study, by \citet{fastrich2018curiosity}, similarly varied whether participants were provided with curiosity-satisfying observations or not, but did not find any significant difference in participants' rating of curiosity or willingness to bid to satisfy their curiosity on a next, unrelated trivia question. On the physiological side, in an fMRI (functional magnetic resonance imaging) experiment performed by \cite{jepma2012neural}, participants were shown blurred pictures, sometimes followed by the corresponding clear picture, sometimes followed by an unrelated clear picture. In the condition with the corresponding clear picture---where curiosity induced by the blurred picture was thought to be relieved--- \citet{jepma2012neural} found both striatal and hippocampal activations were stronger than in the unrelated clear picture condition. Similarly, \cite{ligneul2018relief} performed an fMRI experiment where participants where shown trivia questions, sometimes followed by the corresponding answer, sometimes followed by an unrelated filler screen. In the condition with the corresponding trivia answer, \cite{ligneul2018relief} found that observing the answer yielded a ventral striatal response in the the brain. The striatum has been implicated in both pain relief and reward responses, while the hippocampus has been implicated in memory, which aligns well with the theory that specific curiosity is an uncomfortable experience which can be relieved and the evidence showing that curiosity improves memory. 

Despite evident interest in physiological changes when curiosity is satisfied, there has been minimal empirical work to confirm that curiosity does indeed cease when satisfied. A notable exception is in an experiment performed by \citet{wiggin2019curiosity}. In this experiment, all participants were shown a blurred picture, but while the participants in one condition were then shown the clear version of the same picture, participants in the other condition were not. %
Participants in both conditions responded to the ``10-item state curiosity scale of the State-Trait Personality Inventory (STPI) developed by Spielberger and Reheiser (2009)," and participants who had not been shown the clear picture rated higher on the scale in terms of ``the intensity of feelings and cognitions related to curiosity" (p.~1198). 

Note that \textit{cessation when satisfied} contrasts with the properties of behaviour motivated by extrinsic rewards. Extrinsic rewards, in the terminology of psychology, are material outcomes of an activity like obtaining food, water, or money \citep[p.~1801]{morris2022what}. Extrinsic rewards motivate behaviour repeatedly leading to the same target \citep[p.~1014]{gruber2019how}. For example, animals confined to a box with a lever will learn to repeatedly press the same lever if pressing it results in the mechanism providing the same food reward \cite[p.~504]{skinner1963operant}. 
This kind of directly repetitive behaviour is not exhibited towards curiosity-satisfying observations, as specific curiosity is expected to provide no further motivation towards the same target if the target satisfies the learner's curiosity \citep[p.~1014]{gruber2019how}.

One view that may appear to contradict cessation when satisfied is that proposed by \citet{fastrich2018curiosity}, who have suggested that curiosity may persist even after curiosity-satisfying observations have been provided. In their experiment, they found participants were more likely to demonstrate curiosity for the answer to a trivia question in a sequence of trivia questions if they had been curious for the answer to the preceding question in the sequence---whether or not they had been provided with the answer to that preceding question. \cite{fastrich2018curiosity} explain their findings as suggesting that curiosity \textit{persists} even after the associated answer is provided and curiosity can transfer to a temporally contiguous information gap, and call this effect the curiosity carry-over effect. However, their results do not contradict the property of cessation when satisfied, as in our terminology, while the present \textit{instance} of curiosity ceases when the associated inostensible concept becomes ostensible, this does not imply that a learner is unlikely to become immediately curious again, but for a different inostensible concept. \citeauthor{fastrich2018curiosity}'s (\citeyear{fastrich2018curiosity}) results rather suggest that human learners remain physiologically ``ready" for curiosity for a time interval once curiosity has been induced, whether or not it is satisfied.

Further, we can clarify that the property we are calling ``cessation when satisfied" is not the same as the ``knowledge satiation" described by \citet{murayama2019process}, in which a learner feels that they ``completely understand the topic" (p.~882). In our terminology, satisfaction occurs at the moment of making ostensible the inostensible concept associated with the current instance of specific curiosity---answering a single specific question---and does not imply a feeling of completely understanding an entire topic, where topics are seen as broader categorizations of related knowledge and activities (\citealp[p.~83]{krapp1994interest}; \citealp[p.~11]{renninger2016power}).

\subsection{Voluntary Exposure} \label{sec:voluntary}

\begin{quote}
        \textit{``An active striving to encounter new experiences, and to assimilate and understand them when encountered, underlies a huge variety of activities highly esteemed by society, from those of the scientist, the artist and the philosopher to those of the polar explorer and the connoisseur of wines."} \citet[p.~68]{berlyne1950novelty}.
\end{quote}
    
\begin{tcolorbox}
Our third key property is \textit{voluntary exposure}. This property refers to a preference for curiosity-\textit{inducing} situations, and that learners act on that preference to purposefully make themselves curious.%
\end{tcolorbox}
    
The experience of unresolved curiosity is inherently frustrating, as well-demonstrated by a novel that ends with a cliff-hanger but has no sequel in sight. Curious humans modify their behaviour to alleviate the feeling of unresolved curiosity.\footnote{\citet[pp.~21-22]{fitzgibbon2020seductive} provide an overview of the lengths people will go to to satisfy their curiosity, including paying for non-instrumental information (information that provides no benefit in terms of traditional extrinsic rewards, like money or food) or exposing themselves to pain or risk.} Despite the \emph{aversive quality} or discomfort associated with curiosity,\footnote{\label{foot:aversive}The idea that being in a condition of curiosity is uncomfortable has sparked some debate. \citet[pp.~50,~190--191]{silvia2006exploring} has argued that the idea of curiosity as aversive is a longstanding assumption with little supporting evidence popularized by \citeauthor{loewenstein1994psychology}'s seminal work (\citeyear{loewenstein1994psychology}). The difficulty in disentangling evidence of an aversive quality to curiosity from other possible motivating factors still stands in more recent work \cite[pp.~886--887]{murayama2019process}. Indeed, recent accounts of how emotions are constructed in biological brains and bodies suggests that the experience of curiosity may vary by culture \cite[Ch.~7]{barrett2017how}, and individual differences implicated in interpretation the experience of curiosity may account for some of the controversy. In the computational part of this work (Section \ref{sec:casestudy}), we take inspiration from the aversive quality of curiosity, but our computational analogue of aversive quality is not needed for our computational learner to demonstrate recognizably curious behaviour (Section \ref{sec:setup:ablation:positive}).} humans voluntarily expose themselves to curiosity, choosing to pick up mystery novels and puzzles \textit{because} they will pique curiosity \cite[p.~76]{loewenstein1994psychology}.  We aim to capture this tendency with our third property, \textit{voluntary exposure}.
    
We want to remind you of the separation of curiosity-inducing observations from curiosity-satisfying observations as we introduced in Section \ref{sec:prelim:inducingsatisfying}. While your new book contains examples of both curiosity-inducing observations and curiosity-satisfying observations, if they are associated with the same inostensible concept, then they must be in different places in the book. Re-reading the passage about the butler's shifty behaviour during the officers' interrogation (a plausible curiosity-inducing situation) will not tell you what the butler has done that they don't want the officers to be aware of (the inostensible concept). It is instead in reading the passage where the officers confront the butler about damning evidence of the butler's theft of thousands of dollars worth of their employer's property (a curiosity-satisfying situation) that your curiosity about their behaviour is satisfied.

Voluntary exposure is perhaps best observed via the vast amount of time and money that people across the world devote to activities associated with curiosity. Two of the most obvious activities include engaging with puzzles and mysteries, both of which are hugely popular activities. As examples, the puzzle genre raked in the second-highest total revenue across mobile game genres in the United States and Canada in 2021 \citep{npd2022number} and the mystery genre has held an enduring share of entertainment production over the years in multiple countries \cite[pp.~193--194]{knoblochwesterwick2006mystery}. While mysteries and puzzles are some of the most obvious curiosity-generating activities, narrative elements that induce curiosity are pervasive across genres of storytelling \cite[p.~12]{bermejoberros2022inducing}. Since storytelling features across media (including books, television, movies, games, and news), this single example of activities demonstrates a huge swatch of human life voluntarily engrossed in curiosity-inducing activities at any given time. %

While humans, as a group, seem to be drawn to curiosity-inducing activities, the type of curiosity-inducing activity seems to vary from individual to individual. While one person might be drawn to formulating mathematical proofs, another might prefer crosswords or language puzzles, and another might instead spend time on puzzles of shape and geometry, and yet another may select for mystery novels. All of these individuals demonstrate voluntary exposure to curiosity, yet they are selective \citep[compare][pp.~92--93]{krapp1994interest}. This selectivity is a starting point for our hypothesis that voluntary exposure might be learned over time, as an individual learns a preference for curiosity-inducing situations related to their preferred topics, domains, or puzzle styles. We will discuss this preference further in Section \ref{sec:coherent}, with the property of coherent long-term learning.

\subsection{Transience} \label{sec:transience}

\begin{tcolorbox}
Our fourth key property, \textit{transience}, refers to an instance of curiosity ending when attention is distracted or diverted.
\end{tcolorbox}

As you went to pay for your book, you became intensely curious to learn the current news of a Hollywood star's familial strife, but only while you paid attention to the magazines placed temptingly close to the checkout. Once you've torn yourself away to pay, your mind is happy to resume other functions, so once you're out the door and on your way home to start your new book, the star's struggles  are as good as forgotten (example inspired by \citealp[p.~76]{loewenstein1994psychology}).
    
When attention is distracted, the instance of curiosity ends, and this property is referred to as \textit{transience} \cite[pp.~86,~92]{loewenstein1994psychology}.\footnote{The properties \textit{cessation when satisfied} and \textit{transience} are similar in that both refer to the condition of curiosity ending, but we have separated them to better align with how the terms are used in the literature. There may also be ways the mechanisms for each property should offer different effects. For example, there are some theories that the satisfactory resolution of curiosity is actively rewarding (\citealp[p.~863]{shin2019homo}; \citealp[p.~879]{murayama2019process}).} While some authors have written about curiosity as though it can be sustained over long periods, even over years \cite[e.g.,][pp.~7-8]{engel2015hungry}, transience of curiosity is a frequently recognized property.\footnote{While the term \textit{transience} was used in \citeauthor{loewenstein1994psychology}'s (\citeyear[p.~76]{loewenstein1994psychology}) seminal paper on the information-gap theory of curiosity, the property is sometimes simply referred to as dissipation or decline of curiosity, but specifically that caused by the distraction of attention (\citealp[p.~232]{markey2014curiosity}; \citealp[p.~856]{shin2019homo}; \citealp[p.~152]{dan2020clickbaits}).} 

Like cessation when satisfied, transience appears prominently in theories of curiosity and intuitive examples. Early on, \citet{berlyne1954theory} noted that curiosity can end if distraction occurs (p.~183). More recently, the property of transience appears to have shaped \citeauthor{loewenstein1994psychology}'s (\citeyear{loewenstein1994psychology}) information gap theory: one of the reasons that \textit{attention} to an information gap is key to the formulation is that curiosity is thought to end when attention is distracted (p.~92). %

In recent experiments, \citet[Experiments 1B and 2B: ``Salience"]{golman2021demand} had participants solve puzzles (1B) or identify emotions associated with facial expressions (2B). \cite{golman2021demand} manipulated the amount of time before participants were offered the solution to one of the more challenging puzzles if they failed to solve it (1B) or the amount of time before participants were offered the chance to view their score on the facial emotion recognition test (2B). Participants who were offered the opportunity to satisfy their curiosity immediately were more likely to click multiple times or complete an unrelated task to obtain the solution/score than those who were offered the same opportunity 24 hours later. Distanced from the original context of a concerted effort to solve the puzzle or test questions, participants showed less impetus to acquire the solution or scores. While this experiment is only a partial demonstration of transience, since by offering the solution/score, \cite{golman2021demand} draw attention back to the inostensible concept, this decrease in demonstrated curiosity suggests that for many participants, curiosity has ended and this simple return of attention is insufficient to rekindle curiosity.

The condition of specific curiosity is a concerted effort to make an inostensible concept ostensible. It requires adaptive planning, which is likely resource-heavy, and, in biological learners, active movement of the body towards perceiving curiosity-satisfying observations. Transience helps the learner manage an all-or-nothing effort to satisfy their curiosity, because it means that behaviour and use of attentional resources can be fully reallocated to other matters as needed.

\subsection{Coherent Long-Term Learning} \label{sec:coherent}

\begin{tcolorbox}
    The property of \emph{coherent long-term learning} refers to how specific curiosity works in concert with other mechanisms of attention and value to orient the learner towards inostensible concepts related to the learner's prior knowledge.
\end{tcolorbox}

In this work, we have attempted to be very careful to model specific curiosity as a short-term motivational effect that begins when curiosity is induced and ends when curiosity is satisfied or when attention is diverted. However, curiosity is choosy. Moment-to-moment, humans are faced with a galaxy of unknowns, but the mechanisms of curiosity choose carefully—and it is not as though curiosity simply chooses the most readily available unknown; rather, curiosity often sends us out on a temporally extended plan to make our inostensible concept ostensible. Importantly, curiosity seems to be biased towards learning ideas related to the learner's pre-existing background knowledge \citep[p.~1377]{wade2019role}.

\citeauthor{zurn2022edgework} have recently proposed a connectional account of curiosity (\citeyear{zurn2022edgework}), explicitly critiquing the `acquisitional' metaphors commonly used for curiosity in recent decades. Curiosity is often thought to drive us to acquire information (p.~259-261). The connectional model instead emphasizes curiosity as building connections between ideas (p.~261). The connectional account aligns with \citeauthor{wade2019role}'s (\citeyear{wade2019role}) notion that a learner's level of curiosity is well-predicted by their metacognitive estimates of their own knowledge (p.~1380). If a learner recognizes metacognitively that they have existing knowledge related to a potential learning opportunity, they are well-prepared to make that connection and integrate it into their knowledge base.

By including the property of coherent long-term learning in our list of key properties, we are formally emphasizing the importance of specific curiosity's integration with the learner's current knowledge base. In humans, this integration may occur via the mechanism of individual interest. Individual interest refers to a predisposition to repeatedly engage with a class of content, where a class of content usually refers to a domain or category of knowledge, objects, or ideas. The class of content may be thought of as broad as ‘science’ or ‘playing tennis’ \citep[p.~6]{renninger2016power} or more narrow, like ‘approaches to machine curiosity that offer the benefits of human curiosity’—the best description will be highly individual and depend on the learner’s organization of their knowledge. The connectional account of curiosity can help us think of a class of content as a set of ideas that have been connected in the learner's mind, woven together by the relationships that the learner recognizes among them. Individual interest is distinguished from other motivational concepts by two components: stored knowledge and stored value, both for the particular class of content.

\paragraph{Curiosity $\rightarrow$ Individual Interest:} Curiosity may shape individual interest by increasing both knowledge and value for content areas that a learner experiences curiosity in. By driving learning, curiosity increases knowledge. \citet{rotgans2017relation} have provided initial evidence that individual interest is a consequence of learning, showing small but significant effects that growing knowledge results in increased individual interest. Indeed, the process of continually developing knowledge (availability of "cognitive challenges") in the content area of interest is required to maintain an individual interest \citep[p.~379]{renninger2000individual}. Curiosity provides impetus for a process of continually developing your knowledge.

Curiosity may also play a role in increasing value, as individual interest in a class of content reflects high levels of not only knowledge but value for the content relative to other classes of content \cite[p.~375]{renninger2000individual}. Experimentally, \citet[pp.~559, 566]{ruan2018teasing} found that, when subjects experienced the resolution of curiosity about particular well-known brands, they developed increased positive attitudes towards those brands. Such increases in positive attitudes may reflect increased value. In one experiment, \citet{ruan2018teasing} teased some participants with an animation of a gift card gradually being revealed from an envelope (so these participants needed to wait to find out where the card could be spent) and showed other participants the whole gift card immediately (so these participants immediately knew the card could be spent at Target) (p.~564). When surveyed after, the participants who had to wait for the gift card to be pulled from the envelope had a more positive average attitude toward Target (p.~565). \citet{ruan2018teasing} also found similar results with different manipulations creating and resolving uncertainty about different brands. Further research into this effect is needed, but we hypothesize that, more generally, learners may develop increased positive attitudes towards topics associated with the inostensible concept when curiosity is resolved.

\paragraph{Individual Interest $\rightarrow$ Curiosity:} Individual interest may direct curiosity by directing a learner's attention. Individual interest schools our attention onto aspects of what we perceive that we relate to our pre-existing interests. \citet{renninger2000individual} has described individual interest as acting like a filter on a learner's perception (p.~380). For example, I have an individual interest in curiosity, so when a character in my book says, ``No, I'm not curious," my attention is drawn to how curiosity fits into the situation and how my understanding of curiosity explains or fails to explain the character's lack of motivation. A learner with other individual interests would likely focus on other aspects of the same scene. In this way, individual interest provides can bias curiosity towards inostensible concepts that connect with existing knowledge. 

\paragraph{Curiosity $\longleftrightarrow$ Individual interest:} Given the early evidence we have described, we hypothesize a bidirectional relationship between curiosity and individual interest. Related bi-directional proposals have been previously raised by \citet[p.~186]{arnone2011curiosity} and \citet[p.~185]{engel2009how}. There has been a recent surge in effort to understand curiosity’s relationship with interest (e.g., \citealp{peterson2019curiosity}). More recent work has focused on the direction that experiences of curiosity may build individual interest (\citealp[pp.~863--864]{shin2019homo}; \citealp[p.~814]{peterson2019case}) and substantial work remains to develop a complete account. However, a relationship with some mechanism to re-engage learning related to prior knowledge is likely necessary to provide specific curiosity with the property of coherent long-term learning.

\paragraph{} The property of coherent long term-learning, the last of our five properties, closes the loop of how curiosity can guide a learner over a lifetime. Our list of properties began with the impetus to satisfy our curiosity in a specific, directed way (1, Directedness towards inostensible referents), an effect that ends relatively quickly, either via being satisfied (2, Cessation when satisfied) or via attention being diverted (3, Transience). Our final two properties speak to aspects of curiosity relevant to a learner's entire lifetime: learners should seek curiosity-inducing situations (4, Voluntary exposure) and curiosity should build up knowledge and value, biasing the learner's future experiences of curiosity towards learning opportunities to build on what they already know (5, Coherent long-term learning).

\sectionbreak

In this section, we described five properties of curiosity, and in particular, of specific curiosity (defined by \citet[p.~87]{loewenstein1994psychology} as ``an intrinsically motivated desire for specific information''). %
While specific curiosity is associated with other properties, particularly intensity, association with impulsivity, and a tendency to disappoint when satisfied \citep[p.~17; citing \citeauthor{loewenstein1994psychology}, \citeyear{loewenstein1994psychology}]{chater2016underappreciated}, the set of five properties we described above are expressly valuable to a learner, as we will argue in Section \ref{sec:benefits}, after we have described existing computational alternatives in Section \ref{sec:compu}. As researchers work to design curious machine agents, we believe that these properties are ones we should strive to attain. 

\section{Specific Curiosity for Machine Intelligence} \label{sec:compu}

To create a computational form of specific curiosity, we in essence want an algorithm to exhibit the properties of specific curiosity identified and defined in this manuscript. This means that a robot or computer running such an algorithm would take actions reflecting these properties. In this section, we discuss what infrastructure is needed to make such an algorithm possible, and provide the preliminaries for the framework that we argue is most appropriate---reinforcement learning.

For an algorithm to exhibit the properties of specific curiosity, the machine running the algorithm should be able to decide how to act in the world and exhibit preferences about its available choices; we see this especially in the properties of \textit{directedness} and \textit{voluntary exposure}. Above the other properties, directedness involves a preference for a sequence of actions that should satisfy curiosity, and voluntary exposure involves a preference for curiosity-inducing situations. It is valuable if the agent can learn what kinds of situations induce curiosity and which sequences of actions might lead to those particular situations and develop the appropriate preferences over learning. The capability to demonstrate learned preferences is the primary reason we consider reinforcement learning to be especially appropriate for designing algorithms that reflect machine curiosity,
as reinforcement learning centres around algorithms that use access to sensations of their environments (at least partial) to choose actions that affect the environment \citep[p.~3]{sutton2018reinforcement}. 
Within the framework, instances of reinforcement learning algorithms are often called \textit{agents}, because they have agency to shape their own experience in the world and learn from their actions. This quality makes the framework well-suited for the design of machine curiosity algorithms. 

\subsection{Reinforcement Learning} \label{sec:compu:rl}

In the remainder of the paper, we will rely on language and a choice of framework drawn from reinforcement learning. In this subsection, we introduce the framework and define some of the language that we will help us both to express the differences between the key properties proposed in this paper and existing related methods and to describe our case study in Section \ref{sec:casestudy}.

One way of representing an agent's experience of the world in a reinforcement learning framework is as an alternating sequence of observations and actions marked by time. We think of time as discrete, and at each time step, a single observation is made and a single action is taken, resulting in a sequence of the form 
\begin{equation}
    O_0, A_0, O_1, A_1, ..., O_t, A_t, O_{t+1}, A_{t+1}, ...
\end{equation}
The agent has a set of actions, $\mathcal A$, available to them, so the action taken at time $t$ is denoted $A_t \in \mathcal A$. Each observation, denoted $O_t$ for the observation at time $t$, provides (possibly partial) information about the current \textit{state} of the environment $S_t$. Informally, the state of the environment is the situation that the agent finds itself in; depending on the situation (state), the agent's choice of action will have different effects and could lead to different next situations \cite[pp.~7, 47]{sutton1998reinforcement}. If I'm standing in the open doorway of the bookstore, a step forward could lead me into the splendorous observation of mountains of books; if I'm standing in front of the closed door, a step forward might lead to me bumping my nose.

In classical reinforcement learning, each observation from $t=1$ onwards includes a numerical reward signal $R_t \in \mathbb R$. The agent must choose actions to maximize how much\footnote{Grammatically, you may have expected ``how many rewards" instead of ``how much reward," but within reinforcement learning, each reward can be a different real number, and we are concerned with maximizing \textit{return}, a function involving the sum of rewards over time. For instance, while the learner receives a reward at each timestep, one reward of $76.243$ is going to be more desirable than accumulating three rewards of $-8$, $0$, and $0.3$, so ``how many rewards" wouldn't reflect the meaning of return.} reward accumulates over time, a quantity called the \textit{return}, $G_t$. There are several possible definitions of return, $G_t$, but for simplicity in this paper, we use \textit{discounted return},\footnote{\citet[p.~55]{sutton2018reinforcement} offer further intuition about this choice.} which relies on a \textit{discount rate}, $\gamma \in [0,1)$, to place less value on rewards the further into the future they occur.

\begin{equation}
    G_t = \sum_{k=0}^\infty \gamma^k R_{t+k+1} \label{eq:return}
\end{equation}

It is common for a reinforcement learning agent to keep a running estimate of how valuable different parts of the world are so that they can map their representation of the current state $S_t$ (usually formed using the present observation $O_t$) to an estimate of its \textit{value} and use these estimates to attempt to accumulate more value. A \textit{value function}, denoted $v_\pi$, is defined as the expected \textit{return} moving forward from that state, assuming the agent follows policy $\pi$.

\begin{equation}
    v_\pi(s) := \mathbb{E}_\pi \left[ G_t \middle\vert S_t = s\right]
\end{equation}
We denote an agent's estimated value function as $V$. Estimated value functions offer an intuitive way to think about agent preferences: states with higher estimated value are preferred by the agent. In this way, we could algorithmically express the property of voluntary exposure as the agent estimating increased value for situations that are expected to induce curiosity. 

While there are multiple ways that a reinforcement learning agent might maintain an estimated value function, one of the most important approaches is called \textit{temporal-difference (TD) learning} \cite[Ch.~6]{sutton2018reinforcement}. When the agent transitions from state $S_t$ to state $S_{t+1}$, receiving reward $R_{t+1}$, we can form a new estimate for $V(S_t)$: $R_{t+1} + \gamma V(S_{t+1})$. However, since we may not always arrive in the same next state or receive the same reward when leaving state $S_t$, we usually only want to shift our estimate of $V(S_t)$ towards $R_{t+1} + \gamma V(S_{t+1})$ by a small step. We use a parameter $\alpha$, known as the \textit{step size}, to determine the amount of shift, multiplying $\alpha$ by the difference between the new estimate and the old. This difference, the \textit{TD error}, denoted $\delta$, is defined as
\begin{align}
    \delta := R_{t+1} + \gamma V(S_{t+1}) - V(S_t)
\end{align}
The simplest TD method (and the approach we take in the case study described in Section \ref{sec:casestudy}) updates the estimate of the value of state $S_t$ upon transitioning from $S_t$ to $S_{t+1}$ and receiving a reward of $R_{t+1}$ as follows:
\begin{align}
    V(S_t) \gets V(S_t) + \alpha \delta
\end{align}

The reinforcement learning framework, with these ideas of the state of the world having a specific value to a learning agent, and that an agent's choice of actions over time can influence how valuable the current state is to the agent, has been used to study not only what kind of algorithms make the best choices in such a problem setting, but also how humans and other animals choose actions, especially to consider which algorithms seem to best replicate biological decision-making. 

Within the reinforcement learning framework, there has been a long-standing assumed or hypothesized link between curiosity and \textit{exploration} \cite[p.~109]{fox2020exploration}---some researchers hope that the study of curiosity holds the solution for the exploration--exploitation dilemma. The exploration--exploitation dilemma is a long-studied challenge of reinforcement learning \cite[p.~3]{sutton2018reinforcement}. Classically, reinforcement learning problems have an optimal solution: a policy of behaviour that can obtain the maximal return. However, the learner doesn't start out knowing what the right policy is. To learn a good policy, the learner has to balance taking actions which have offered the best value in their experience so far (exploiting what they've learned) with taking actions that they haven't tried enough times to be certain of those actions' values (exploring alternative possibilities). In the following section, we will describe a number of existing methods inspired by curiosity, and many of these methods are explicitly designed to improve exploration. In this work, however, we do not assume that curiosity should contribute to exploration in this return-driven sense. Indeed, curiosity might be most interesting in the context where the learner does not have a persistent objective.

\subsection{Computational Approaches Inspired by Curiosity: Intrinsic Rewards} \label{sec:intrinsic}

The argument that reinforcement learning is an appropriate framework for computational approaches to curiosity has been embraced by many authors over the past few decades. 
The mechanisms that have been inspired by curiosity vary widely, with many using the amount of error in their machine-learning predictions (`prediction error') or ideas from information theory in the interests of simulating other constructs, like confidence \citep{schmidhuber1991curious}, learning progress \citep[p.~269]{oudeyer2007intrinsic}, surprise (\citealp[p.~14]{white2014surprise}), interest/interestingness (\citealp[p.~435]{gregor2014curiosity}; \citealp[pp.~5-6]{frank2014curiosity}), novelty (\citealp[p.~435]{gregor2014curiosity}; \citealp[pp.~1,~5]{singh2004imrl}), uncertainty \citep[pp.~1-2]{pathak2017curiosity}, compression progress (\citealp[p.~44]{graziano2011artificial}),
competence (\citealp[pp.~2417-2418]{oddi2020integrating}),
 and information gain (\citealp[p.~4]{bellemare2016unifying}; \citealp[pp.~2-3]{houthooft2016vime}; \citealp[p.~139]{still2012information}; \citealp[pp.~5-6]{frank2014curiosity}).

Most of the existing methods inspired by curiosity are centred on generating special reward-like signals, called \textit{intrinsic reward}. In this section, we provide detail on intrinsic-reward methods, including their benefits and limitations. Because intrinsic reward is the approach most commonly associated with curiosity, this section sets up the context for a discussion of computational specific curiosity as an alternative. Specific curiosity may address some of the limitations of intrinsic reward and offer a better choice for some applications of machine curiosity. Our description of specific curiosity provides a specification for computational approaches that aligns with an interdisciplinary understanding of curiosity found in the literature, in part inspired by observing a poor alignment between intrinsic-reward methods and biological curiosity. 

Intrinsic rewards can be described in relation to the term reward ($R_t$) that we described in \hyperref[sec:compu:rl]{the preceding subsection}. As you may recall, reward is given as 
part of the observations the agent makes of the environment.
The designer of the agent's learning algorithm cannot change the reward and so their algorithm must solve the optimization problem as it stands. Intrinsic rewards, on the other hand, are defined as part of the agent's learning algorithm (they are \textit{intrinsic to the agent}), but can be optimized for just as the original reward signal could be. For clarity, the original reward signal is often called \textit{extrinsic reward} to distinguish it from \textit{intrinsic reward} in the intrinsic reward literature.\footnote{Use of the term intrinsic reward in \textit{computational} reinforcement learning, as described here, differs from its use in psychology. \citet[pp.~1--4,~12]{oudeyer2009what} offer a discussion of how the terms extrinsic, intrinsic, external, and internal reward and motivation are used within the contexts of psychology versus computational systems.} 

Intrinsic reward is usually either (a) treated as a reward bonus added to the extrinsic reward provided to the environment, or (b) treated as the only reward signal, with the learner effectively ignoring any reward provided by the environment. If the intrinsic reward at time $t$ is written $R^I_t$, then standard algorithms for maximizing return can be used on the new, modified return (compare with Equation \ref{eq:return}):
\begin{align}
    \text{(a)} \quad \sum_{k=0}^\infty \gamma^k \left( R_{t+k+1} + R^I_{t+k+1} \right) & & \text{(b)} \quad \sum_{k=0}^\infty \gamma^k R^I_{t+k+1}
\end{align}

While many intrinsic reward designs have been inspired by curiosity, there is a wider body of literature about intrinsic rewards that doesn't always reference curiosity. However, many intrinsic rewards in this wider literature are designed for the same reasons that researchers often want to include curiosity in their algorithms, like improved exploration of the environment or allowing for self-directed learning. For this reason, readers interested in learning more about current machine curiosity methods may wish to explore the larger literature on computational intrinsic rewards.\footnote{ \citet{oudeyer2009what},
\citet{baldassarre2013intrinsically}, and
\citet{linke2020adapting}, 
offer overviews and surveys of intrinsically motivated computational systems.}

\subsubsection{Benefits of Intrinsic-Reward Approaches} \label{sec:intrinsic:benefits}
Intrinsic rewards have been very useful for increasing exploration on some important testbeds \citep{pathak2017curiosity,burda2019exploration,bellemare2016unifying}, and they have been used to perform well on problems where the objective outcome metric is unavailable to the agent \citep{linke2020adapting} or to generate developmental behaviour \citep{oudeyer2007intrinsic}. In intrinsic-reward approaches, the agent is rewarded for being in interesting (novel, surprising, uncertainty-reducing, etc.) states. These rewards encourage the agent to stay in or return to the same state repeatedly. Repeatedly visiting the same state has some important benefits for learning. 

\begin{enumerate}
    \item \textbf{Offers a simple way to recognize and remain on an exploration `frontier.'} Repeatedly visiting states that have not yet been visited many times can mean staying on the frontier of a part of the world that the agent has yet to explore. By \textit{frontier}, we mean states of the world from which, if the agent takes a particular action, they can end up in a state of the environment that they have never experienced before. If the agent occasionally takes a random action,\footnote{Many reinforcement learning agents occasionally take random actions \cite[p.~7]{hauser2018human}. Such agents learn about the world and develop estimates about which actions will let them accumulate the best return, and take a best action (according to that metric) most of the time. However, the agent occasionally takes one of its other possible actions, just in case its estimates were wrong. This design element to take a random action is considered a type of exploration strategy, and has good properties for ensuring the agent tries all possible actions from any state an infinite number of times \citep[p.~103]{sutton2018reinforcement}, at least if the agent has infinite time! Two popular examples are $\epsilon$-greedy exploration and soft-max/Boltzmann policy exploration.} staying on a frontier makes it more likely that it will end up visiting unexplored parts of the world via such a random action\footnote{Or, with carefully designed search control, the agent could be biased to take actions it has never taken from a given state before.} than if the agent largely stayed in the middle of the part of the world already explored. 
    
    \item \textbf{Offers a way to check if an action results in a consistent reaction.} Another perceived benefit of doing the same thing repeatedly is to check for consistency. 
    In Section \ref{sec:prelim:inducingsatisfying}, we described repeating a test of the floorboard to decide if it was the source of a peculiar noise. Similarly, for the experimental section of this paper, we completed multiple trials of each experiment because we are interested in patterns that hold over time, rather than one-time outliers. This benefit relates to an important assumption in many uses of reinforcement learning: that the world is a little bit random. When we want to estimate the value of a particular state, we are really interested in an \textit{average} so we must observe a state multiple times to form a reasonable estimate. Because of this assumption, many exploration methods try to ensure the agent visits each state of its domain multiple times.
    
    One way algorithm designers have encouraged agents to make exploratory visits to each state multiple times is through intrinsic rewards that decay over visits. This is an important area of study in exploration for reinforcement learning, and some notable approaches include Upper-Confidence bounds for Reinforcement Learning (UCRL, \citealt{ortner2007logarithmic} and UCRL2, \citealt{jaksch2010near}), Model-based Interval Estimation (MBIE, \citealt{strehl2005theoretical,strehl2008analysis}), and Random Network Distillation (RND, \citealt{burda2019exploration})---even the early curiosity system by \cite{schmidhuber1991curious} is based on a decaying bonus (assuming it is applied in a deterministic environment). The purpose of the decay is to only temporarily encourage visits to any given state, enough to obtain sufficient samples. 

\end{enumerate}

\subsubsection{Limitations of Intrinsic-Reward Approaches} \label{sec:intrinsic:limitations}

Although intrinsic-reward approaches have important benefits, they are limited in their ability to achieve those benefits and lack some of the benefits we might expect from an analogue of biological curiosity.

\paragraph{Detachment:} One limitation relates to the first benefit we described---that an intrinsic-reward approach offers a simple way to recognize and remain on an exploration `frontier,' because it doesn't always work. In particular, since most intrinsic rewards are designed to decrease as the agent returns to the same state over and over again, it is possible for the agent to essentially use up the intrinsic reward without ever taking the actions required to continue into the nearby unexplored part of the world. This is an example of the problem \citet[Supplementary Material, p.~20]{ecoffet2021first} called \textit{detachment}, where an agent leaves and fails to return to parts of its environment that are likely on a frontier---likely to be close to new parts of the environment. This problem of detachment makes intrinsic reward approaches to seeking never-before-seen states quite brittle and unable to achieve this desired benefit in some situations.

\paragraph{Reactivity:} If the goal of including a mechanism is to encourage the agent to experience something \textit{new}, intrinsic reward offers an inelegant approach, as it can only drive that goal indirectly. A reward can only be provided for observing a state once it has been observed---at which point it is no longer new. As \citet[p.~1]{shyam2019model} put it, intrinsic reward methods are \textit{reactive} and cannot direct a learner towards novel observations. The reward becomes associated with something already observed, not with novelty itself. To best achieve this goal, the agent should be directed towards the new part of the world, rather than pushed to dither near it. Of course, this is easier said than done, and so methods used to return to frontier states, like Go-Explore \citep{ecoffet2021first}, then focus on actions that may lead to novel states, instead of focusing on staying in such states, offer a useful interim measure.

\paragraph{Lack of motivation in non-stationary environments} Another limitation relates to the second benefit we described: offering a way to check for consistency. Many types of intrinsic reward---decay-based intrinsic rewards in particular---only offer a way to check for consistency if we assume the environment is stationary. By stationary, we mean that patterns and distributions in the environment never change, so once you have collected enough samples to be confident in a pattern or distribution, you never have to return to collect more. If the environment is non-stationary, the pattern could change completely while you're not looking, so you must regularly return to check if you want to be sure of your estimates.

Decay-based rewards, in particular, are generally not designed to encourage an agent to return to parts of the environment that it has already visited a sufficient number of times. However, there are intrinsic rewards designed to account for this concern: one of the earliest intrinsic rewards, used as part of \citeauthor{sutton1990bonus}'s (\citeyear{sutton1990bonus}) Dyna-Q+ agent, was an additive intrinsic reward that, for a given state, grew with the amount of time since the agent's last visit. The longer it has been since the agent's last visit to that state, the more the value for the state would grow, motivating the agent to return.

Of course, Dyna-Q+ relies on a model of the environment. In reinforcement learning, a \textit{model} of the environment, sometimes \textit{transition model}, is traditionally refers to a function that takes a state and an action to take from that state and returns a next state and reward, mimicking the environment \cite[pp.~217--218]{sutton1990bonus}. Models of the environment are notoriously challenging to formulate for real-world applications where environments are so large and complex that building full models would extend beyond real memory and computational limitations. However, the benefits of Dyna-Q+ point to a need to address these challenges to achieve effective curiosity or exploration: without being able to ``think about" or simulate experiences far from your current position in the world, it will likely often be impossible to develop specific intentions to observe parts of the world containing the information that an agent needs or wants most. 

\paragraph{Reliance on repetition of state:} Another limitation connects to the second key benefit---that intrinsic-reward approaches are useful for checking if an action results in a consistent reaction. This benefit may not align with our goals for designing computational curiosity. Curiosity may be misaligned with an underlying assumption about \emph{state} that is typical in computational reinforcement learning. We mentioned state in Section \ref{sec:compu:rl} as the situation that the agent finds themself in.  

State repetition is important for the trial-and-error aspect of reinforcement learning, as reinforcement learning is designed to evaluate how well an action went last time so the agent can adjust their behaviour next time they are in the same situation. If I didn't much like bumping my nose on the door last time, I might choose a different action when I'm next faced with a closed door.

State is usually thought of as essentially separate from the agent, and more importantly, as repeatable, meaning an agent can experience the same state multiple times. Of course, in large complex worlds, the exact same situation isn't likely to repeat multiple times, but with some generalization, this assumption is very helpful. Important features of the state can repeat multiple times and be useful for predicting reward. For example, imagine you're a rat in a box with a lever. Let's say that when a light in the box is turned on, pulling the lever results in the appearance of chocolate for you to eat, but when the light is off, nothing happens when the lever is pulled. In this case, thinking of \textit{light on} and \textit{light off} as repeatable features of state can prove very useful in optimizing your chocolate intake.

However, this assumption that state repeats should be complicated in the case of curiosity. Why? With curiosity, a learner's goal is to change their situation by making changes to their own \textit{knowledge state}. By knowledge state, we mean the state of what the learner knows---what the agent has learned from its observations of the world. One curious learner wants to change their knowledge state from not knowing who the killer is in the book they are reading to include knowing who the killer is. Another wants to get into a knowledge state where they know if it was their own action that generated peculiar noises from the floorboard. In comparison to traditional reinforcement learning state, which we can call \textit{environment state}, knowledge state is similar in that the learner can take actions to change it, but it is different in that it isn't helpful to think about returning to previous knowledge states. A learner's knowledge state is continually changing and does not have the same repeatability as environment state: it is much more useful to think of the agent's knowledge growing and adapting with each new observation of the world. Sure, an agent might forget things, but that doesn't mean it ever returns to a prior knowledge state.
    
When checking if an action results in a consistent reaction, the knowledge state of the agent actually changes after each trial. The inostensible concept of interest is not the result of a single trial, but actually some statistic about the distribution of possible results. For the agent trying to learn the value of a state, the inostensible concept might be the mean value, and for the scientist, the inostensible concept is more likely to be some underlying pattern or truth about the world. Appropriate directed behaviour, in this case, is to experience the same environmental state features multiple times, but each visit provides new information and leads to achieving a different knowledge state.

Specific curiosity still needs to make use of repeatable features of environment state. In fact, we believe that an agent learning what features of the environment tend to repeatably lead to curiosity-inducing situations might be critical to the property of voluntary exposure, (e.g. sections of bookstores labelled `Mysteries' could be a good feature). And without learning about repeating features of environment state, how could we plan directed action to satisfy our curiosity? (c.f., \citealp[p.~183]{berlyne1954theory})

\subsubsection{Specific Curiosity in Relation to the Limitations\\of Intrinsic Reward Approaches}
We believe that specific curiosity can address some of the limitations of intrinsic reward approaches. However, we also recognize that specific curiosity appears to function for a different purpose than intrinsic reward methods and compare the functions and goals of each type of method in this discussion.

\paragraph{Detachment:} By being \textit{specific}, curiosity has different goals than the methods that suffer from detachment. Specific curiosity does not attempt to cover an entire frontier and doesn't regret losing track of a state that is likely to be near novel states. Specific curiosity may be best-suited for huge environments where there is so much possible novelty that the learner needs to be choosy about which new information they seek. Intrinsic reward methods are generally not so choosy about the novelty they seek.

\paragraph{Reactivity:} Specific curiosity is less defined by reactivity and is a forward-thinking method. The core piece of specific curiosity is the planning to go retrieve a particular piece of information to create the right knowledge at the right time.

A curiosity-inducing situation seems to stem from an update to the learner's knowledge state that results in the agent recognizing an inostensible concept, or specific piece of knowledge that they don't have. In large, complex worlds where a learner can't expect to do everything it is possible to do, specific curiosity helps the agent to go get the right observations for the agent's knowledge state at the right time. 

In summary, while it may sometimes be reasonable to think of learners returning to the same environment state and action, this is not a return to the same knowledge state.

\subsubsection{Goals in Reinforcement Learning}

The word \textit{goal} lives a conflicted life within the terminology of reinforcement learning. One traditional use of the word goal is specifically in reference to maximizing return \cite[pp.~6, 53]{sutton2018reinforcement}, in reference to the \textit{reward hypothesis}, stated by \citet[p.~53]{sutton2018reinforcement} as:
\begin{quote}
    {\em ``That all of what we mean by goals and purposes can be well thought of as the maximization of the expected value of the cumulative sum of a received scalar signal (called reward).}

\end{quote}
And yet, when speaking to the intuition around reinforcement learning, there is longstanding use of the the word goal to refer to abstract accomplishments like \textit{grasp a spoon} or \textit{get to the refrigerator} (\citealp[Ch.~1.2]{sutton1998reinforcement}; \citealp[p.~5]{sutton2018reinforcement}). If we assume the reward hypothesis holds for human learners, the reward signals generated in our bodies were evolved over millions of years to shape our behaviour towards such goals, and it isn't obvious on what basis our reward signal is generated \cite[p.~469]{sutton2018reinforcement}.

The use of \textit{goal} as specifically related to maximizing return is inspired by the way \textit{goal} can be used in the context of human and animal motivation and behaviour, but defining \textit{goal} this way is limiting. More recently, taking a computational approach has led authors like \citet{grace2015specific} to define specific curiosity as ``the search for observations that explain or elaborate a particular goal concept'' (p.~262). We suggest that further consideration of what is meant by goal is needed when approaching the relationships between objectives as they relate to both environment state and knowledge state, as described above, and when attempting to broker the relationship between human and machine curiosity literature.

\subsubsection{Approaching the Five Properties in the Computational Literature} 

Computational reinforcement learning researchers have shown strong interest in aspects of the properties of directedness towards inostensible referents, cessation when satisfied, voluntary exposure, and transience. Their exploration has not always been done in the name of curiosity, however. For example,, the idea of directedness (though not necessarily towards inostensible referents) parallels work done on options (as early as \citealp{sutton1999between}) and planning. The study of \textit{options}, a mathematical abstraction of short-term policies, has resulted in a growing body of research. Part of the appeal of options is their potential to get an agent from point A to point B (which could be thought of as a goal) without emphasis on the path to get there.  Purposeful exploration using options and related ideas has been actively pursued by researchers such as \cite{machado2019efficient}.
The options framework, in particular, further reflects cessation when satisfied and aspects of transience via termination conditions for each option. Some termination conditions are naturally defined by goal states, so the directed behaviour ceases upon reaching a goal state, much like cessation when satisfied; other termination conditions can be based on when the option hasn't succeeded in reaching its goal state in a reasonable amount of time, one of the aspects of transience (\citealp[p.~212]{stolle2002learning}). However, as \cite{colas2022autotelic} have pointed out, most work with options to date has largely only considered goals within the distribution of goals previously encountered (p.~1177). One notable exception is the IMAGINE architecture, in the design of which \cite{colas2020language} leveraged the compositionality of language to generate goals---which could be seen as a step towards leveraging the compositionality of concepts to generate inostensible concepts.

Prior work has further aimed to address the lack of directedness that is a characteristic of intrinsic-reward methods. For example, the Model-Based Active eXploration algorithm presented by \citet{shyam2019model} uses planning to allow the agent ``to observe novel events" (p.~1). They care about unknowns and about creating paths to them. The Go-Explore family of algorithms also centres on the idea of taking a direct sequence of actions to move to a specific state for the purpose of exploring from it, as per \cite{ecoffet2021first}. In these examples and others, it is clear that recent work has begun to seek ways to avoid the reactive approach to designing machine curiosity. %

\citet{murayama2019process} developed a model rooted in reinforcement learning to describe the reward process involved in knowledge acquisition, designed to help explain curiosity and interest. 
\citet{yasui2020empirical} ``also found that methods which add a bonus to their value function tended to explore much more effectively than methods which add a bonus to their rewards" (p.~ii). This is part of a growing body of evidence in the literature that additive reward bonuses do not in many cases reflect or lead to the same results as human curiosity. As stated by \citet[p.~1016]{gruber2019how}, ``the effects of reward and curiosity are not additive, and reward has been shown to undermine curiosity and its effect on memory" (in reference to \citealp{murayama2010neural, murayama2011money}). Finally, active perception is a field of computing science concerned with building systems that take action to change what the system perceives towards specific goals. The needs that arise when considering how to design algorithms for specific curiosity overlap substantially with the concerns of active perception. 

In summary, it is encouraging to see a wide body of literature begin to move toward effecting what could be well considered properties of specific curiosity. In the section that follows, we expand on some of the benefits that each of the properties of specific curiosity can bring to curious reinforcement learning agents by way of a concrete implementation and empirical study that highlight how multiple properties work together as a unified whole to generate curious behaviour in a learning machine.

\section{Case Study: A Simple Learning Agent That Exhibits Properties of Specific Curiosity} \label{sec:casestudy}

We now present a case study that illustrates one possible way that three of the key properties of specific curiosity might be implemented to shape the behaviour of a reinforcement learning agent. Our intent is for this example to help the reader more deeply understand the properties of specific curiosity identified above, and how the computational principles they represent might be translated to algorithms and implementation. To support this understanding, the case study is designed to model our running bookstore example, so the agent, like you, has the opportunity to discover its own analogue of your corner bookstore.

We specifically hope to show that, even in a simple and focused setting, using the properties of specific curiosity we've highlighted as guidelines allows us to see machine behaviour emerge that approximates specific curiosity from the animal learning domain. Further, we aim to depict how these properties are modular and amenable to extension as future, more insoluble aspects of specific curiosity become computationally clear and tractable. This example is not, however, to be interpreted as a recommendation for a final or definitive computational implementation of specific curiosity. We diverge from the more common practice of fully tackling a problem without domain knowledge, instead implementing hand-designed rules of thumb or expert knowledge as solutions for some of the more challenging, unsolved aspects of computational specific curiosity, such as the process for recognizing inostensible concepts. The intended purpose of this section is for the reader to gain insight and motivation to further investigate the way the properties of specific curiosity might be integrated into different machine learning frameworks and problem settings.

To this end, we offer three sets of experiments. Sections \ref{sec:agent} and \ref{sec:domain} describe the base agent and base domain, respectively, that will be used throughout---agent interactions with the base domain are directly explored in the first set of experiments (Sections \ref{sec:setup:basic} and \ref{sec:resultsdisc:basic}). In our second set of experiments, we investigate agent behaviour when the domain is perturbed in terms of domain geometry and span (Sections \ref{sec:setup:geometry} and \ref{sec:resultsdisc:geometry}). In our third and final set of experiments (Sections \ref{sec:setup:ablation} and \ref{sec:resultsdisc:ablation}), we examine the ablation of individual properties of specific curiosity within the agent and the impact this has on agent behaviour. 

\subsection{Agent Implementation} \label{sec:agent}

In this section, we provide the specification for an agent that, if truly exhibiting the behaviour expected from the biological literature on specific curiosity, would be expected to:
    \begin{enumerate}
        \item take a largely direct route to a curiosity-satisfying situation, which we term a {\em target} (directedness),
        \item not repeatedly return to situations that had satisfied curiosity (cessation when satisfied), and
        \item develop a preference for (increased estimated value for) parts of the world that repeatedly offer curiosity-inducing observations (voluntary exposure).
    \end{enumerate}
    
The full algorithm followed by our curious agent is described in Algorithm \ref{alg:sc}. 
Sections \ref{sec:agent:directed}--\ref{sec:agent:voluntary} provide detail on how each property is included in the algorithm. The agent parameters used for our experiments are shown in Table \ref{tab:parameters}.

\begin{algorithm}[!t]
\caption{A specific example of specific curiosity}\label{alg:sc}
\begin{algorithmic}[1]
\State Initialize $\alpha$, $\epsilon$, $\gamma$, $\gamma_{curious}$, $V$, $x$ \label{algline:init}
\State Initialize $V_\textit{curious}$, $R_\textit{curious}$ to zeros \label{algline:zeroinit}
\While{$alive$} \label{algline:while}
\If{agent observation $x$ induces curiosity} \label{algline:induce}
\State generate a new curiosity target
\State generate $R_{curious} = \left\{
  \begin{array}{@{}ll@{}}
    0, & \text{if transitioning to target} \\
    {\color{blue} \bf -1}, & \text{otherwise}
  \end{array}\right.$  \Comment{{\color{blue} \textbf{A\kern-0.13em versive Quality}}} \label{algline:aversive}
\State $V_{curious} \gets {ValueIteration}(R_{curious},\gamma_{curious})$ \label{algline:valiter}
\EndIf
\If{there is currently a curiosity target (ie. the agent is curious)} \label{algline:ifcurious}
\State $x'$, $R \gets$ {\color{blue} \bf move greedily w.r.t. \boldmath{$V_{curious}(x)$}}  \Comment{{\bf \color{blue} Directed Behaviour}} \label{algline:directed}
\Else
\State $x'$, $R \gets$ move $\epsilon$-greedily w.r.t. $V(x)$ \Comment{Ties broken uniform randomly} \label{algline:vaction}
\EndIf
\State $\delta \gets R + \gamma \cdot V(x') - $ {\color{blue}\boldmath{$[ V(x) + V_{curious}(x)] $}} \Comment{ {\color{blue} \bf Voluntary Exposure} }  \label{algline:delta}%
\State $V(x) \gets V(x) + \alpha \delta$\; 
\If{agent observation $x'$ is the target} \label{algline:iftarget}
\State destroy the current target \label{algline:destroy}
\State {\color{blue} \bf reinitialize {\boldmath{$V_{curious}$}} to zeros} \Comment{{\color{blue} \bf Cessation when Satisfied}} \label{algline:ceases}
\EndIf
\State $x \gets x'$ 
\EndWhile \label{algline:endwhile}
\end{algorithmic}
\end{algorithm}

As a note on the scoping of our empirical work: In the design of the agent used in this case study, we aim to demonstrate interactions between the first three of the five key properties of specific curiosity we contributed in the sections above (directedness, cessation when satisfied, and voluntary exposure). This scope is deliberate: we place our initial focus on foundational properties of specific curiosity that for clarity of investigation can be well studied and perturbed in isolation from the experimental variability of long-term information search and the shifting focus (transience) related to life-long learning. We address these remaining two properties and their conceptual connection to our observed results in the discussion sections below, and explicitly in Section \ref{sec:benefits}.

We further contain the scope of these initial experiments by limiting the comparison of secondary computational operations that are involved in specific curiosity but that might have a variety of possible algorithms and implementations---in such cases we chose the clearest, simplest implementation of the many possible alternatives. Specifically, in Section \ref{sec:prelim:inducingsatisfying}, we noted the importance of separating curiosity-inducing observations from curiosity-satisfying situations. However, recognizing appropriate curiosity-inducing observations and estimating where in the world the appropriate satisfying observations can be found are complex issues. In this initial case study we chose to isolate the key properties from these complexities so as to better see the impact of the properties themselves on agent behaviour. We achieved this isolation by assuming the existence of an oracle-like mechanism that indicates that curiosity has been induced and indicates the location of an observation that would satisfy it. In what follows, we often refer to this particular location in the domain as the \textit{target} of curiosity, in reference to the idea that, while there may be many possible ways of making the inostensible concept of focus ostensible, the agent selects one potential curiosity-satisfying situation and then aims its behaviour towards experiencing that situation. We refer in what follows to the mechanism for recognizing curiosity-inducing situations and suggesting appropriate targets as a \textit{curiosity-recognizer module}.

\afterpage{\clearpage}

\subsubsection{Base Algorithm}

Since we are conceptualizing specific curiosity as resulting in a binary state of curiosity---at a given moment, the agent is either curious or not---we can start with a base algorithm in our experiments that determines the baseline agent behaviour when the agent is not curious. For simplicity, since the intent of this work is to explore behavioural change and not task optimality, we chose TD(0) \cite[pp.~120-121]{sutton2018reinforcement} as our base algorithm,\footnote{We herein do not rely on eligibility traces to prevent confounding their impact during analysis with the way a system might present its developed preference for curiosity-inducing situations; we expect the practical impact of accumulating or replacing eligibility traces to be one of speeding up the acquisition of preference for curiosity inducing situations, but this is a detailed comparison intended for future work.} with an $\epsilon$-greedy policy\footnote{Epsilon-greedy ($\epsilon$-greedy) behaviour refers to choosing the action that has the highest estimated value (being \textit{greedy)} nearly all of the time, but a small percentage of the time, choosing randomly from the available actions. The `epsilon,' $\epsilon$, in $\epsilon$-greedy is a parameter that sets how likely it is that a given action will be random rather than greedy. For more information on epsilon-greedy behaviour, see \citet[pp.~27-28]{sutton2018reinforcement}.} with respect to its estimated value function $V$, with ties broken by equiprobable choice. This behaviour is defined in Line \ref{algline:vaction} of Algorithm \ref{alg:sc}. Further, while the agent is not in a state of curiosity, its learning follows the standard TD(0) learning update: 
\begin{align}V(x) \gets V(x) + \alpha \delta\text{ where }\delta = R_{t+1} + \gamma V(S_{t+1}) - V(S_t)
\end{align}
Note that, in Line \ref{algline:delta} of Algorithm \ref{alg:sc}, when the agent isn't curious, our $V_\textit{curious}$ is zero everywhere, so the learning update simplifies to the standard TD(0) update.

\subsubsection{Recognizing Curiosity-Inducing Observations}
To enter a state of curiosity, the algorithm relies on a curiosity-recognizer module, which, upon a curiosity-inducing observation, generates an associated target location (Line \ref{algline:induce}). In our bookstore analogue, looking around the bookstore offers a curiosity-inducing observation, like an intriguing back-of-book blurb, and upon this observation, the reader/agent automatically has a target observation or set of target observations in mind. The target might be observing the first page of the book, and based on the target, the agent can guess how best to act to achieve the target (open the book) and proceed.

As we mentioned earlier in Section \ref{sec:agent}, recognizing when an observation should induce curiosity and estimating where an appropriate satisfier might be found are complex issues with solutions beyond the scope of this paper. For this case study, we instantiated a specific location in the domain to induce curiosity and a set of locations of possible satisfiers. Each time curiosity is induced by visiting the curiosity-inducing location, one location for a satisfier is chosen randomly from the set; we refer to this location as the target. This simple target generator acts as the curiosity-recognizer module in our experiments. The exact locations used for our experiments will be described with the domains in Sections \ref{sec:domain} and \ref{sec:setup:geometry}. We use this simplified curiosity-recognizer module to recognize when curiosity is induced.

\subsubsection{Directedness} \label{sec:agent:directed}

Once curiosity is induced, the agent changes its behaviour. To achieve the property of directedness, the agent is no longer $\epsilon$-greedy with respect to $V$ and is instead fully greedy with respect to $V_\textit{curious}$, a temporary value function. As mentioned earlier in Section \ref{sec:agent}, the key property of $V_\textit{curious}$ is that it is a gradient leading the agent towards the target provided by the curiosity recognizer: if one location is fewer actions away from curiosity's satisfier than another, the former location has higher value. An agent acting greedily with respect to the temporary value function will travel directly to curiosity's satisfier.

In our implementation, the function $V_\textit{curious}$ is generated via value iteration \citep{sutton2018reinforcement} in Line \ref{algline:valiter}.
Value iteration generates appropriate gradations in the value function, even taking into account any known obstacles or required detours between the agent's current location---or any given location---and the location of the target. See Figure \ref{fig:results-finegrain}(a, $V_\textit{curious}$) for a visualization of a gradient generated by value iteration. Value iteration is performed using the agent's transition model of the space, but uses a special reward model, $R_\textit{curious}: \mathcal S \times \mathcal A \times \mathcal S \to \mathbb R$, which maps a transition from any location other than the target to $-1$, but maps a transition from the target to $0$. Equivalently:
\begin{align} \label{eq:rcurious}
    R_\textit{curious}(s,a,s') = \left\{ \begin{array}{cl}
        0 & \text{if }s\text{ is the target}  \\
        -1 & \text{otherwise}
    \end{array}\right.
\end{align}
This choice was inspired by the characteristic aversive quality of curiosity mentioned in Section \ref{sec:voluntary}. 

Note that in this simplified agent, we provided the agent a perfect transition model of the world, so that its value iteration produces an exactly direct gradient to the target. The agent could instead learn this model from experience. Future work will need to consider the implications of \textit{not} giving the agent a perfect model, as using a perfect model is a simplification rarely possible in real-world settings.

\subsubsection{Cessation When Satisfied} \label{sec:agent:ceases}

The property of cessation when satisfied refers to the agent's behaviour no longer being affected by curiosity once the agent has observed the target of its curiosity. Once the agent has visited the target (Line \ref{algline:iftarget}), the agent is no longer curious and returns to its base behaviour. In the algorithm, this return to base behaviour is achieved by removing the target (Line \ref{algline:destroy}) and zeroing out $V_\textit{curious}$ (Line \ref{algline:ceases}). The agent will only become curious again when it has another curiosity-inducing observation as recognized by the curiosity-recognizer module. 

\subsubsection{Voluntary Exposure} \label{sec:agent:voluntary}

After several cycles of curiosity being induced, followed, and satisfied, if a particular part of the world repeatedly induces curiosity, the agent can learn a preference for returning to that part of the world. This learning process exemplifies voluntary exposure. In our running example, if you visited your corner bookstore by largely random choice during a few strolls around your neighbourhood and each time you found your curiosity sparked by excellent reads, you might find yourself heading to the bookstore directly to shortcut the process.

While we designed directedness and cessation when satisfied as simple behaviours, voluntary exposure was more interesting because we wanted our design to let the agent learn where in the world it might repeatedly become curious, and therefore voluntarily expose itself to those parts of the world---in reference to our running example, returning to the bookstore. We made a simple change to the TD update that would let the temporary value function, $V_\textit{curious}$, influence the enduring value function, $V$. This change can be found in Line \ref{algline:delta} of Algorithm \ref{alg:sc}:
\begin{align*}
\delta \gets R + \gamma \cdot V(x') - {\color{blue}\boldmath{[ V(x) + V_{curious}(x)]}}
\end{align*}
This change means that when the agent is curious, the temporary value function, $V_\textit{curious}$, affects the learning update to its estimated value function, $V$. Since $V_\textit{curious}$ is negative everywhere, the enduring value for any locations the agent visits while curious will increase.

This design choice came from intuition more than a strong theoretical underpinning. Our intuition was that the experience of curiosity should affect internal value estimates more enduringly, but the result should not be an enduring push towards curiosity's satisfiers, as we might see if visiting a satisfier were intrinsically rewarding. Instead, we hoped to see an enduring effect of increased preference for curiosity-inducing situations, or, algorithmically, increased value. Our initial experiments were designed to uncover whether this algorithmic choice would offer behaviour and value function estimates characterized by the property of voluntary exposure, which we would observe as a learned preference (increased value) for locations where the agent repeatedly makes curiosity-inducing observations. 

\sectionbreak

In summary, the agent implemented in this section was designed to incorporate three of the five key properties highlighted in this paper: directedness, cessation when satisfied, and voluntary exposure. As a reminder, this design represents only an initial example of how these properties might be simply achieved, and meant to inspire other approaches to agents exhibiting specific curiosity. In the remainder of this experimental section, we describe experiments designed to help us better understand the effects of our algorithmic choices for the agent.

\begin{table}
    \centering
    \begin{tabular}{r|r@{.}l}
        $\alpha$ & $0$&$01$ \\
        $\epsilon$ & $0$&$2$ \\
        $\gamma$ & $0$&$9$\\
        $\gamma_\textit{curious}$ & $0$&$9$\\
        initial $V: \mathcal S \to \mathbb R$ & \multicolumn{2}{c}{$V(s) = 0$ $\forall s \in \mathcal S$}
    \end{tabular}
    \caption{Parameters used at initialization in our experiments.}
    \label{tab:parameters}
\end{table}
\afterpage{\clearpage}

\subsection{Primary Domain} \label{sec:domain}

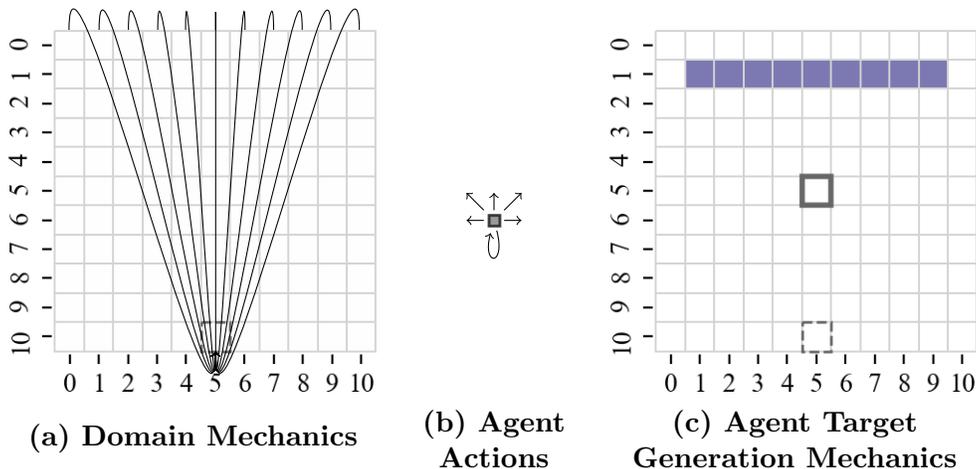
\begin{figure}[t!]
\begin{center}
\begin{tikzpicture}
\node (domain) at (0,0) {\includegraphics{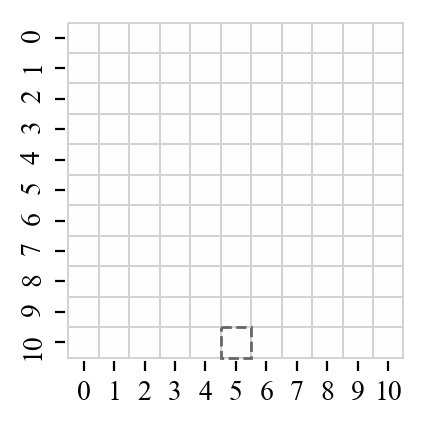}};
\node (start) at (0.3,-1.71) {};
\node (0) at (-1.65,2.3) {};
\node (1) at (-1.25,2.3) {};
\node (2) at (-0.86,2.3) {};
\node (3) at (-0.47,2.3) {};
\node (4) at (-0.1,2.3) {};
\node (5) at (0.3,2.3) {};
\node (6) at (0.69,2.3) {};
\node (7) at (1.07,2.3) {};
\node (8) at (1.45,2.3) {};
\node (9) at (1.83,2.3) {};
\node (10) at (2.21,2.3) {};
\node (dlabel) at (0, -3) {\bf (a) Domain Mechanics};
\draw [->] (0) to [out=90,in=270] (start);
\draw [->] (1) to [out=90,in=270] (start);
\draw [->] (2) to [out=90,in=270] (start);
\draw [->] (3) to [out=90,in=270] (start);
\draw [->] (4) to [out=90,in=270] (start);
\draw [->] (5) to [out=90,in=270] (start);
\draw [->] (6) to [out=90,in=270] (start);
\draw [->] (7) to [out=90,in=270] (start);
\draw [->] (8) to [out=90,in=270] (start);
\draw [->] (9) to [out=90,in=270] (start);
\draw [->] (10) to [out=90,in=270] (start);

\node (agentbg) at (4,-0.1) {\includegraphics{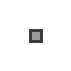}};
\node (agent) at (4,-0.1) {};
\node (left) at (3.5,-0.1) {};
\node (right) at (4.5,-0.1) {};
\node (up) at (4,0.4) {};
\node (upleft) at (3.5,0.4) {};
\node (upright) at (4.5,0.4) {};
\draw [->] (agent) to (left);
\draw [->] (agent) to (right);
\draw [->] (agent) to (up);
\draw [->] (agent) to (upleft);
\draw [->] (agent) to (upright);
\path (agent) edge [loop below] (agent);

\node (oracle) at (8,0) {\includegraphics{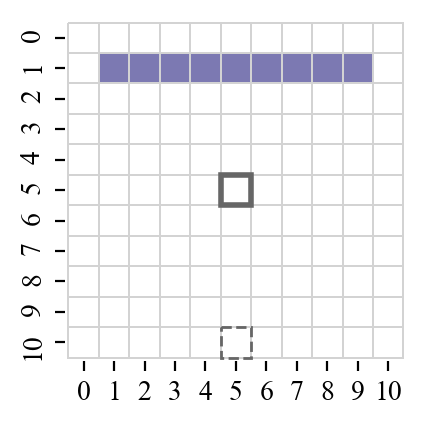}};
\node[align=center] (olabel) at (8.0, -3) {\bf (c) Agent Target \\ \bf Generation Mechanics};
\node[align=center] (alabel) at (4, -3) {\bf (b) Agent \\ \bf Actions};
\end{tikzpicture}

\end{center}
\caption{This image provides a graphical expression of the mechanics of the primary domain described in Section \ref{sec:domain}. The junction location is shown at the bottom with a grey dashed outline in (10, 5). The arrows in (a) from the top row back to the start location represent teleportation back to the junction location when the agent takes an upward action off the top of the grid. The grey rectangle shown in (b) will represent the agent in later figures, and (b) also visually shows the six actions available to the agent from any location. For clarity, the target generation mechanics needed for the curiosity-recognizing module (not considered inherent to the domain) are shown separately in (c). The curiosity-generating location has a thick solid grey outline. The possible locations for curiosity targets to be generated, across the second row from the top, are highlighted in purple.}
\label{fig:domain}
\end{figure}

With the goal that our experiments should use a simple and focused setting to highlight machine behaviour approximating biological specific curiosity, we designed a primary domain mirroring our running example of the corner bookstore. The domain is a simple gridworld, meaning that the agent occupies a single square in a grid. Our primary domain is an 11 by 11 grid, depicted in Figure \ref{fig:domain}. In this paper, our references to locations on the grid are 0-indexed using (\textit{row}, \textit{column}) notation. 

Actions can move the agent one space per step either up, on either upward diagonal, left, or right---but never down or on a downward diagonal. The agent also has a stay-here action that allows it to stay on the same location. These actions are shown visually in Figure \ref{fig:domain}(b). Directly left or right actions that would take an agent beyond the left or right boundary of the grid instead return the agent to the square where it attempted the action. Similarly, diagonal actions that would take an agent beyond the left or right boundary of the grid instead simply move the agent up. Any action that would take the agent beyond the upper boundary of the grid teleports the agent to the midpoint of the lowest row of the grid, $(10,5)$, which we will refer to as the \textit{junction location}.\footnote{The choices to have no downward actions and to teleport off the top of the grid to the midpoint of the bottom of the grid may seem unexpected. This choice was made to allow greater clarity in the visual presentation of the outcomes of the case study. By removing backtracking, the visit counts for each state more clearly show where the agent chooses to move. There are other choices, such as standard cardinal direction actions. The choice to teleport to the \textit{junction location} rather than treating the grid like a simple cylinder simplifies the learning problem by making it more likely that the agent will return to a state it has already learned about, speeding up the learning process. We evaluated a range of alternatives without some of these constraints on the domain and movement, but they have been omitted from this manuscript for what brevity we can hope to preserve; key observations in settings with backward motion, cardinal motion, cylindrical wrapping, and others are well captured by the presented results.}

A location in the centre of the grid at position $(5,5)$ is considered a permanent \textit{curiosity-inducing location}, analogous to the bookstore in our example. This choice makes the curiosity-recognizer module mentioned in Section \ref{sec:agent} very simple. When the agent enters the curiosity-inducing location, the module generates a target. Much like your corner bookstore, the curiosity-inducing location reliably induces curiosity in the agent when visited. Every target is generated in row 1 of the grid (the second row from the top) with equal probability of being placed in any of the 9 grid columns besides those directly neighbouring the left or right boundary, i.e., a target chosen from the locations in the row between and including $(1,1)$ and $(1,9)$. Different stories have different endings and different narratives to take the reader to them, so the targets generated at your corner bookstore vary.

While a reward function is usually included as part of an experimental domain for reinforcement learning, we did not include a reward function as part of the domain for the experiments described in this paper.\footnote{As a reminder from Section \ref{sec:compu:rl}, typically the goal pursued by reinforcement learning agents is to maximize their accumulation of extrinsic reward---the reward provided by the environment, usually defined as part of the domain. This goal is not directly relevant to the core of this paper, which is more focused on isolated mechanics of curiosity.} For a standard reinforcement learning agent that expects a reward for its learning algorithm, we could equivalently define $R_t$ to be $0$ at every time $t$. We leave the exploration of how best to balance the scale of $V_\textit{curious}$ with the value generated by a nonzero reward function to future work.

In the experiments showcased in this paper, we initialized each trial with the agent located at the curiosity-inducing location, as the random behaviour to find the curiosity-inducing location is not especially relevant to the mechanics central to this paper. We did run experiments with the agent starting at other locations: the results are not meaningfully affected, but the learning time is extended.

The primary domain described in this section acted as the environment that the agent interacts with in our first and third sets of experiments and as a starting point for domain modifications in the second set of experiments. By using a clear analogue of the bookstore throughout, our intention was to make behaviours characterizing specific curiosity obvious.

\subsection{First Set of Experiments: Base Domain and Agent}

\subsubsection{Experimental Setup: Visit Count and Value Study in the Primary Domain with the Base Agent} \label{sec:setup:basic}

As we noted early in Section \ref{sec:casestudy}, we wanted to design experiments to seek out machine behaviour approximating biological specific curiosity. To achieve this, we observed visit counts (where an agent goes) and the agent's estimated value function (what locations an agent learns to prefer). 

To measure what the agent does or how the agent acts, we use \textit{visit counts}. We use the term visit counts to refer to an array of integers, one integer for each location on the grid equal to the number of times the agent has visited that location. At any given timestep, the values in the visit count array will be identical to the values in the preceding timestep, except at the location that the agent visits, which will be larger by 1. At the end of a trial, the visit counts help us see where the agent spent more time and where it spent less time. Graphical examples of visit counts can be found in the right column of Figure \ref{fig:results-aggregate}. 

To gain insight into what the agent learns and how its persistent value function changes over time, we can represent the persistent value function as an array with the value equal to the estimated value of that location. Graphical examples of the persistent value function can be found in the left column of Figure \ref{fig:results-aggregate}. The agent's curiosity value function can be represented similarly, and, in the context of Algorithm \ref{alg:sc}, a record of the curiosity value function at each timestep can provide insight as to \textit{why} the agent acted in a particular way or learned a particular change in the persistent value function. Graphical examples of the curiosity value function can be found in the second row of Figure \ref{fig:results-finegrain}.

As our initial experiment, we recorded the estimated value function $V$, the curiosity value function $V_\textit{curious}$, and the visit counts of the agent described by Algorithm \ref{alg:sc} in the primary domain in 30 trials of 5000 timesteps each (with each timestep referring to an iteration over the loop in Lines \ref{algline:while}-\ref{algline:endwhile} of Algorithm \ref{alg:sc}). For each trial, we recorded the value functions at each timestep. Recording these values allowed us to create frame-by-frame animations for each trial showing the agent's movement through the grid over time along with the changing value functions. An example of agent motion and value learning in video format is provided as supplementary material: \url{https://youtu.be/TDUpB7OefFc}. 

To account for stochasticity in the agent's behaviour, we also aggregated the final estimated value functions and visit counts (after 5000 timesteps) over all 30 trials. Similarly, we aggregated the estimated value of the curiosity-inducing location and potential target locations at each timestep over all 30 trials. Observing the changes in the estimated value function, in particular, allowed us to test our hypothesis of voluntary exposure: that the curiosity-inducing location would strongly accumulate value, while the locations of the targets would accumulate relatively little value. 

Overall, this initial experiment in the primary domain allowed us to look for patterns in the agent's behaviour and learning and then compare those patterns to the expectations we developed through conceptual analysis of the properties of curiosity in Section \ref{sec:properties}.

\subsubsection{Results and Discussion: Visit Count and Value Study in the Primary Domain with the Base Agent} \label{sec:resultsdisc:basic}

\begin{figure}
\begin{center}
\makebox[\textwidth][c]{
\includegraphics[valign=t]{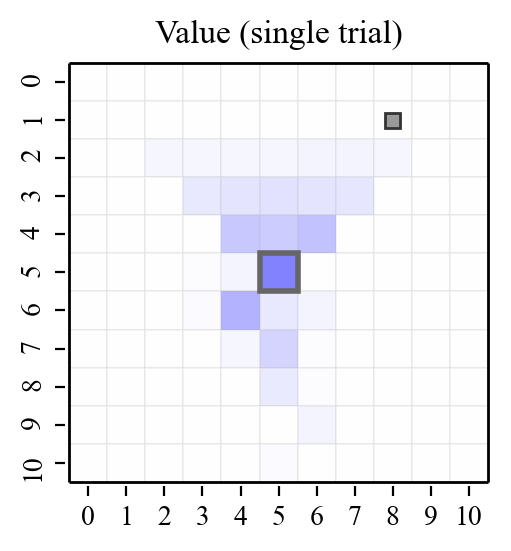}\includegraphics[trim=0 0 0 -11,clip,valign=t]{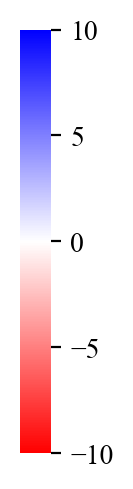}\includegraphics[valign=t]{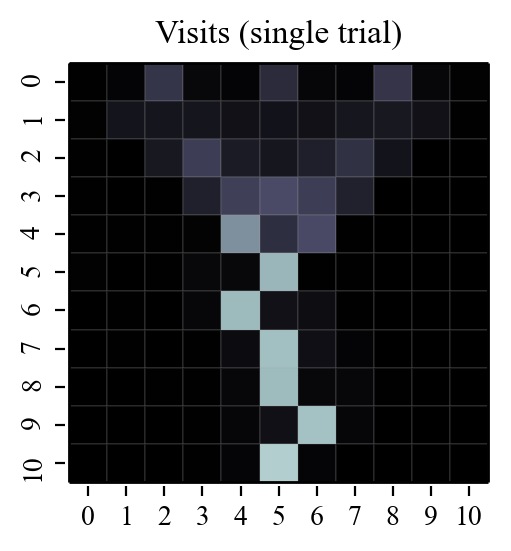}\includegraphics[trim=0 0 0 -11,clip,valign=t]{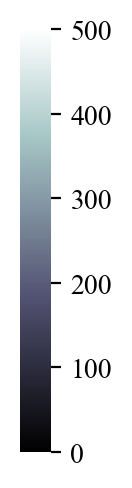}
}
{\bf (a) \hspace{12em} (b)}\\
\makebox[\textwidth][c]{
\includegraphics[valign=t]{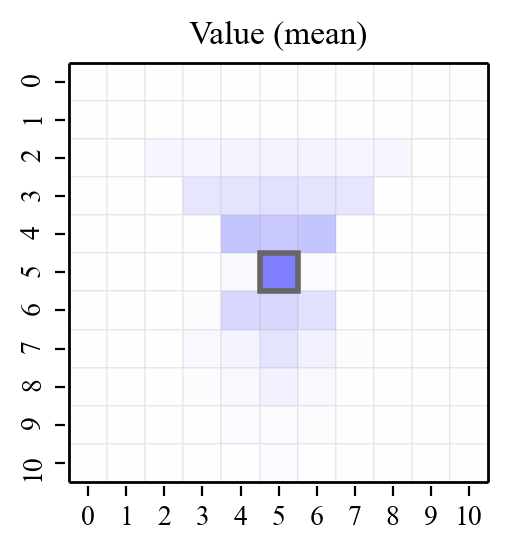}\includegraphics[trim=0 0 0 -11,clip,valign=t]{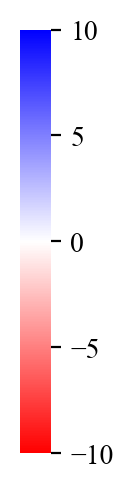}\includegraphics[valign=t]{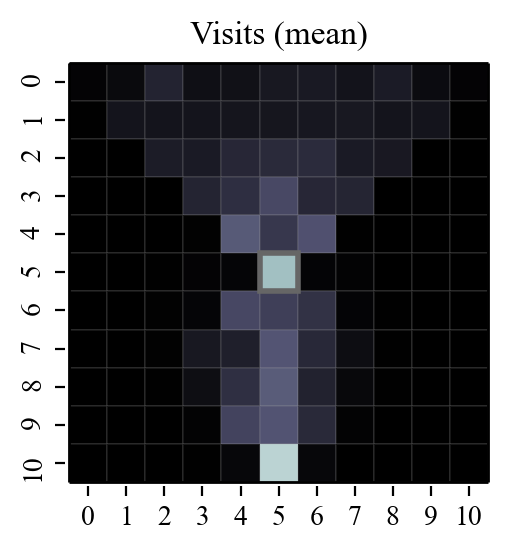}\includegraphics[trim=0 0 0 -11,clip,valign=t]{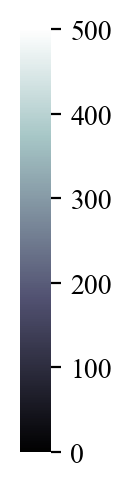}
}
{\bf (c) \hspace{12em} (d)}\\
\end{center}
\caption{This figure shows the learned value function and visit counts in the primary domain for a simple reinforcement learning agent that exhibits properties of specific curiosity. From this figure, we can see that the agent learned to value the curiosity-inducing location and therefore follow a direct path to that location, but it does not learn to value the targets of its curiosity. Shown here are (a,c) the learned value function $V$ and (b,d) the total visits the agent made to each state in the 11 x 11 grid domain. Totals plotted for trials of 5000 steps, with (a) and (b) showing value and visit counts for one representative trial, while (c) and (d) are averaged over 30 independent trials.}\label{fig:results-aggregate}
\end{figure}

One question that motivated these experiments was: Does the agent learn to value the curiosity-inducing location, emulating the property of \textit{voluntary exposure}? In particular, an agent demonstrating specific curiosity would learn a preference to return to the curiosity-inducing situation (think the bookstore) and \textit{not} learn a preference to return to the curiosity-satisfying targets (think the specific pages of each book). Figure \ref{fig:results-aggregate}(c) shows that the final value function, aggregated over all trials, had this property, with the curiosity-inducing location having the highest persistent value of all locations in the grid. We can also see a gradient leading from the bottom row of the grid up to the curiosity-inducing location, showing that, after 5000 steps, the agent had a persistent preference to move to the curiosity-inducing location. Figure \ref{fig:results-aggregate}(d) shows that this preference was reflected in the agent's behaviour: visits were concentrated between the junction location (where the agent starts each upward traversal of the grid) and at the curiosity-inducing location.

\begin{figure}
    \centering
    \includegraphics[trim=0 10 0 10, clip]{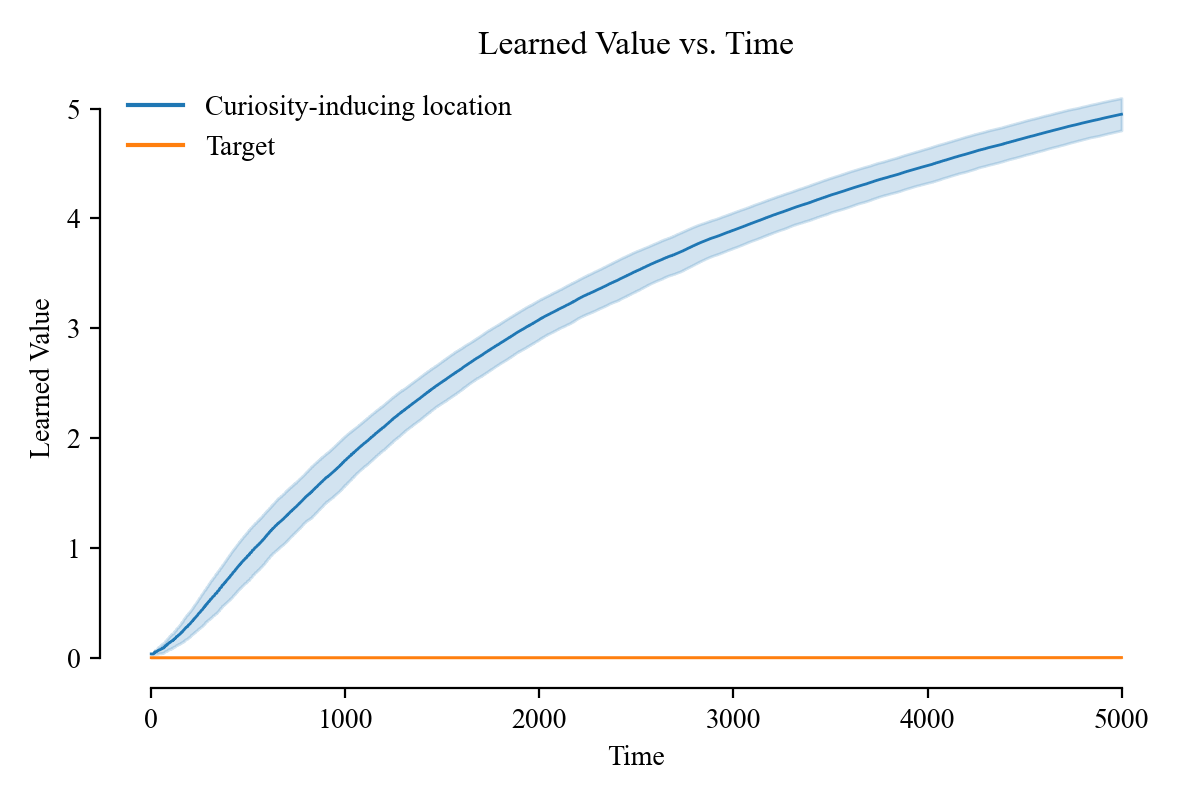}
    \caption{This figure shows the mean (line) and standard deviation (shaded area) of the estimated value of the curiosity-inducing location (in blue) and of all the possible target locations (in orange) over time, considering 30 trials. The estimated value of the curiosity-inducing location grows sublinearly while the learned values of the targets hover around $0$ throughout with little variation or growth.}
    \label{fig:results-lineplot}
\end{figure}

While it is promising to see this indication of voluntary exposure at the end of learning, we also would hope to see the difference in preference between the curiosity-inducing location and potentially curiosity-satisfying targets learned smoothly over time. Indeed, this desired pattern can be seen in the learning curves in Figure \ref{fig:results-lineplot}. 

\begin{figure}
    \makebox[\textwidth][c]{
    \begin{tikzpicture}
    \node[above right] (Rcurious) at (0,10.8) {\includegraphics[valign=t]{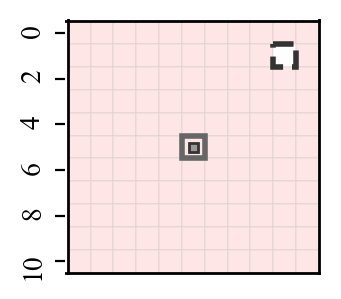}
    \includegraphics[valign=t]{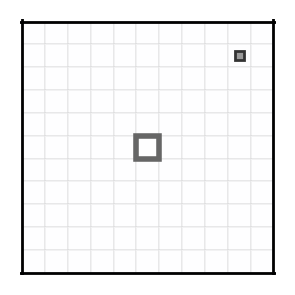}
    \includegraphics[valign=t]{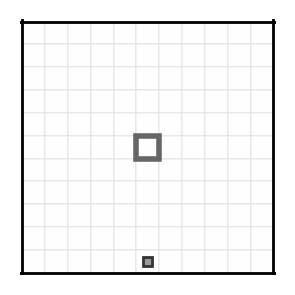}
    \includegraphics[valign=t]{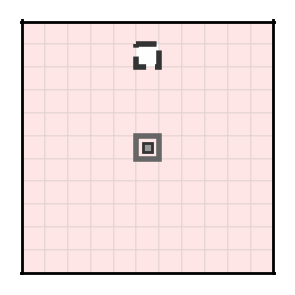}
    \includegraphics[trim=0 0 0 4,clip,valign=t]{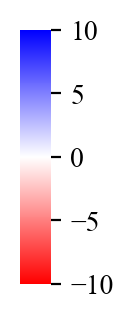}};
    \node[above right] (Rcuriouslabel) at (1,14) {$R_\textit{curious}$};
    \node[above right] (alabel) at (1,11.4) {(a)};
    \node[above right] (elabel) at (4.85,11.4) {(e)};
    \node[above right] (ilabel) at (8.75,11.4) {(i)};
    \node[above right] (mlabel) at (12.6,11.4) {(m)};
    
    \node[above right] (Vcurious) at (0,7.3) {\includegraphics[valign=t]{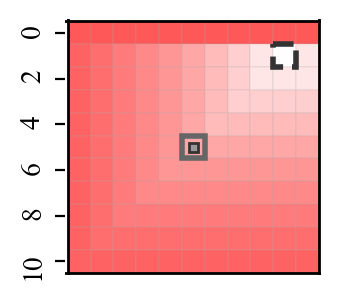}
    \includegraphics[valign=t]{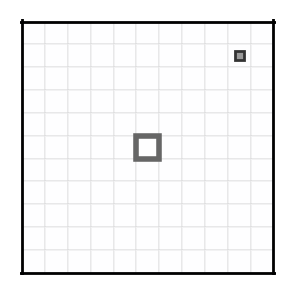}
    \includegraphics[valign=t]{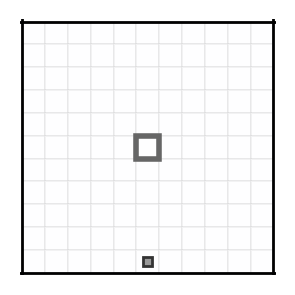}
    \includegraphics[valign=t]{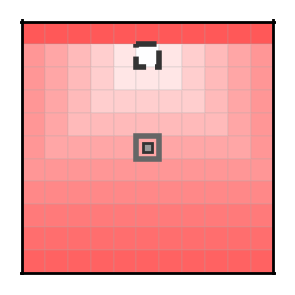}
    \includegraphics[trim=0 0 0 4,clip,valign=t]{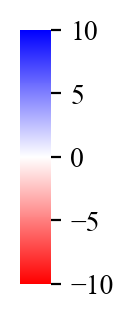}};
    \node[above right] (Vcuriouslabel) at (1,10.5) {$V_\textit{curious}$};
    \node[above right] (blabel) at (1,7.9) {(b)};
    \node[above right] (flabel) at (4.85,7.9) {(f)};
    \node[above right] (jlabel) at (8.75,7.9) {(j)};
    \node[above right] (nlabel) at (12.6,7.9) {(n)};

    \node[above right] (V) at (0,4) {\includegraphics[valign=t]{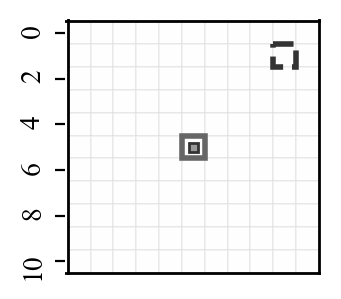}
    \includegraphics[valign=t]{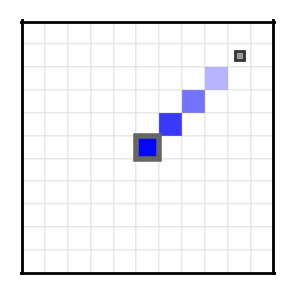}
    \includegraphics[valign=t]{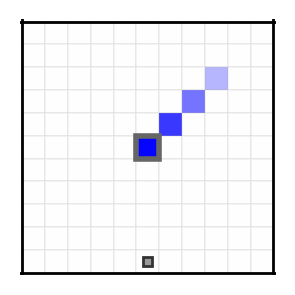}
    \includegraphics[valign=t]{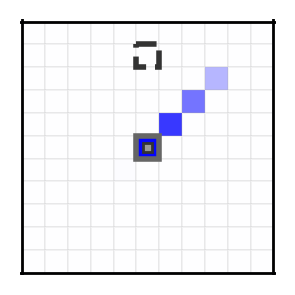}
    \includegraphics[trim=0 0 0 0,clip,valign=t]{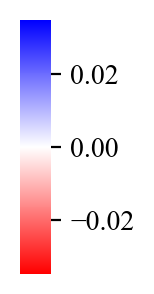}};
    \node[above right] (Vlabel) at (1,7.1) {$V$};
    \node[above right] (clabel) at (1,4.5) {(c)};
    \node[above right] (glabel) at (4.85,4.5) {(g)};
    \node[above right] (klabel) at (8.75,4.5) {(k)};
    \node[above right] (olabel) at (12.6,4.5) {(o)};
    
    \node[above right] (VisitCounts) at (0,0) {\includegraphics[valign=t]{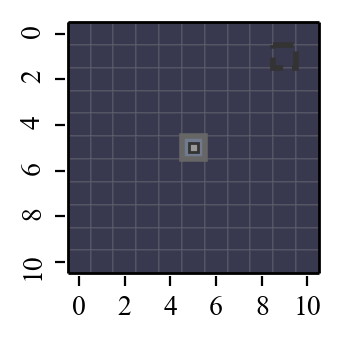}
    \includegraphics[valign=t]{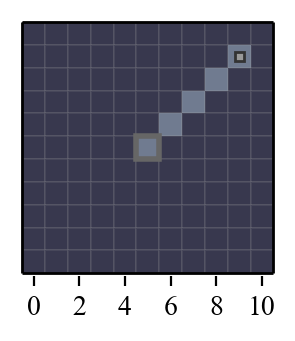}
    \includegraphics[valign=t]{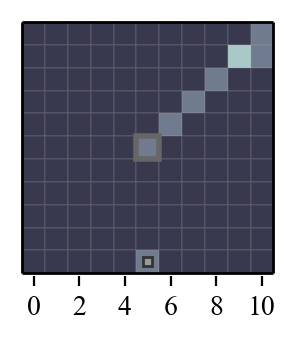}
    \includegraphics[valign=t]{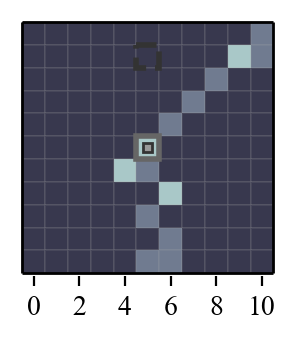}
    \includegraphics[trim=0 0 0 -0.5,clip,valign=t]{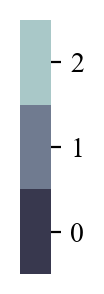}};
    \node[above right, text=white] (Visitlabel) at (1,3.6) {Visit Counts};
    \node[above right, text=white] (dlabel) at (1,1) {(d)};
    \node[above right, text=white] (hlabel) at (4.85,1) {(h)};
    \node[above right, text=white] (llabel) at (8.75,1) {(l)};
    \node[above right, text=white] (plabel) at (12.6,1) {(p)};
    
    \draw[ultra thick, ->] (2.6,0) -- (\textwidth,0);
    \draw (2.6,3pt) -- (2.6,-3pt);
    \draw (2.6,0) node[below=3pt] {$t=0$};
    \draw (6.5,3pt) -- (6.5,-3pt);
    \draw (6.5,0) node[below=3pt] {$t=3$};
    \draw (10.35,3pt) -- (10.35,-3pt);
    \draw (10.35,0) node[below=3pt] {$t=7$};
    \draw (14.2,3pt) -- (14.2,-3pt);
    \draw (14.2,0) node[below=3pt] {$t=16$};
    
    \end{tikzpicture}
    }
    \caption{\label{fig:results-finegrain}This figure is meant to offer intuition into the agent's learning behaviour by showing the agent's internal learned value function $V$, curiosity value function $V_\textit{curious}$, the curiosity reward function $R_\textit{curious}$ used to generate $V_\textit{curious}$, and the visit counts at the initialization of the trial ($t = 0$), the first visit to an induced target ($t=3$), after it has crossed off the top of the grid back to the bottom centre ($t=7$) and the second visit to the curiosity-inducing location ($t=16$).  Note the difference in scale between $V$ and $V_\textit{curious}$. While it is not visually obvious, location (6, 4) has a value $V$ of approximately $0.0003$ at $t=16$---the first step in learning a path to the curiosity-inducing location.}
\end{figure}

To understand how the agent learned to travel directly to the curiosity-inducing location, it can be helpful to follow the agent through a cycle of curiosity being induced, followed, and satisfied. The first such cycle in one trial is followed in Figure \ref{fig:results-finegrain}. The agent started at the curiosity-inducing location at $t=0$, where curiosity is triggered. The leftmost column of Figure \ref{fig:results-finegrain} shows the temporary reward function ($R_\textit{curious}$), the temporary value function ($V_\textit{curious}$), the persistent value function ($V$), and the visit counts at time $t=0$. For the agent, the induction of curiosity meant generating a curiosity-satisfying target (in the figure, the target has a dashed line border and is located near the top right of the grid). An associated temporary reward function, $R_\textit{curious}$, was generated, shown in panel (a), which was used to compute an appropriate temporary value function, $V_\textit{curious}$, shown in panel (b). 

Acting according to the property of \textit{directedness}, the agent moved directly to the target and reached that target at $t=3$, as shown in panel (h). At each step, the agent's persistent value function was updated according to Line \ref{algline:delta}, so we see the gradient we saw in $V_\textit{curious}$, panel (b), reflected in the learned value in panel (g). The further from the target, which is where $V_\textit{curious}$ is more negative, the more positive value was accumulated into the persistent value function. 

When the agent observed the target at time $t=3$, its curiosity was satisfied, and in accordance with the property of \textit{ceases when satisfied}, the target-driven behaviour ended. This means that $R_\textit{curious}$ and $V_\textit{curious}$ were zeroed out for all locations, as shown in panels (e) and (f), respectively. In this initial cycle, the agent's behaviour was wandering and largely random (as can be observed via its visits in panels (l) and (p)) until the agent reached a location adjacent to a location that has accumulated some persistent value---in this case, the agent reaches a location adjacent to the curiosity-inducing location, where a greedy action would be to move to the curiosity-inducing location. At time $t=16$, the agent visited the curiosity-inducing location where the cycle restarted with a new target.

We have seen the agent exhibit the properties of directedness, cessation when satisfied, and voluntary exposure, which was the desired result. However, this experiment was performed in a very small domain, so a next obvious question is whether these properties would still be exhibited in larger domains. Is the agent still able to learn a persistent preference for the curiosity-inducing location when the domain is larger, or when there are many possible targets? These questions motivated our second set of experiments, described in the next section.

\subsection{Second Set of Experiments: Domain Geometry}

\subsubsection{Experimental Setup: Domain Geometry Manipulations} \label{sec:setup:geometry}

While the patterns we observed through the experiments described in Sections \ref{sec:setup:basic} and \ref{sec:resultsdisc:basic} are promising reflections of specific curiosity, we were curious about whether we would observe the same patterns in a larger domain. In a larger domain, there is more space for the agent to get `lost,' and not pick up the patterns of behaviour demonstrating learned voluntary exposure and repeated cycles of curiosity. For this reason, in our second set of experiments, we manipulated the geometry, or shape, of our original $11 \times 11$ domain to make similar wide ($11\times101$) and tall ($101\times 11$) domains. In these domains, we ran four experiments:

\begin{enumerate}
    \item 30 trials of 5000 steps in wide ($11\times101$) domain
    \item 30 trials of 5000 steps in wide ($11\times101$) domain without a junction location
    \item 30 trials of 5000 steps in tall ($101\times 11$) domain with curiosity-inducing location near the bottom of the grid
    \item 30 trials of 5000 steps in tall ($101\times 11$) domain with curiosity-inducing location in the centre of the grid
\end{enumerate}

We explain these experiments in more detail in this section. In each of these domains with manipulated geometry, each key aspect of the primary domain has an analogue. The agent had the same six actions available (left, left-up diagonal, up, right-up diagonal, right, and stay-here). The targets were uniformly selected from the second row from the top of the grid: from $(1,1)$ to $(1,99)$ in the wide domain and from $(1,1)$ to $(1,9)$ in the tall domain.

In three of the four experiments, the junction location has an analogue: when the agent moves off the top of the grid, it is returned to the centre of the bottom row of the grid, which is $(10,50)$ in the wide domain and $(100,5)$ in the tall domain. In the second experiment, however, we removed the junction location, making the domain a true cylinder. When the agent moves off the top of the grid, it arrives at the bottom of the grid in the column it attempted to move into (e.g., if the agent moved on a left-up diagonal, it would arrive one column to the left of where it was along the top, unless it was against the left edge, in which case it would arrive in the bottom row in the same column).

Removing the junction location allowed us to explore how important it is for a curiosity-inducing location to be near the agent when the agent isn’t curious. If the agent fails to find a distant curiosity-inducing location, it might not demonstrate the key properties of specific curiosity. Understanding this effect has important implications for the design of an appropriate curiosity-recognizing module. For example, we may need to ensure the module has a sufficiently low threshold for the induction of curiosity to obtain useful behaviour.

We further explored this concern by manipulating the location of the curiosity-inducing location in the tall domain. It was not obvious where to put the curiosity-inducing location in the tall domain: five rows up from the bottom, or in the vertical centre of the grid? %
As the third and fourth experiments of this set, we tried both natural possibilities for the curiosity-inducing location, with the third experiment performed with the curiosity-inducing location at $(95,5)$ and the fourth with it at $(50,5)$.

By manipulating the geometry of our original domain, we hoped to find out whether the initial patterns we observed in the first set of experiments generalized to larger domains. Further, larger domains might illuminate other patterns of behaviour that might improve our choices in the design of future, more sophisticated algorithms for machine curiosity.

\subsubsection{Results and Discussion: Domain Geometry Manipulations} \label{sec:resultsdisc:geometry}

Through these experiments with larger geometry-manipulated domains, we learned three key lessons:
\begin{enumerate}
    \item \textbf{Even in expanded domains, following Algorithm \ref{alg:sc} still results in properties of directedness, cessation when satisfied, and voluntary exposure.} Of these properties, we were least certain that we would observe voluntary exposure, but by the end of every trial of these experiments, the persistent value is highest at the curiosity-inducing location, which reflects this property. For an aggregate view, see Figures \ref{fig:results-aggregate-wide}a and \ref{fig:results-aggregate-tall}a,b.
    
    \begin{figure}
    \begin{center}
    \textbf{(a) Learned Value Function $V$}\\
    \makebox[\textwidth][c]{
    \includegraphics[trim=7 0 7 20,clip,valign=top]{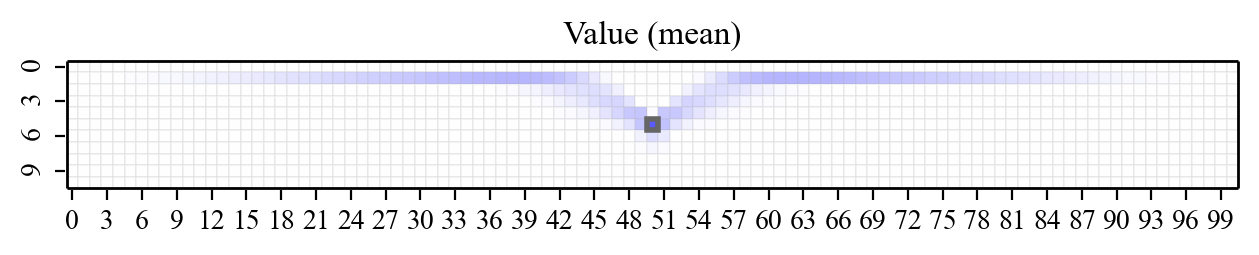} \includegraphics[trim=14 -11 6 0,clip,valign=top]{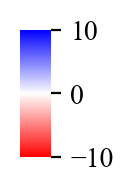}
    }
    \textbf{(b) Visit Count}\\
    \makebox[\textwidth][c]{
    \includegraphics[trim=7 0 7 20,clip,valign=top]{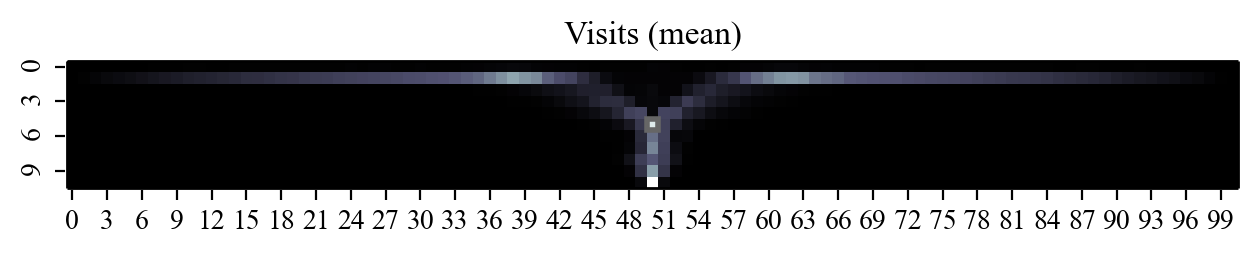}
    \includegraphics[trim=14 -11 6 0,clip,valign=top]{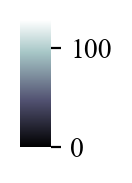}
    }
    \end{center}
    \caption{This figure shows the learned value function and visit counts in the modified wide domain for our simple reinforcement learning agent exhibiting properties of specific curiosity. This figure shows how any locations that are visited repeatedly while curious will accumulate value. Shown here are (a) the learned value function $V$ and (b) the total visits the agent made to each location. Totals plotted for trials of 5000 steps and averaged over 30 independent trials. Note that the scale of the visit counts plot differs from that in Figure \ref{fig:results-aggregate}. }\label{fig:results-aggregate-wide}
    \end{figure}
    
    \begin{figure}
    \centering
    \scalebox{0.95}{%
    \begin{tikzpicture}
        \node[above] (valuelow) at (0,0) {\includegraphics[trim=8 0 0 0,clip,valign=t]{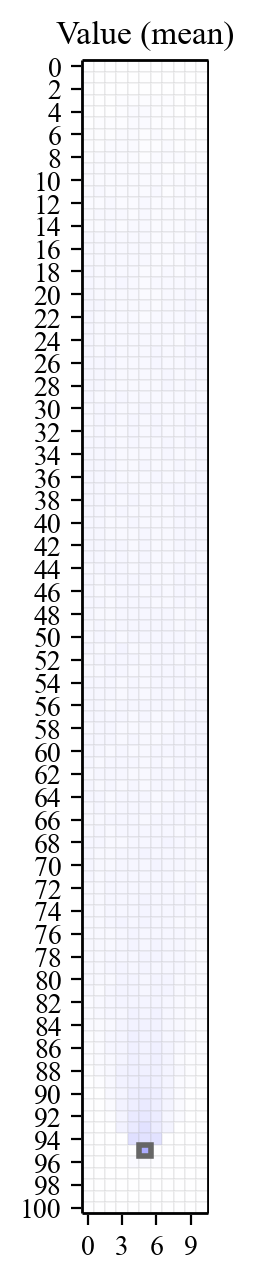}};
        \node (a) at (0,0) {\textbf{(a)}};
        \node[above] (valuemid) at (2.5,0) {\includegraphics[trim=8 0 0 0,clip,valign=t]{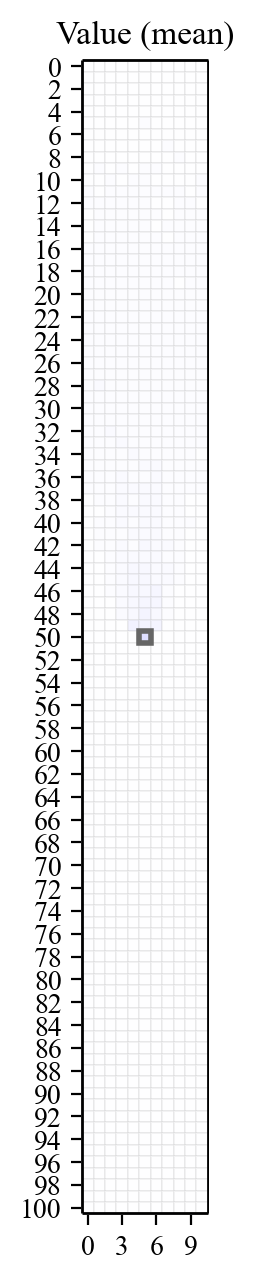}};
        \node (b) at (2.5,0) {\textbf{(b)}};
        \node[above] (valuebar) at (4.7,0.3) {\includegraphics[trim=8 0 0 -7,clip,valign=t]{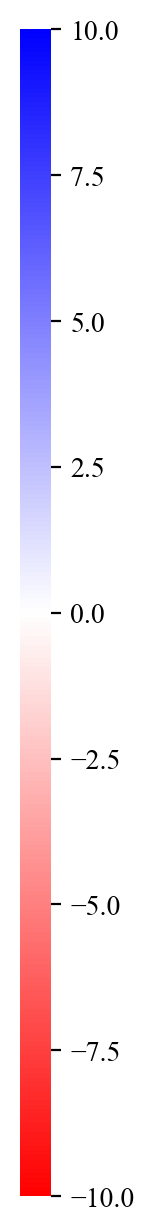}};
        \node[above] (visitslow) at (7,0) {\includegraphics[trim=8 0 0 0,clip,valign=t]{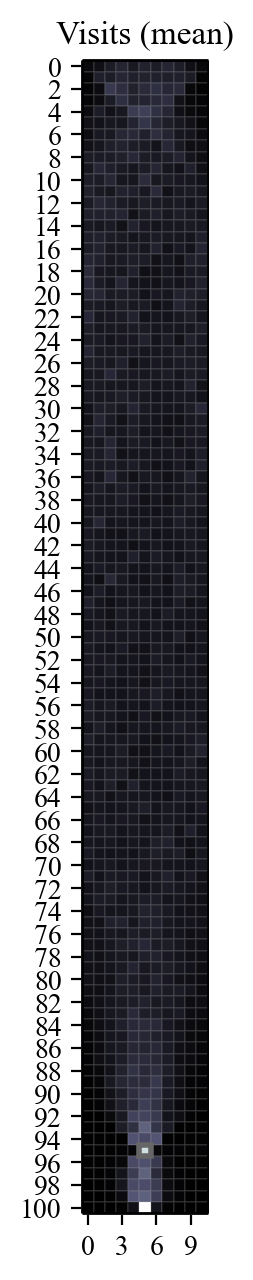}};
        \node (c) at (7,0) {\textbf{(c)}};
        \node[above] (visitslowbar) at (9,0.3) {\includegraphics[trim=8 0 0 -9,clip,valign=t]{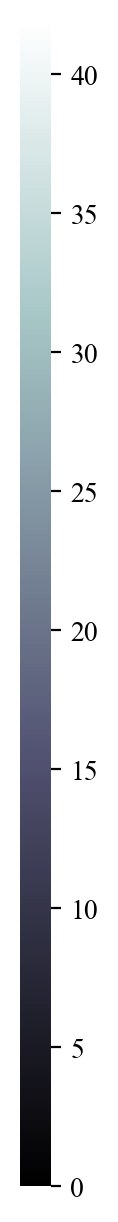}};
        \node[above] (visitsmid) at (11,0) {\includegraphics[trim=8 0 0 0,clip,valign=t]{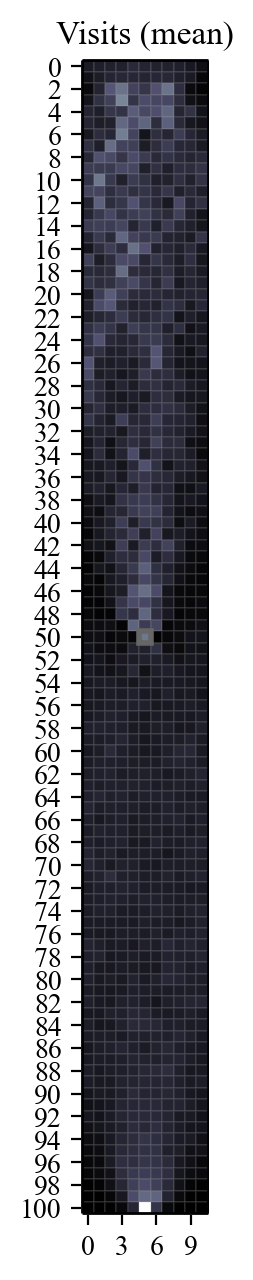}};
        \node (d) at (11,0) {\textbf{(d)}};
        \node[above] (visitsmidbar) at (13,0.3) {\includegraphics[trim=8 0 0 -10,clip,valign=t]{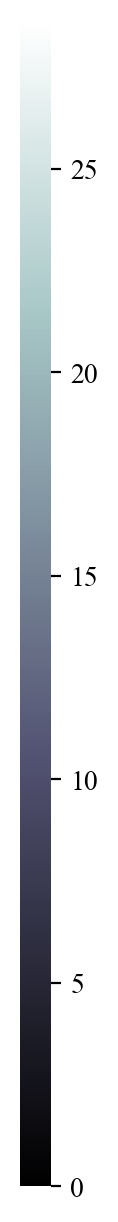}};
    \end{tikzpicture}
    }
    \caption{\label{fig:results-aggregate-tall}This figure shows the learned value function and visit counts in the modified tall domain for our simple reinforcement learning agent exhibiting properties of specific curiosity. This figure shows that learning is slowed when time to complete a cycle of curiosity is increased, and slowed even more when the curiosity-inducing location isn't near any repeatedly visited location. Shown here are the learned value function, $V$, with (a) the curiosity-inducing location at $(95,5)$ and (b) the curiosity-inducing location at $(50, 5)$, and the total visits the agent made to each location. Totals averaged over 30 independent trials of 5000 steps each.}
    \end{figure}
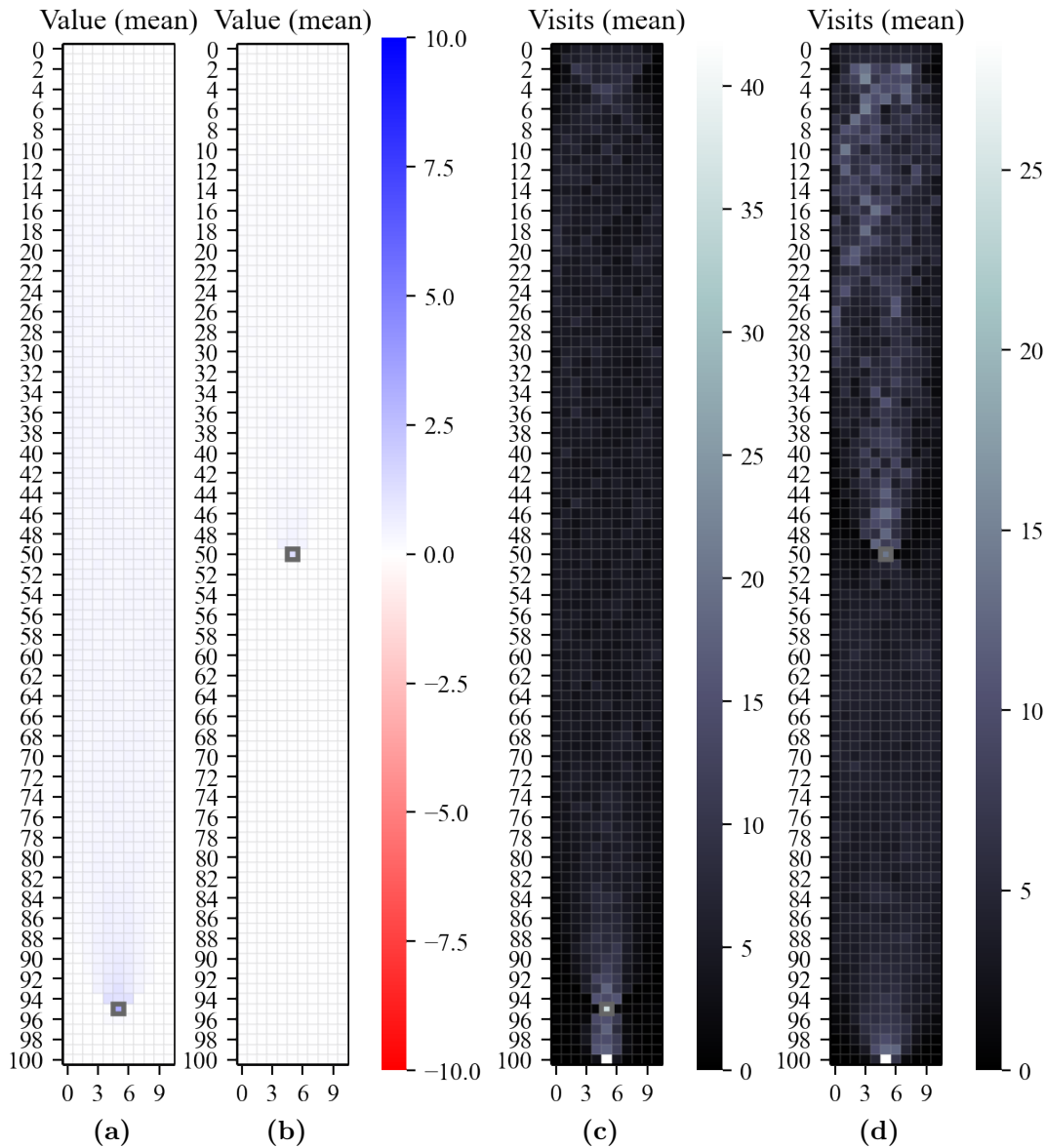
    
    Directedness and cessation when satisfied are determined directly by the algorithm and do not rely on any learning, so it is unsurprising to see these properties reflected in videos of the agent's behaviour. The directed behaviour of the agent is also reflected in the visit counts for the wide and tall domains shown in Figures \ref{fig:results-aggregate-wide}b and \ref{fig:results-aggregate-tall}c,d, primarily in the upward-opening funnel shape from the curiosity-inducing location, which occurs because once the agent is in state curiosity, it only takes upward (or upward diagonal) actions to reach the targets at the top of the grid.
    
    \item \textbf{Any part of the world that is repeatedly visited while the agent is in state curiosity acquires persistent value.} We already saw this phenomenon in the primary domain (Figure \ref{fig:results-aggregate}a,c), as persistent value accumulated in the funnel shape of locations leading from the curiosity-inducing location towards the targets. However, this phenomenon is more pronounced in the wide and tall domains: in Figures \ref{fig:results-aggregate-wide}a and \ref{fig:results-aggregate-tall}a,b, while a direct path from from the junction location to the curiosity-inducing location has accumulated some value, the magnitude of that value is imperceptible on the scale used for those figures, while the upward funnels are clear.
    
    In the wide domain, this upward funnel includes some of the potential target locations (see the distinctive `bird-wing' shape in Figure \ref{fig:results-aggregate-wide}a and the spread of orange lines in Figure \ref{fig:results-lineplot-wide}), which might raise concern if you remember that we were aiming for targets \textit{not} to accumulate value---remember, once you've satisfied your curiosity, you don't read the same page over and over again. However, this is a special case, where these locations accumulate value when they are visited for a different purpose: passing through them on the way to a curiosity-satisfying target. Depending on the context, there may be benefits to learning to value processes that have helped satisfy curiosity in the past, or this may be an undesirable side effect.
    
    \begin{figure}
    \centering
    \includegraphics{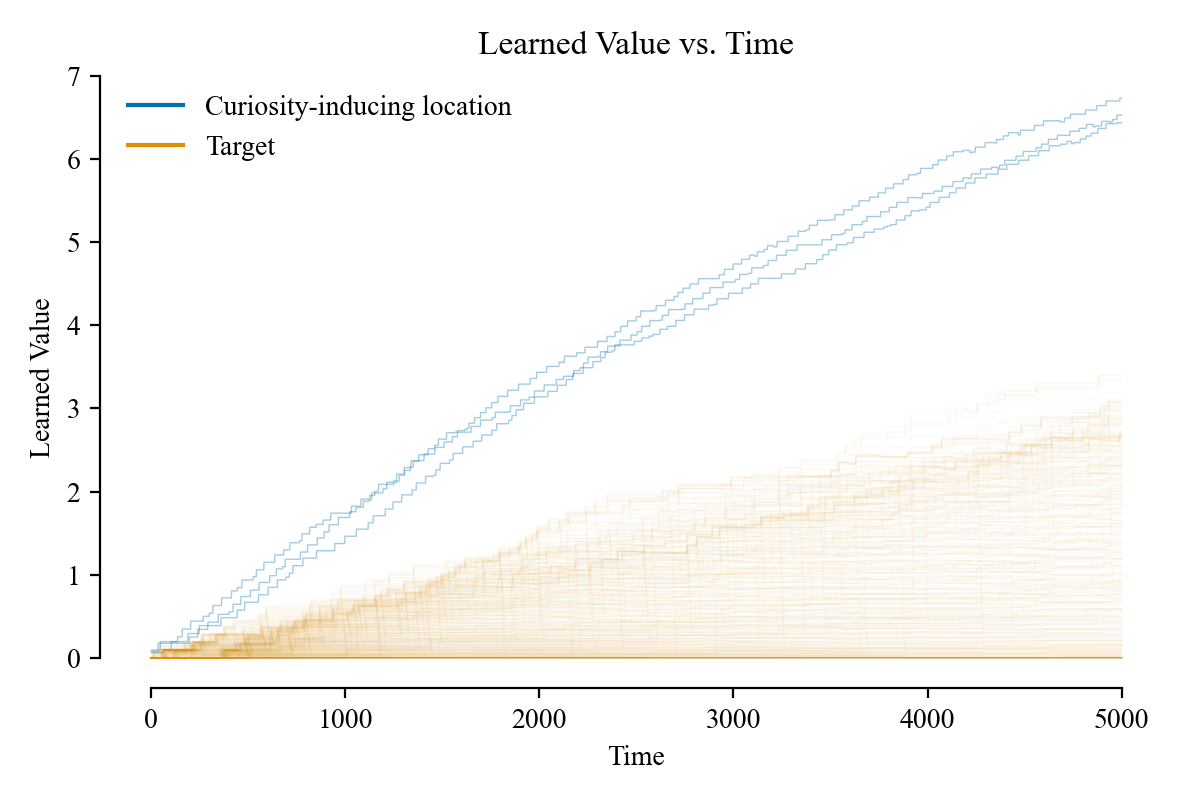}
    \caption{This figure shows the persistent value of the curiosity-inducing location (blue) and target locations (orange) over time for three trials. While the growth pattern for the curiosity-inducing location is similar to that seen for the primary domain (Figure \ref{fig:results-lineplot}), in the wide domain, some of the target locations grow in value over time. There are three blue lines, with each showing the value of the curiosity-inducing location for a single trial. In orange, the value for each target is shown as a separate line (meaning there are 297 separate orange lines, 99 for each trial).}
    \label{fig:results-lineplot-wide}
\end{figure}
\afterpage{\clearpage}

    The accumulation of value in any area visited by the agent while curious is important in the context of our exploration of whether an agent might `get lost' if the curiosity-inducing location is too far away: the agent \textit{can} get stuck in these areas of accumulated value and not find its way back to the curiosity-inducing location. We observed this exact problem when we removed the junction location from the wide world: the agent spends the majority of the trial example trial used to generate Figure \ref{fig:results-single-wide} in an area to the left of the curiosity-inducing location, where it had previously accumulated value on the way to a target. 
    
    \begin{figure}
    \begin{center}
    \textbf{Visit Count (Single Trial in Wide Domain, No Junction Location)}\\
    \makebox[\textwidth][c]{
    \includegraphics[trim=180 20 300 20,clip,valign=top]{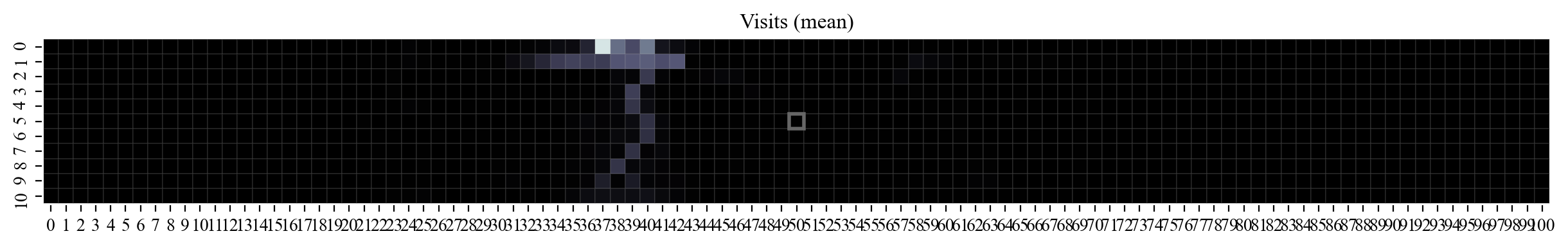}
    \includegraphics[trim=14 8 6 0,clip,valign=top]{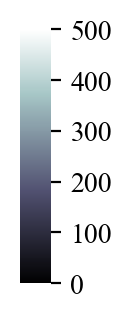}
    }
    16 \hfill 40 \hfill 65    \ \ \ \ \ \ \ \ \ \ 
    \end{center}
    \caption{This figure shows the visit counts in a single trial in the wide domain with the junction location removed. While the agent's persistent value function \textit{is} greatest at the curiosity-inducing location, as desired for voluntary exposure, the agent still doesn't find its way back to the curiosity-inducing location because it gets stuck re-visiting an area of the grid that accumulated value while the agent travelled from the curiosity-inducing location to a target. In this trial, the agent only visited the curiosity-inducing location twice (the first visit resulting in a target to the right of the curiosity-inducing location, and the second resulting in a target to the left of the curiosity-inducing location. }\label{fig:results-single-wide}
    \end{figure}
    
    Thinking this scenario out beyond the 5000 steps of one trial, the agent should gradually learn that this `sticky' area is not valuable. While the agent is not curious, the value of the locations it visits slowly return toward zero. In Figure \ref{fig:results-lineplot-no-junction}, we see that the potential target locations visited repeatedly in this trial gradually decrease in value over time. Because the agent rapidly learned a persistent value function where the curiosity-inducing location has the highest persistent value, after many time steps, it should theoretically return to the curiosity-inducing location once the value of these areas had decreased sufficiently. However, we can see from the shape of those curves in Figure \ref{fig:results-lineplot-no-junction} that this decrease will be ineffectually slow.
    
    \begin{figure}
    \begin{center}
    \textbf{Persistent Value over Time (Single Trial in Wide Domain, No Junction Location)}\\
    \makebox[\textwidth][c]{
    \includegraphics[trim=7 0 7 20,clip,valign=top]{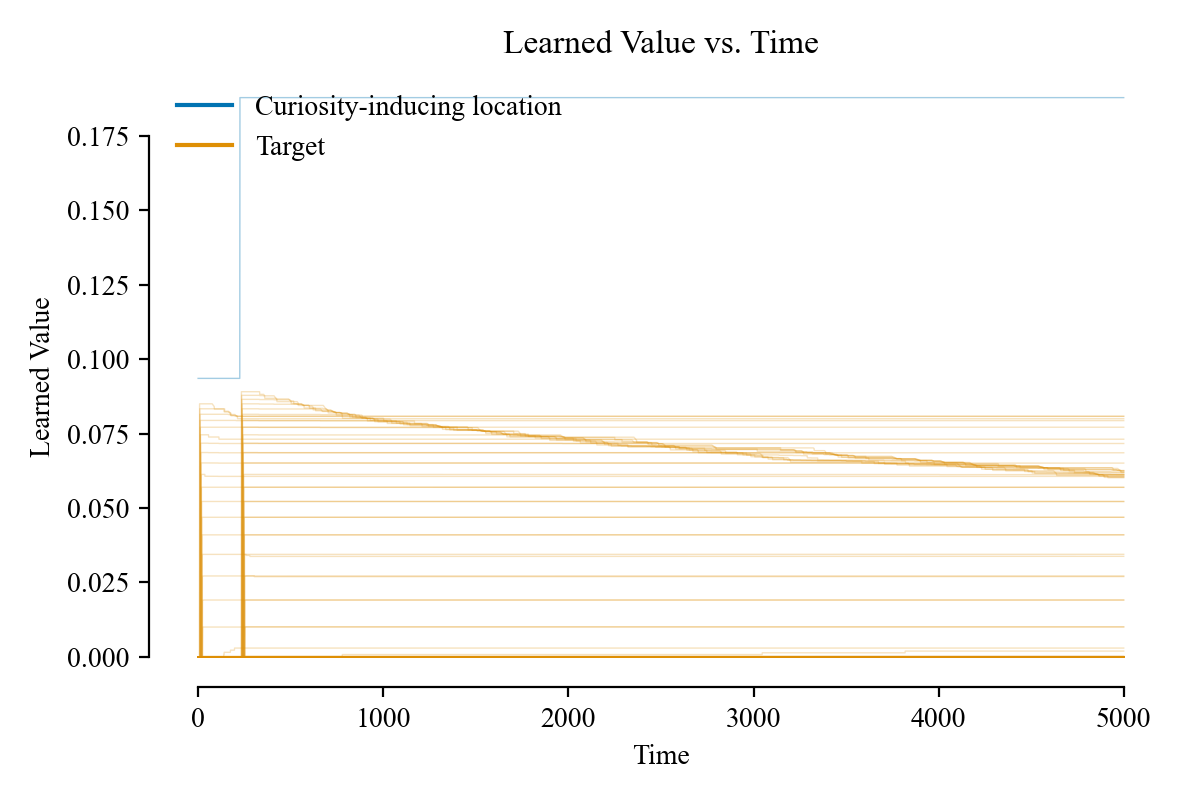}}
    \end{center}
    \caption{This figure shows the persistent value of the curiosity-inducing location (blue) and target locations (orange) over time for one trial in the wide domain with no junction location. While the curiosity-inducing location accumulates the most value, the agent gets stuck re-visiting a region of the grid that was on the way to a previous target. Some of the potential target locations are in this region, and so we can see their value grow when the agent visits them while curious, When the agent returns to these locations after curiosity has been satisfied, their value slowly declines. This decline is so slow that the agent will not unlearn its preference for them in a timeframe that we would consider reasonable. The value for each potential target location is shown in orange as a separate line (meaning there are 99 separate orange lines).}\label{fig:results-lineplot-no-junction}
    \end{figure}
    
    In many cases, we suspect that this limitation would not pose a problem. For example, the existence of a junction location is typical to biological learners: where the curiosity-inducing location is like a bookstore, the junction location is much like a home---a place the agent returns to regularly. Once you've learned a path from your home to the bookstore, you are readily able to follow your desire to expose yourself to curiosity. If you didn't return home, however, you might not figure out how to get back to the bookstore, as we observed in our experiments. This observation of our agent getting stuck is the most extreme example of our third and following lesson.
    
    \item \textbf{Learning voluntary exposure requires multiple visits, and the less likely the agent is to return to a curiosity-inducing location, the slower this learning process will be.} In the wide world with the junction location removed, the agent rarely followed any repetitive path to the curiosity-inducing location. In many trials, the agent visited the curiosity-inducing location more than once, but did not have the opportunity to learn a habitual path. In these grid worlds, a single visit to the curiosity-inducing location extends the learned path by only one location. For readers unfamiliar with the learning behaviour of model-free reinforcement learning algorithms, you can think that, every time the agent stumbles upon a path it has already noted, it notes where it was before entering the path, then follows the path the rest of the way. This new note adds one more location to the path. Algorithmically, these `notes' are made as increased persistent value. This procedure means that while developing increased value for the curiosity-inducing location occurs with even a single visit, developing \textit{behaviour} that reflects voluntary exposure takes multiple visits.
    
    The two experiments in the tall domain reflect our third lesson with more gradation. When the curiosity-inducing location is near the junction location, the agent learns a direct path between the two relatively quickly. When the curiosity-inducing location is placed further away, the agent skips by the curiosity-inducing location more often and spends more time wandering in the part of the domain above the curiosity-inducing location---slowed down by the `sticky' parts of that region that have accumulated value by being visited when the agent is in a state of curiosity. As a result, the curiosity-inducing location accumulates more value and visits overall when it is placed close to the junction location (Figure \ref{fig:results-aggregate-tall}a,c) than when it is placed further away (Figure \ref{fig:results-aggregate-tall}b,d).
\end{enumerate}

These lessons are valuable because, as described in Section \ref{sec:voluntary}, assuming curiosity can be used to direct agents towards fruitful learning opportunities, it is desirable for our agents to effectively and efficiently learn voluntary exposure to curiosity-inducing situations. Using Algorithm \ref{alg:sc} or an adaptation of it will require recognizing the effect of domains on whether the agent will visit a curiosity-inducing location enough times while following its non-curious policy to learn habitual paths. With these lessons in mind, our next set of experiments probes the interplay of the properties within Algorithm \ref{alg:sc}.

\subsection{Third Set of Experiments: Ablation of Properties}

\subsubsection{Experimental Setup: Ablation Study} \label{sec:setup:ablation}

The third and final set of experiments is an ablation study. The term \textit{ablation} comes from neuroscience, where one way to experimentally learn about the function of part of the brain is to destroy that part and see how the behaviour of the learner changes. In our case, particular design elements were included in the algorithm to account for each of three key properties of specific curiosity and in this set of experiments, we ablated (i.e., removed) each of these design elements in turn---directedness, cessation when satisfied, and voluntary exposure---running the same experiment as described in Section \ref{sec:setup:basic} and observe what has changed from the results we observed in Section \ref{sec:resultsdisc:basic}. For each property, a reminder from Sections \ref{sec:agent:directed}-\ref{sec:agent:voluntary} of how the property is incorporated into Algorithm \ref{alg:sc} and a description of how the algorithm proceeds with the property removed is included in the latter part of this subsection.

Beyond using ablations to study the design elements for each key property, in this section we also include an experiment with an ablation of the design element included to account for the aversive quality of specific curiosity. While we pointed out that there is some controversy in whether specific curiosity should be characterized as aversive in Footnote \ref{foot:aversive} and did not argue for aversive quality to be a key property for the implementation of machine specific curiosity, we did include aversive quality in designing Algorithm \ref{alg:sc}, as described in Section \ref{sec:agent:voluntary}. An aversive quality may not be necessary for specific curiosity generally, but removing it should have a notable effect on the results of using Algorithm \ref{alg:sc}, as the aversive quality both guides the agent to the target and determines the value that is learned in the persistent value function, so we tested its importance via an ablation of the associated algorithmic elements, detailed below. 

However, since aversive quality defines the curiosity value function, $R_\textit{curious}$, we expected that removing it completely for an ablation should result in uninteresting, random behaviour: the agent will neither have a guide to the target to use for directedness nor learn to value the curiosity-inducing location for voluntary exposure. For this reason, asking what happens when aversive quality is ablated entirely is less interesting than asking what happens if it is replaced with \textit{positive} quality. How does the agent's learning and behaviour change if $R_\textit{curious}$, rather than being negative everywhere except the target, is \textit{positive} everywhere, most positive at the target? To answer this question, we additionally ran an experiment where we modified the curiosity reward function in this manner, as detailed below.

Running this series of ablations should allow us to better understand Algorithm \ref{alg:sc} by demonstrating how each property contributes to the agent's learning and behaviour. Each of these experiments is described in more detail in the following subsections.

\paragraph{Ablation of Directedness} \label{sec:setup:ablation:directed}
To ablate directedness, we removed Line \ref{algline:directed} and the \textbf{if} statement structure around it. 
\begin{algorithm}
\begin{algorithmic}[1]
\setcounter{ALG@line}{7}
\If{there is currently a curiosity target (ie. the agent is curious)} 
\State $x'$, $R \gets$ {\color{blue} \bf move greedily w.r.t. \boldmath{$V_{curious}(x)$}}  \Comment{{\bf \color{blue} Directed Behaviour}} 
\Else
\EndIf
\end{algorithmic}
\end{algorithm}

In Algorithm \ref{alg:sc}, the agent follows the gradient value function $V_\textit{curious}$ greedily to the target, but in the ablation, the agent instead follows an $\epsilon$-greedy policy with respect to $V$, whether or not a target exists. Equivalently, Line \ref{algline:vaction},
\begin{align*}
    x', R \gets \text{move epsilon-greedily w.r.t. } V(x) &\qquad\triangleright\text{Ties broken uniform randomly,}
\end{align*}
always determines the agent's next action and get the next state, $x'$, and reward, $R$.

\paragraph{Ablation of Cessation When Satisfied}
To ablate cessation when satisfied, we removed Lines \ref{algline:destroy} and \ref{algline:ceases} of Algorithm \ref{alg:sc} and the \textbf{if} statement structure around them. 
\begin{algorithm}
\begin{algorithmic}[1]
\setcounter{ALG@line}{13}
\If{agent observation $x'$ is the target} 
\State destroy the current target 
\State {\color{blue} \bf reinitialize {\boldmath{$V_{curious}$}} to zeros} \Comment{{\color{blue} \bf Cessation when Satisfied}}
\EndIf
\end{algorithmic}
\end{algorithm}

With these lines removed, if the agent visits the target, the target remains and the agent continues to greedily follow the gradient value function $V_\textit{curious}$.

\paragraph{Ablation of Voluntary Exposure}
To ablate voluntary exposure, we removed the edit we made to the learning update in Line \ref{algline:delta}. As a reminder, Line \ref{algline:delta} in the original algorithm was as follows:
\begin{align*}
    \delta \gets R + \gamma \cdot V(x') -  {\color{blue}\boldmath{[ V(x) + V_{curious}(x)] }} &\qquad \triangleright {\color{blue} \textbf{Voluntary Exposure}}
\end{align*}
The ablation reverts that line to the standard TD error, as follows:
\begin{align*}
    \delta \gets R + \gamma V(x') - V(x).
\end{align*}
With the $V_\textit{curious}(x)$ term removed, the temporary value function does not affect updates to the persistent value function.

\paragraph{Ablation of A\kern-0.13em versive Quality}
To ablate aversive quality, we removed Line \ref{algline:aversive}, 
\begin{align*}
    \text{generate } R_{curious} = \left\{
  \begin{array}{@{}ll@{}}
    0, & \text{if transitioning into target state} \\
    {\color{blue} \bf -1}, & \text{otherwise}
  \end{array}\right.  &\qquad \triangleright {\color{blue} \bf \textbf{A\kern-0.13em versive Quality}}
\end{align*}
which accounts for the aversive quality of specific curiosity in Algorithm 1. Without Line \ref{algline:aversive}, $R_\textit{curious}$ remains zero for all state transitions, as $R_\textit{curious}$ was initialized to zero in Line \ref{algline:zeroinit}.

\paragraph{Replacing A\kern-0.13em versive Quality with Positive Quality} \label{sec:setup:ablation:positive}
In addition to ablating aversive quality, we also tested replacing it with positive quality. To achieve this replacement, we modified $R_\textit{curious}$. In the original algorithm, the special reward function, $R_\textit{curious}$, is negative everywhere except at the target, inspired by the aversive quality of curiosity. A different, but still appropriate gradient (temporary value function) could be formulated using an alternative, positive reward model, $\tilde{R}_\textit{curious}$, that would similarly direct the agent towards the target.  While there are many possible definitions, we used the following definition:
\begin{align}
    \tilde{R}_\textit{curious}(s,a,s') = \left\{ \begin{array}{cl}
        1 & \text{if }s\text{ is the target}  \\
        0 & \text{otherwise}
    \end{array}\right.
\end{align}

When $\tilde{V}_\textit{curious}$ is generated via value iteration from $\tilde{R}_\textit{curious}$, it should guide the agent to the target much like the original $V_\textit{curious}$ does. However, in the original learning update in Line \ref{algline:delta}, subtracting the non-positive $V_\textit{curious}(x)$ meant that the agent learned a positive value. To get the same effect with the with the newly defined, non-negative $\tilde{R}_\textit{curious}$, $\tilde{V}_\textit{curious}(x)$ must be \textit{added}; consequently, we modified Line \ref{algline:delta} to the following. 
\begin{align*}
    \delta \gets R + \gamma \cdot V(x') -  {\color{blue}\boldmath{ V(x) + \tilde{V}_{curious}(x) }} &\qquad \triangleright {\color{blue} \textbf{Voluntary Exposure}}.
\end{align*}

\subsubsection{Results and Discussion: Ablation Studies} \label{sec:resultsdisc:ablation}

Our primary result from our ablation studies was that ablating any algorithmic element that supports a key property or that supports the aversive quality results in behaviour that no longer reflects specific curiosity. In particular, the agent no longer exhibits the cycles of curiosity we observed in the primary domain or the wide and tall domains (with junction location). In this section, we will examine the resultant behaviour for each experiment in this set and their implications.

\begin{figure}
    \scalebox{0.95}{%
    \begin{tikzpicture}
    \node[above right] at (0,10.55) {\includegraphics[trim=0 0 0 19,clip,valign=t]{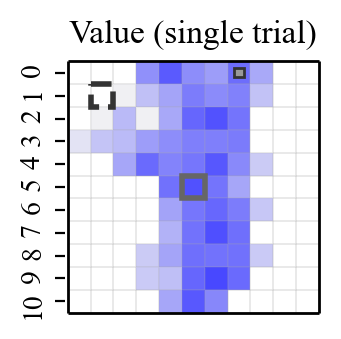}};
    \node[above right] at (4.1,10.6) {\includegraphics[trim=0 0 0 19,clip,valign=t]{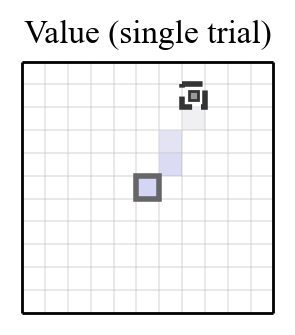}};
    \node[above right] at (7.6,10.6) {\includegraphics[trim=0 0 0 19,clip,valign=t]{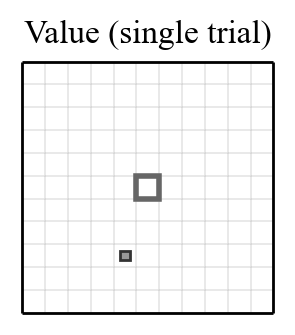}};
    \node[above right] at (11.1,10.6) {\includegraphics[trim=0 0 0 19,clip,valign=t]{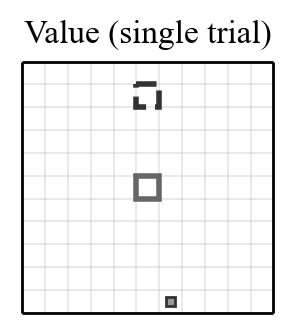}};
    \node[above right] at (14.6,10.49) {\includegraphics[trim=0 4 0 4,clip,valign=t]{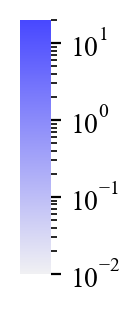}};
    \node[above right] (SingleTrialValueLabel) at (1,14.1) {Persistent Value, $V$ (Single Trial)};
    \node[above right] (alabel) at (1,11) {(a)};
    \node[above right] (elabel) at (4.6,11) {(e)};
    \node[above right] (ilabel) at (8.1,11) {(i)};
    \node[above right] (mlabel) at (11.6,11) {(m)};
    
    \node[above right] at (0,6.65) {\includegraphics[trim=0 0 0 19,clip,valign=t]{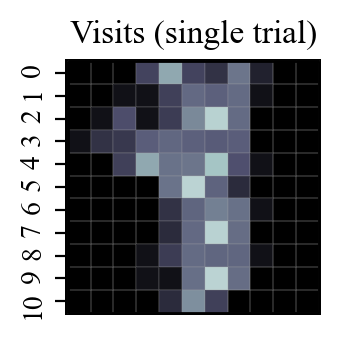}};
    \node[above right] at (4.1,6.7) {\includegraphics[trim=0 0 0 19,clip,valign=t]{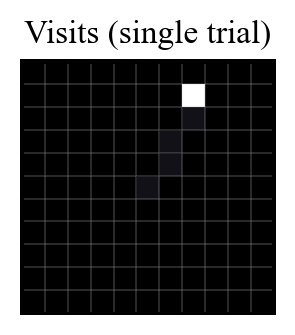}};
    \node[above right] at (7.6,6.7) {\includegraphics[trim=0 0 0 19,clip,valign=t]{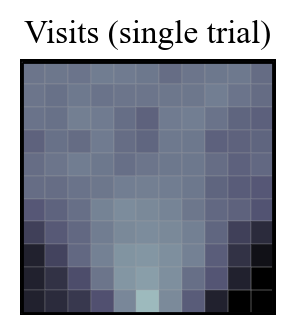}};
    \node[above right] at (11.1,6.7) {\includegraphics[trim=0 0 0 19,clip,valign=t]{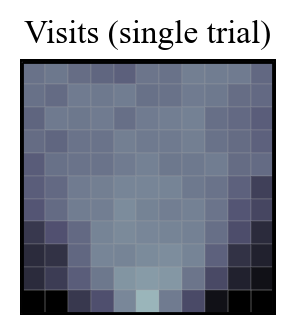}};
    \node[above right] at (14.6,6.7) {\includegraphics[valign=t]{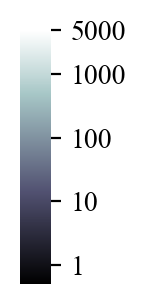}};
    \node[above right] (SingleTrialVisitsLabel) at (1,10.2) {Visits (Single Trial)};
    \node[above right, text=white] (blabel) at (1,7.1) {(b)};
    \node[above right, text=white] (flabel) at (4.6,7.1) {(f)};
    \node[above right, text=white] (jlabel) at (8.1,7.1) {(j)};
    \node[above right, text=white] (nlabel) at (11.6,7.1) {(n)};

    \node[above right] at (0,2.75) {\includegraphics[trim=0 0 0 19,clip,valign=t]{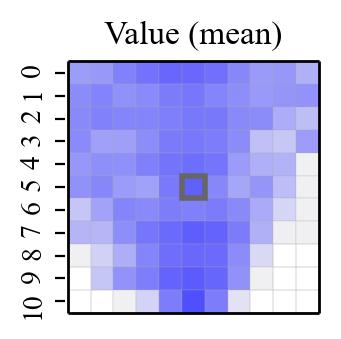}};
    \node[above right] at (4.1,2.8) {\includegraphics[trim=0 0 0 19,clip,valign=t]{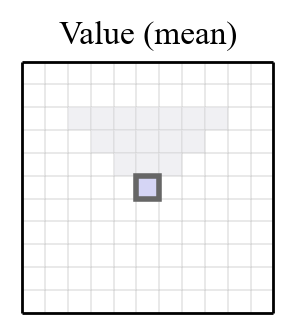}};
    \node[above right] at (7.6,2.8) {\includegraphics[trim=0 0 0 19,clip,valign=t]{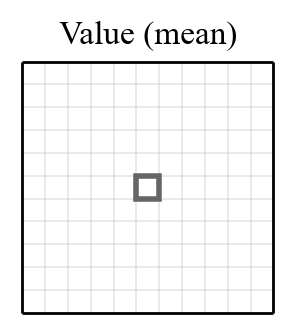}};
    \node[above right] at (11.1,2.8) {\includegraphics[trim=0 0 0 19,clip,valign=t]{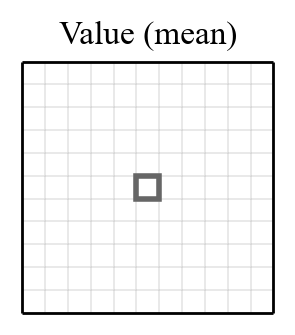}};
    \node[above right] at (14.6,2.55) {\includegraphics[trim=0 0 0 4,clip,valign=t]{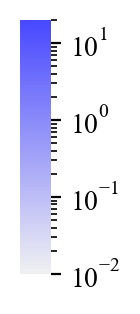}};
    \node[above right] (MeanValueLabel) at (1,6.3) {Persistent Value, $V$ (Mean over 30 trials)};
    \node[above right] (clabel) at (1,3.2) {(c)};
    \node[above right] (glabel) at (4.6,3.2) {(g)};
    \node[above right] (klabel) at (8.1,3.2) {(k)};
    \node[above right] (olabel) at (11.6,3.2) {(o)};
    
    \node[above right] at (0,-1.15) {\includegraphics[trim=0 0 0 19,clip,valign=t]{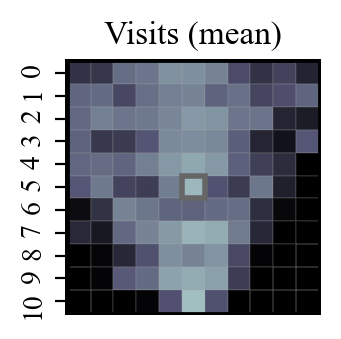}};
    \node[above right] at (4.1,-1.1) {\includegraphics[trim=0 0 0 19,clip,valign=t]{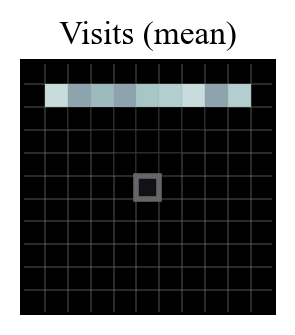}};
    \node[above right] at (7.6,-1.1) {\includegraphics[trim=0 0 0 19,clip,valign=t]{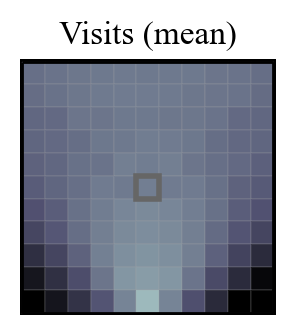}};
    \node[above right] at (11.1,-1.1) {\includegraphics[trim=0 0 0 19,clip,valign=t]{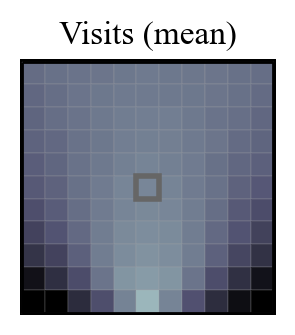}};
    \node[above right] at (14.6,-1.1) {\includegraphics[trim=0 0 0 7,clip,valign=t]{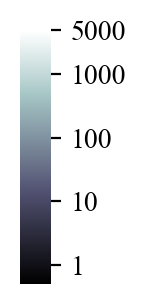}};
    \node[above right] (Visitlabel) at (1,2.4) {Visit Counts (Mean over 30 trials)};
    \node[above right, text=white] (dlabel) at (1,-0.7) {(d)};
    \node[above right, text=white] (hlabel) at (4.6,-0.7) {(h)};
    \node[above right, text=white] (llabel) at (8.1,-0.7) {(l)};
    \node[above right, text=white] (plabel) at (11.6,-0.7) {(p)};
    
    \node[align=center, anchor=north] at (2.6,-0.65) (directed) {Directedness\\Ablation};
    \node[align=center, anchor=north] at (6.1,-0.65) (ceases) {Cessation\\When\\Satisfied\\Ablation};
    \node[align=center, anchor=north] at (9.6,-0.65) (voluntary) {Voluntary\\Exposure\\Ablation};
    \node[align=center, anchor=north] at (13.1,-0.65) (aversive) {Aversive\\Quality\\Ablation};
    
    \end{tikzpicture}
    }\vspace{-21pt}
    \caption{\label{fig:results-aggregate-ablation}This figure shows the persistent value function and visit counts in the primary domain with each ablation. From this figure, we can see that all of the properties are used together to achieve behaviour that learns to value the curiosity-inducing location, but not the targets. A single ablation is shown in each column. The top and third rows show the learned value function $V$ with zero-valued locations in white, while the second and bottom rows show the visit counts with zero-valued locations in black, each after 5000 time steps. The first two rows show a single representative trial for each ablation, while the bottom two rows are averaged over 30 trials. All subfigures are on logarithmic scales.}%
\end{figure}
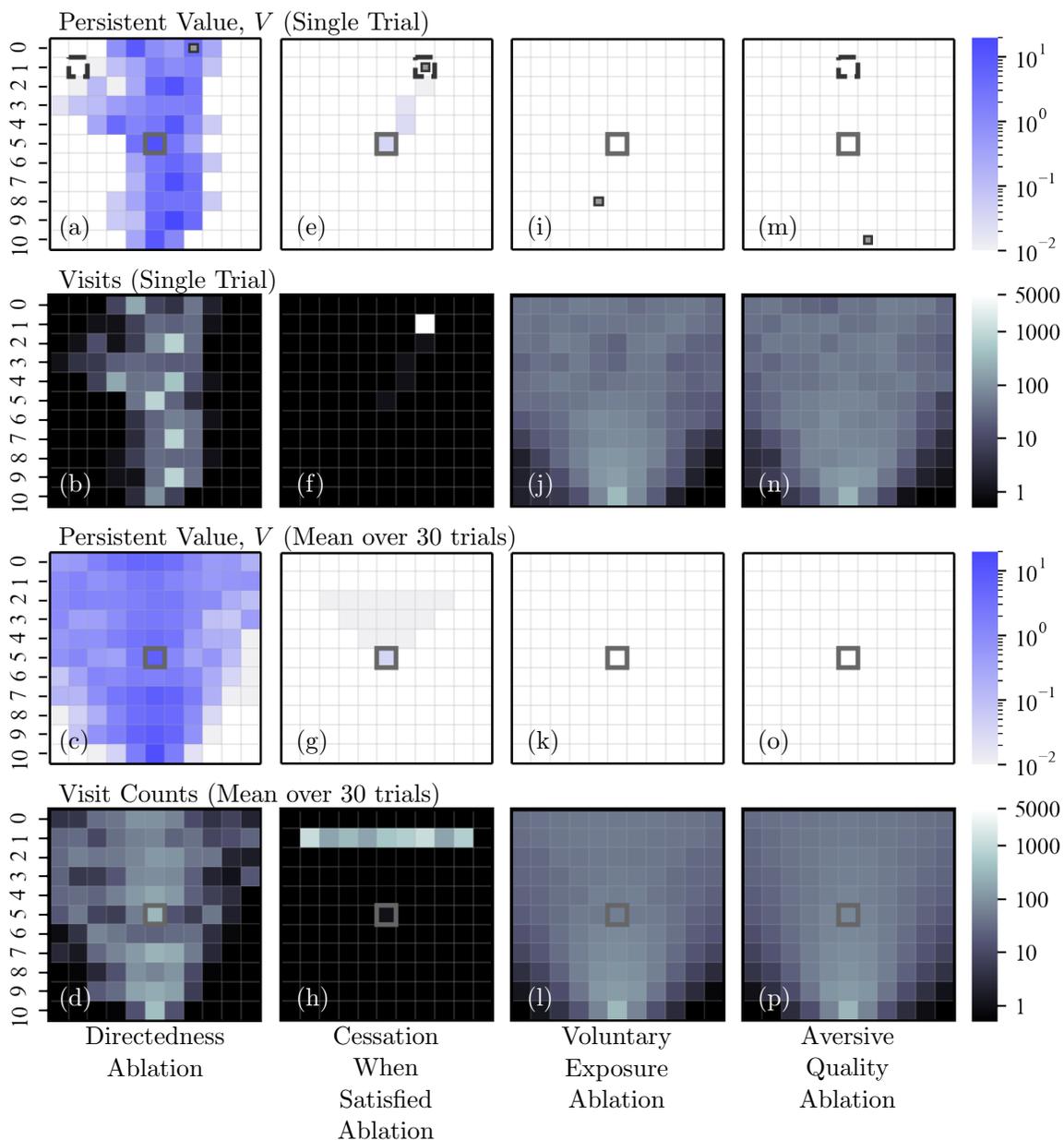

\paragraph{Ablation of Directedness}

\begin{figure}
    \centering
    \includegraphics{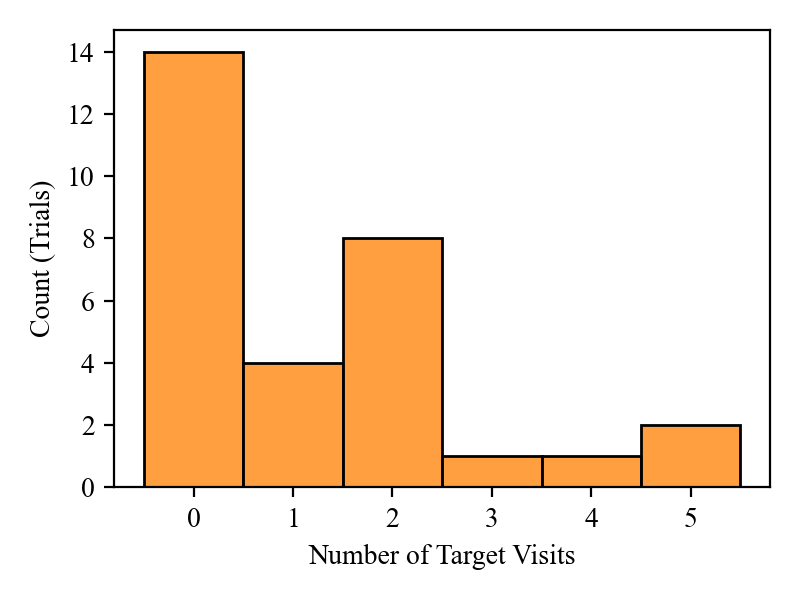}
    \caption{This figure shows a histogram of the number of visits to targets in each trial when directedness is ablated. The histogram is right-skewed. The height of a bar for a given number of target visits is the number of trials with exactly that number of visits. The experiment included 30 trials total.}
    \label{fig:directed_hist}
\end{figure}
\afterpage{\clearpage}

\begin{figure}
\begin{center}
\textbf{Learned Value vs. Time}
\begin{tikzpicture}
    \node (directedness) at (0,0) {\includegraphics{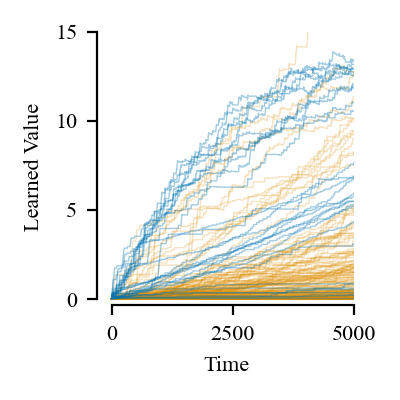}};

    \node (cessation) at (5,0) {\includegraphics{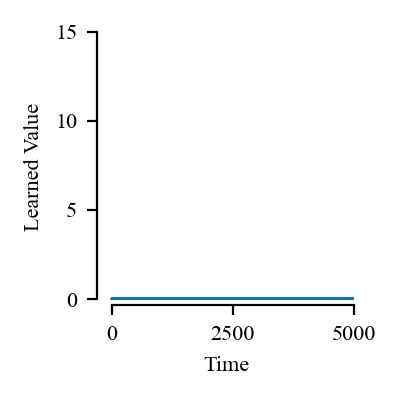}};

    \node (voluntary) at (10,0) {\includegraphics{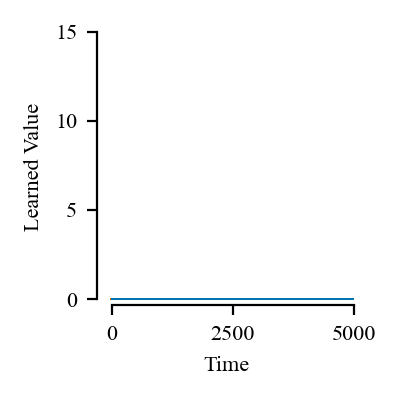}};

    \node (aversive) at (0,-6) {\includegraphics{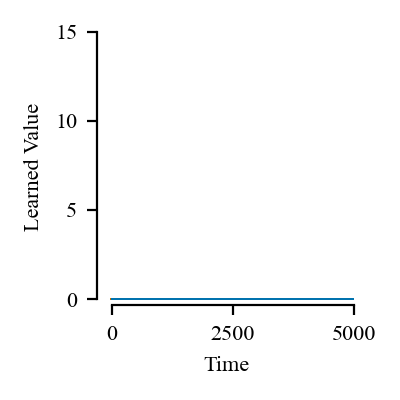}};

    \node (positive) at (5,-6) {\includegraphics{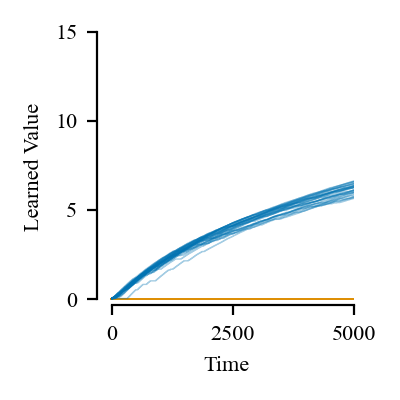}};

    \node (original) at (10,-6) {\includegraphics{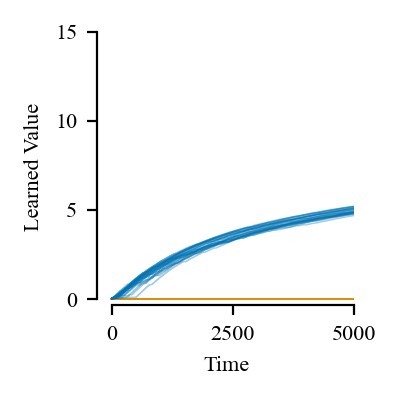}};

    \node[align=center, anchor=north] (dirlabel) at (0.4,-2.3) {Directedness Ablation};
    \node[left=0cm of dirlabel] {\textbf{(a)}}; 
    \node[align=center, anchor=north] (ceslabel) at (5.4,-2.3) {Cessation When \\Satisfied Ablation};
    \node[left=0cm of ceslabel] {\textbf{(b)}}; 
    \node[align=center, anchor=north] (vollabel) at (10.4,-2.3) {Voluntary Exposure\\Ablation};
    \node[left=0cm of vollabel] {\textbf{(c)}}; 
    \node[align=center, anchor=north] (avelabel) at (0.4,-8.3) {Aversive Quality\\Ablation};
    \node[left=0cm of avelabel] {\textbf{(d)}};
    \node[align=center, anchor=north] (poslabel) at (5.4,-8.3) { Positive Replacing\\Aversive Quality};
    \node[left=0cm of poslabel] {\textbf{(e)}};
    \node[align=center, anchor=north] (orilabel) at (10.4,-8.3) {\textbf{Original}\\\textbf{Algorithm \ref{alg:sc}}};
    \node[left=0cm of orilabel] {\textbf{(f)}};
\end{tikzpicture}
\end{center}
\caption{This figure shows the estimated value of the curiosity-inducing location (blue) and target locations (orange) over time for all thirty trials for each ablation and for the original Algorithm \ref{alg:sc}. Panel (f) shows the same data as Figure \ref{fig:results-lineplot}. In the directedness ablation (a), both the curiosity-inducing locations and targets grow over time, with large variation. When cessation when satisfied is ablated (b), the values of both the curiosity-inducing location and the targets remain constant over time, with the value of the curiosity-inducing location reaching $0.315$ during the agent's first and only visit to the curiosity-inducing location at time $t=0$ and the value of the targets remaining $0$ throughout. The learned value for the ablations of voluntary exposure (c) and aversive quality (d) remains zero everywhere. When aversive quality is replaced with positive quality (e), the learned values for the curiosity-inducing location and the targets are similar to those in the original algorithm, but the value of the curiosity-inducing location grows slightly more quickly over time.}\label{fig:results-multilineplots}
\end{figure}
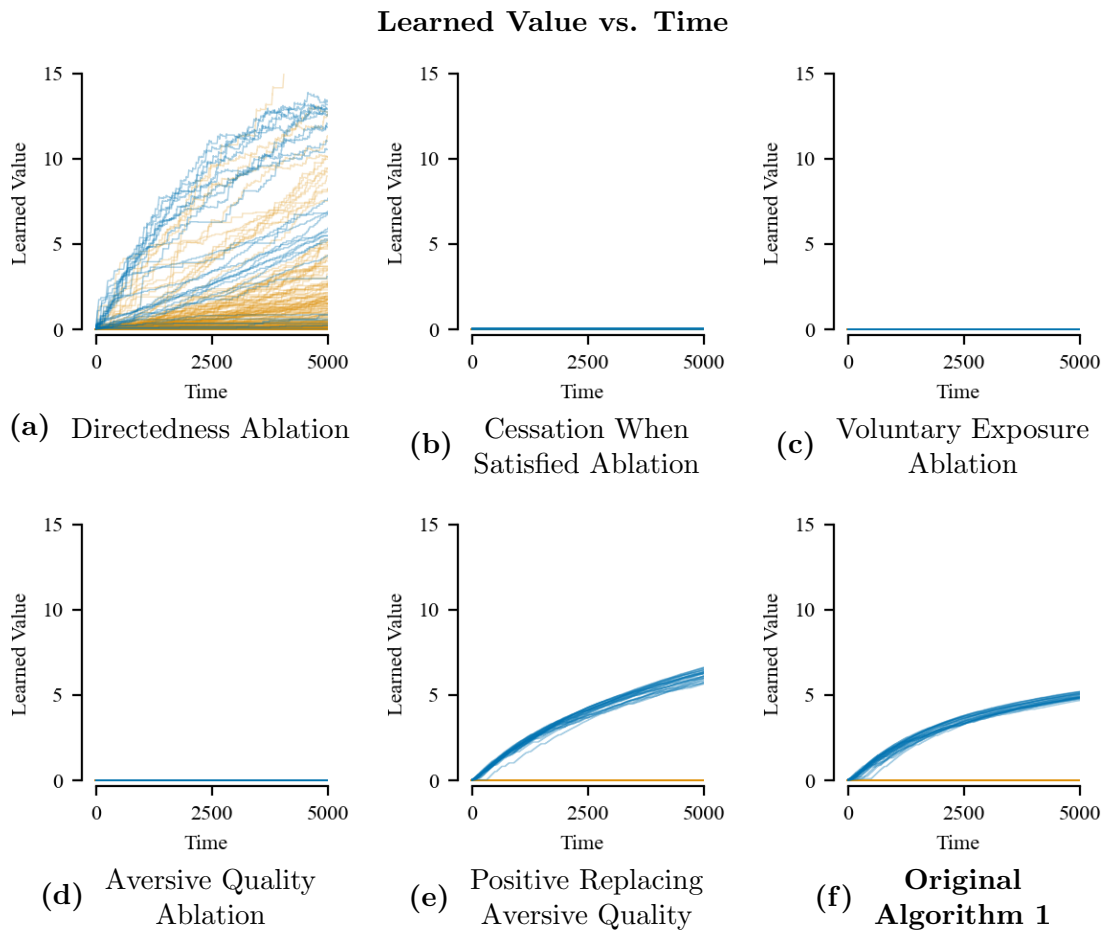

When directedness is ablated, arbitrary paths through the domain accumulate value. This learning behaviour contrasts with what happens when using the original Algorithm \ref{alg:sc}, where direct paths from the curiosity-inducing location to the appropriate satisfier accumulate value (Figure \ref{fig:results-aggregate}a,c). Because the agent with the ablation chooses randomly when faced with equally-valuable maximally-valued alternatives, exactly which path accumulates value varies from trial to trial. This randomness results in the visual difference between the value function for a single trial (top panel of Figure \ref{fig:results-aggregate-ablation}a) and the aggregated value function across trials (third panel from the top of Figure \ref{fig:results-aggregate-ablation}a). Further, the persistent values for the curiosity-inducing location and the potential targets vary substantially from trial to trial, depending on which path through the domain the agent gets stuck on (Figure \ref{fig:results-multilineplots}a). 

\begin{figure}[ht]
\centering
\scalebox{0.95}{%
\begin{tikzpicture}

\tikzstyle{every node}=[font=\small]

\node[anchor=south west] at (0,0) (bot)  {\includegraphics{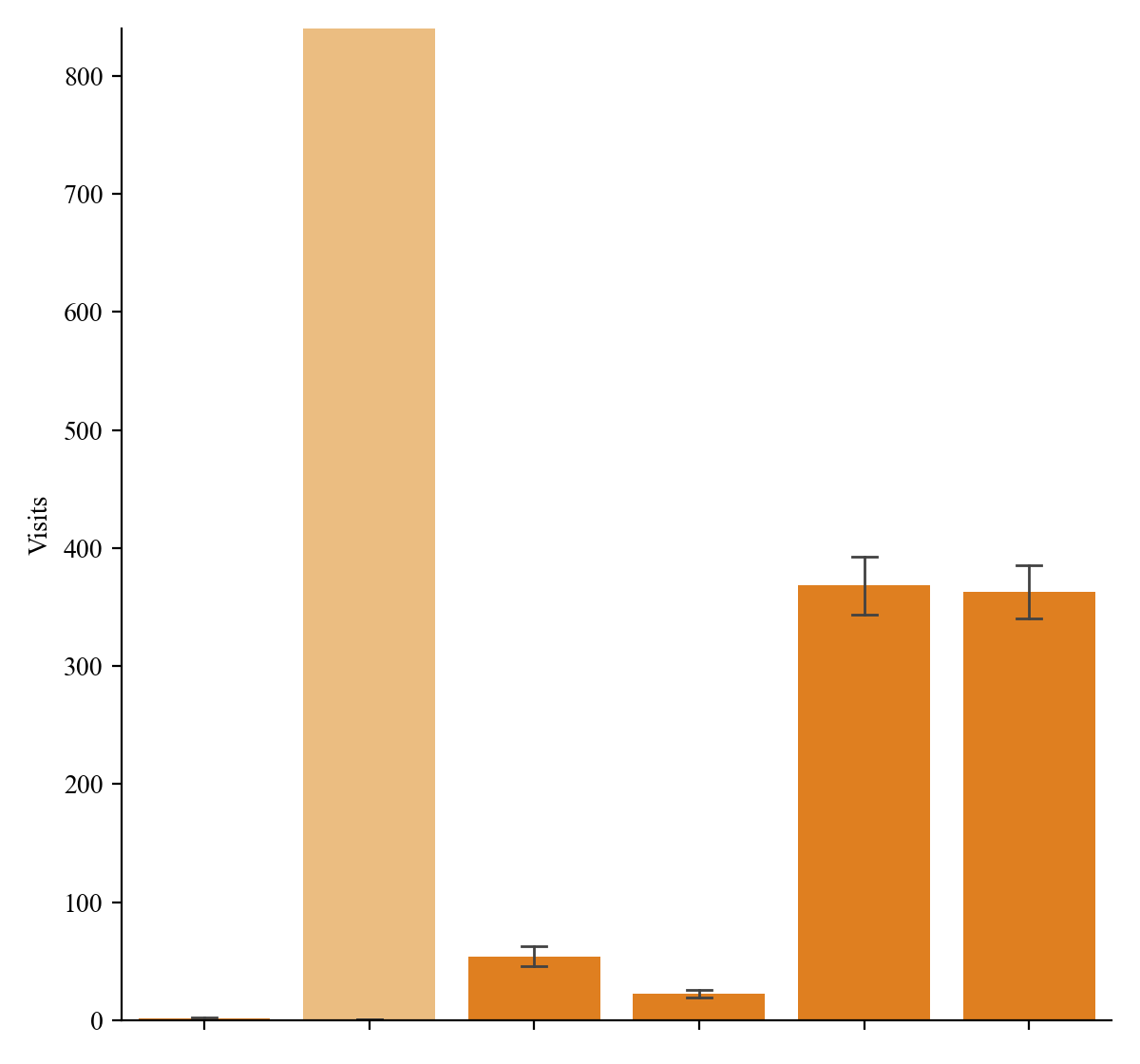}};
\node[anchor=south west] at (6.2,6.9) (top) {\includegraphics{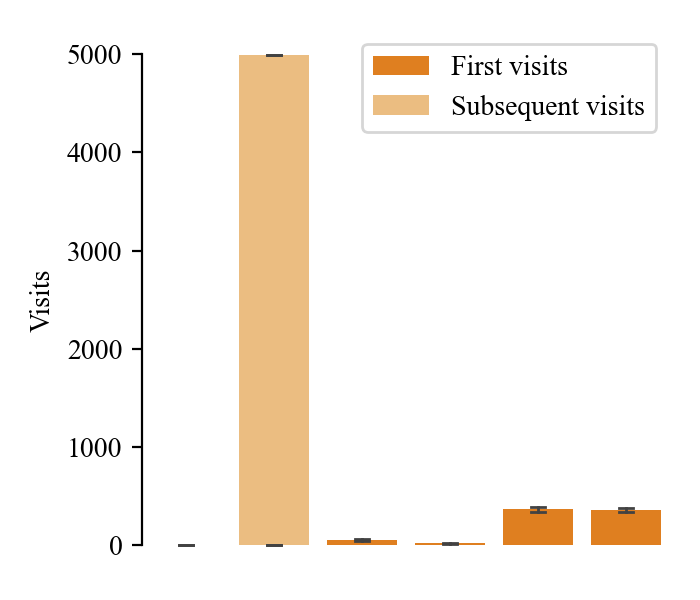}};
\fill [white, path fading=south] (0,14) rectangle (2,13.5); %
\fill [white, path fading=south] (4,13.98) rectangle (6,9); %

\node[align=center, anchor=north] at (2.85,0.6) (directed) {Directedness\\Ablation};
\node[align=center, anchor=north] at (5.1,0.6) (ceases) {Cessation\\When\\Satisfied\\Ablation};
\node[align=center, anchor=north] at (7.3,0.6) (voluntary) {Voluntary\\Exposure\\Ablation};
\node[align=center, anchor=north] at (9.47,0.6) (aversive) {Aversive\\Quality\\Ablation};
\node[align=center, anchor=north] at (11.7,0.6) (positive) {Positive\\Replacing\\Aversive\\Quality};
\node[align=center, anchor=north] at (13.9, 0.6) (original) {\textbf{Original}\\\textbf{Algorithm \ref{alg:sc}}};

\node[align=center, anchor=south] at (2.85, 0.8) {1.2};
\node[align=center, anchor=south] at (5.06, 0.75) {1.0};
\node[align=center] at (5.06, 13.98) {4996.0};
\node[align=center, anchor=south] at (7.28, 1.73) {54.2};
\node[align=center, anchor=south] at (9.47, 1.16) {22.3};
\node[align=center, anchor=south] at (11.7, 6.93) {368.3};
\node[align=center, anchor=south] at (13.9, 6.82) {362.9};

\node[align=center] at (3.5,3) (cesslabel) {First\\visits};
\node at (4.7,0.63) (cessbar) {};
\draw[->] (cesslabel) to [out=270,in=100] (cessbar);
\end{tikzpicture}
}
\caption{\label{fig:barplot}This figure compares the number of times the agent visits a target (specified by the curiosity recognizer, not just possible target locations) for each experiment in Section \ref{sec:setup:ablation} with the original algorithm. The ablation of cessation when satisfied has two stacked bars: dark orange showing the number of \protect\say{first visits} to a target (counting only one visit after the target has been generated) and light orange showing the number of subsequent visits. The upper right inset graph is zoomed out to show the full bar. The original algorithm and the modification replacing aversive quality with positive quality have similar target visit counts while the other ablations result in substantially fewer first visits. Error bars show the standard deviation across 30 trials.} %
\end{figure}

On average, ablating directedness results in the fewest number of visits to generated targets as compared to the original algorithm and the other ablations (mean of $1.2$, comparison shown in Figure \ref{fig:barplot}). While the agent sometimes chooses a path that visits the target, that target is removed once it is visited (cessation when satisfied). Because the agent has already accumulated so much value on its meandering path, it tends to remain on that path. If the next target is not generated on or near that path, then the agent is unlikely to visit it. The result is that the distribution of the number of target visits across trials is right skewed, with the agent failing to visit any targets at all in nearly half of the trials (Figure \ref{fig:directed_hist}).

With directedness ablated, the agent's behaviour is characterized not by cycles of curiosity, but by randomly chosen cycles which continually accumulate more value. The agent does not seek out a satisfier, so unless it stumbles on a satisfier by chance, it can stay in a state of `curiosity,'\footnote{Of course, this state no longer reflects curiosity in any way, and is more reflective of wireheading \citep[for one description of the term wireheading, see][]{yampolskiy2014utility}.} continually accumulating value in a randomly chosen region of the domain with no off switch.

\afterpage{\clearpage}

\paragraph{Ablation of Cessation When Satisfied}

When cessation when satisfied is ablated, the agent takes a direct path from the curiosity-inducing location to the target and remains at that target for the remainder of the trial. In each trial, the agent has one first visit to a target, and 4996 subsequent visits (Figure \ref{fig:barplot}). As an example, the visit counts and persistent value at the end of a single trial are shown in the top two panels of Figure \ref{fig:results-aggregate-ablation}, showing how the agent accumulated persistent value on its path to its first target much like the agent following Algorithm \ref{alg:sc} shown in Figure \ref{fig:results-finegrain}g. Since this ablation agent's target is not removed, the agent does not move on from this location. The agent therefore only visits one target in every trial and does not benefit from curiosity motivating it towards multiple new experiences. 

Of the ablation experiments, the ablation of cessation when satisfied is the only experiment where the agent consistently learns a persistent value function that is maximal at the curiosity-inducing location. Such a value function would reflect voluntary exposure, but since the agent remains fixated on a target, it never has the opportunity to reflect the behaviour component of voluntarily visiting curiosity-inducing situations. Neither the value of the curiosity-inducing location nor the targets changes over time, with the value of the curiosity-inducing location reaching $0.315$ during the agent’s first and only visit to the curiosity-inducing location at time $t = 0$ and the value of the targets remaining $0$ throughout (Figure \ref{fig:results-multilineplots}). Because the agent remains fixated on a single target, the agent spends little time visiting areas with accumulated persistent value, instead spending the rest of its time at the target (Figure \ref{fig:results-aggregate-ablation}e--h).

The removal of cessation when satisfied might remind some readers of the reactive behaviour of intrinsic-reward learners, who are driven to visit a novel state repeatedly. Despite this parallel, the ablation of cessation when satisfied is not directly comparable to intrinsic reward methods. As we discussed in Section \ref{sec:intrinsic:benefits}, multiple computational intrinsic rewards are designed to decay as the agent visits its target over and over. In our ablation, the level of motivation remains static throughout each trial. We experimented with a decaying motivation level, but do not include the (rather uninteresting) results here because the conceptual purpose of intrinsic rewards is so unlike that of specific curiosity that the comparison is inappropriate in our test domain. Again, two primary benefits of this decaying property of intrinsic rewards are promoting multiple visits to check for consistency (for example of a stochastic reward) or staying on an exploration frontier. In our simple, rewardless domain, there is no benefit to repeated visits, nor are the curiosity targets generated on an exploration frontier.

\paragraph{Ablation of Voluntary Exposure}

When voluntary exposure is ablated, no persistent value accumulates in any part of the domain (Figure \ref{fig:results-aggregate-ablation}c and the line plot in Figure \ref{fig:results-multilineplots} are zero everywhere). This occurs because the learning update step that flips value from the curiosity value function into the persistent value function has been removed. However, the agent does still demonstrate directed behaviour between the curiosity-inducing location and the targets. As a result, there is a faint but visible funnel shape above the curiosity-inducing location in the bottom panel of Figure \ref{fig:results-aggregate-ablation}c (compare with the bottom panel of Figure \ref{fig:results-aggregate-ablation}d, which reflects a true random walk through the domain). This directed behaviour helps the agent make more (first) target visits than any of the other ablations (mean of $54.2$, see Figure \ref{fig:barplot}), though still far fewer than an agent following the original Algorithm \ref{alg:sc}.

\paragraph{Ablation of Aversive Quality}

When the aversive quality of curiosity is ablated, $V_\textit{curious}$ is not generated, so the agent experiences no difference in value or reward throughout the domain. For this reason, the agent acts randomly throughout each trial. The resulting estimated value function and visit counts are shown in Figure \ref{fig:results-aggregate-ablation}(d). No value is accumulated anywhere in the grid, as emphasized by Figure \ref{fig:results-multilineplots}(d), which shows that the estimated value for all of the targets and the curiosity-inducing location remain zero throughout each trial.

\paragraph{Replacing Aversive Quality with Positive Quality}

\begin{figure}
\centering
\makebox[\textwidth][c]{
\begin{tikzpicture}
    \node (value1label) at (1.75,2) {\textbf{Value (single trial)}};
    \node (value1positive) at (0,0) {\includegraphics[trim=0 0 0 20,clip,valign=t]{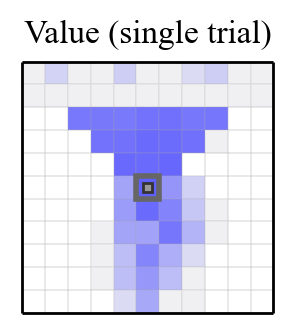}};
    \node (a1) at (-1.2,-1.2) {\textbf{(a)}};
    \node (value1aversive) at (3.5,0) {\includegraphics[trim=0 0 0 20,clip,valign=t]{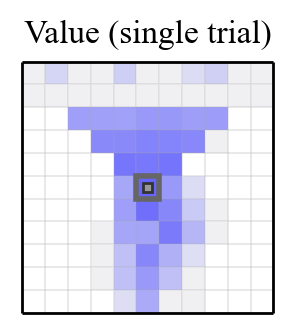}};
    \node (b1) at (2.3,-1.2) {\textbf{(b)}};

    \node (valuemlabel) at (8.75,2) {\textbf{Value mean}};
    \node (valuempositive) at (7,0) {\includegraphics[trim=0 0 0 20,clip,valign=t]{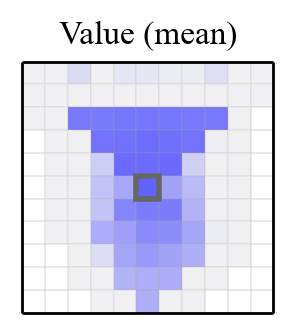}};
    \node (a2) at (5.8,-1.2) {\textbf{(a)}};
    \node (valuemaversive) at (10.5,0) {\includegraphics[trim=0 0 0 20,clip,valign=t]{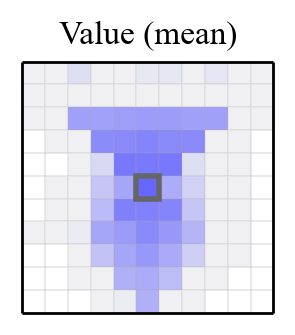}};
    \node (b2) at (9.3,-1.2) {\textbf{(b)}};
    
    \node (value1bar) at (13,0) {\includegraphics[trim=0 0 0 -1,clip,valign=t]{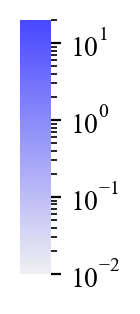}};

    \node (visits1label) at (1.75,-2) {\textbf{Visits (single trial)}};
    \node (visits1positive) at (0,-4) {\includegraphics[trim=0 0 0 20,clip,valign=t]{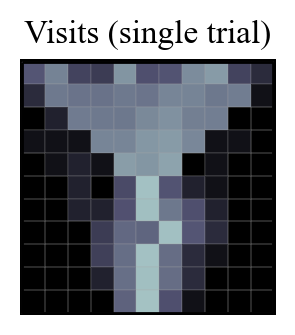}};
    \node[text=white] (a3) at (-1.2,-5.2) {\textbf{(a)}};
    \node (visits1aversive) at (3.5,-4) {\includegraphics[trim=0 0 0 20,clip,valign=t]{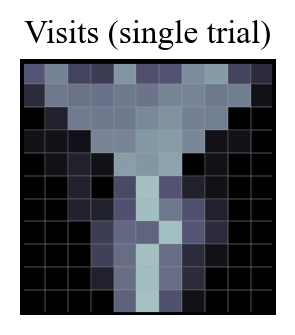}};
    \node[text=white] (b3) at (2.3,-5.2) {\textbf{(b)}};

    \node (visitsmlabel) at (8.75,-2) {\textbf{Visits (mean)}};
    \node (visitsmpositive) at (7,-4) {\includegraphics[trim=0 0 0 20,clip,valign=t]{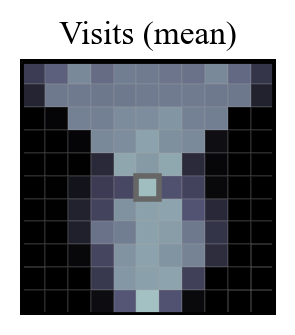}};
    \node[text=white] (a4) at (5.8,-5.2) {\textbf{(a)}};
    \node (visitsmaversive) at (10.5,-4) {\includegraphics[trim=0 0 0 20,clip,valign=t]{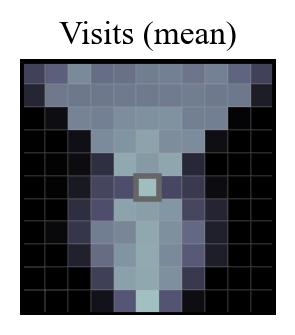}};
    \node[text=white] (b4) at (9.3,-5.2) {\textbf{(b)}};

    \node (value1bar) at (13,-3.85) {\includegraphics[valign=t]{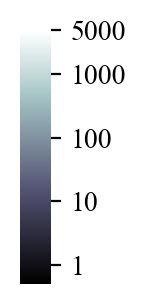}};
\end{tikzpicture}
}
\caption{This figure shows the final learned value functions and visit counts for the experiment where (a) aversive quality is replaced with positive quality alongside the same for (b) the original algorithm (same data as Figure \ref{fig:results-aggregate}, but on a logarithmic scale) for visual comparison. The behaviour and value learned with positive quality (a) is very similar to that of the original algorithm (b)---indeed, given the same random seed, the behaviour is identical for 1071 steps---but value accumulates at different rates in each case, so the value functions do differ by more than just scale. All subfigures are on logarithmic scales.}
\label{fig:results-aggregate-modblation}
\end{figure}
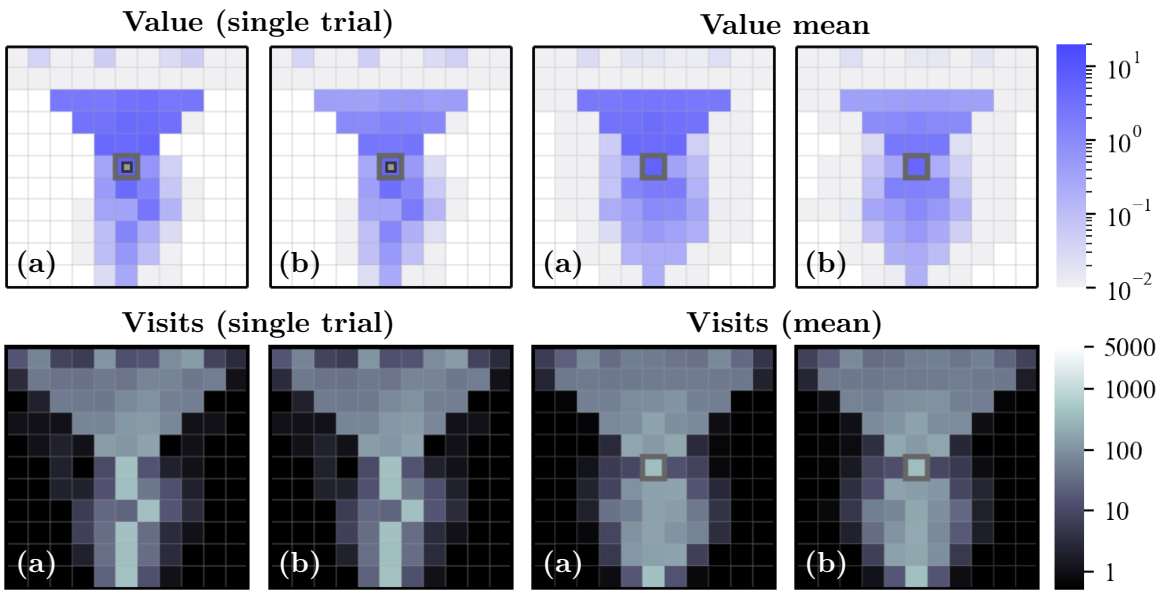
\afterpage{\clearpage}

More interesting than ablating aversive quality is replacing it with positive quality. In this experiment, the agent's behaviour is very similar to that of of the agent following original Algorithm \ref{alg:sc} as described in Section \ref{sec:resultsdisc:basic}. The number of visits to generated targets for the agent with this replacement are within error of that of the original algorithm, shown in Figure \ref{fig:barplot}. Both agents' final persistent value functions and visit counts are similar (Figure \ref{fig:results-aggregate-modblation}). The main difference between the persistent value functions is a matter of scale, in that the estimated values for the experiment using positive quality are generally higher. This difference is also visible in the associated lineplot in Figure \ref{fig:results-multilineplots}, where the value of the curiosity-inducing locations grows more quickly when aversive quality is replaced with positive quality. However, the difference is not only in scale; for example, note that squares $(7,0)$ and $(4,1)$ have different mean values between in Figure \ref{fig:results-aggregate-modblation}a (positive quality) and \ref{fig:results-aggregate-modblation}b (aversive quality).

The agent using positive quality should and does behave differently than the agent following the original Algorithm \ref{alg:sc}, because the value functions generated by $R_\textit{curious}$ and $\tilde{R}_\textit{curious}$ have different shapes. For this reason, the persistent value function accumulates value at different rates in each case. However, a takeaway from this experiment is that using a negative value function, or what we call the aversive quality of Algorithm 1, is not necessary for creating cycles of behaviour reflecting specific curiosity. 

In humans, it may be true that the information seeking associated with specific curiosity ``is motivated by the aversiveness of not possessing the information more than it is by the anticipation of pleasure from obtaining it" \cite[p.~92]{loewenstein1994psychology}, but from the perspective of our simplistic computational RL agent, our choice of implementation for each did not result in appreciably different behaviour.
\sectionbreak

Taken together, the experiments in our ablation study show us that, in the context of Algorithm 1, the properties of directedness, cessation when satisfied, and voluntary exposure work together, and that curious behaviour is noticeably impaired when any one property is missing. 

\section{General Discussion: Benefits of the Properties of Specific Curiosity} \label{sec:benefits}

Our ablation study provides initial evidence for the interconnected nature of the properties of specific curiosity---effective learning behaviour isn't achieved via one or two properties; the properties work together. Indeed, the benefits of each property are so interwoven that they are best understood via their combined influence on the whole of specific curiosity. 

\paragraph{Flexible specialization to a learner's context:} In Section \ref{sec:coherent}, we noted that the property of coherent long term-learning, the last of our five properties, closes the loop of how curiosity can guide a learner over a lifetime. Curious biological learners, including humans, live long lives, but certainly not long enough to experience every possible situation that the world could throw at them. Further, humans have found ways to survive in a diverse set of possible climates, cultures, and contexts. We believe specific curiosity supports that ability.

Some of what we learn is passive---we learn just by `being there.' Our brains persistently and automatically take the observations from our senses and work to integrate them into our knowledge of the world \citep[p.~138]{chater2018mind}. This passive learning helps build up a foundation of knowledge that is somewhat local to the learner's particular context. Then, specific curiosity insists that we learn actively, almost any time we aren't attending to obvious needs to keeping our bodies going and species alive. And in particular, the property of coherent long-term learning biases our active learning towards specific concepts that we are ready to build onto our existing knowledge \citep{wade2019role}, often towards new information defines a connection across a gap in our existing knowledge \citep{loewenstein1994psychology}, much of which may have been passively learned. The better connected our knowledge is, the more useful it is. 

Very importantly, curiosity supports us when our context changes. By being biased to direct the learner towards information to support connections to the learners' existing knowledge, specific curiosity may direct us to learn new information that will help us transfer our existing skills and knowledge into a novel context. How many of our curiosity questions start by orienting on ``Wait, that wasn't what I was expecting"? In those kinds of situations, whether we observed a toy performing an unexpected function or a suspect in our mystery performing a suspicious action, there is a waiting connection to be made. The jack-in-the-box doesn't appear except when ...? People don't dump heavy body-sized bags into the lake in the dead of night except when ...? Dark fluid doesn't end up on white-paper walls except when ...? In these situations, making our inostensible referent ostensible repairs the broken understanding created by our prior generalizations failing to hold in a new context, giving us a more accurate foundation of knowledge on which to act.

\paragraph{Specialization as contribution to societal knowledge:} Looking at our favourite biological model of curiosity, the human, another key feature of humans is that they are social. Humans in particular seem to get an incredible benefit from individuals having different specialties \cite[p.~7]{hauser2018human}. If each individual instead developed unspecialized, broad knowledge, then the overlap---the knowledge held by our entire society---would be similarly broad, but unfortunately shallow. We would know very little about many things, as a group. Instead, the overlap of all these narrow, deep specializations developed over time lends itself to providing not only broad, but deep knowledge for our larger society, networked together by humanity's social nature.

When a piece of specialized knowledge turns out to be generally applicable, it can be transferred via social contact across a connected network of learners, a more general societal benefit. While we noted that humans are our favourite model of curiosity, the societal transmission of new, specialized behaviours---innovations---appears to benefit social non-human animals too. One example involves birds, British blue tits, who famously discovered how to pierce the foil caps on milk bottles to access the cream on top. The behaviour was first observed in 1921, but by the end of the 1940s, the behaviour was widespread across the U.K. (\citealp[p.~1226]{aplin2013milk}, \citealp{yong2014scientists}). Experiments by \cite{aplin2013milk} involving teaching new foraging behaviours to blue tits have provided further evidence that blue tits socially transmit new, useful behaviours across their communities (p.~1230). 

As another example, researchers on the isolated Japanese islet of Koshima observed a macaque (a variety of monkey) washing the sand off of a potato---a new behaviour that they had never observed before \citep[pp.~2--3]{kawai1965newly}. In the years thereafter, the researchers observed a wave of social learning until nearly the whole colony seemed to clean their potatoes before eating (p.~4). Interestingly, the same macaque who seems to have come up with the potato-cleaning behaviour appeared to later be the first macaque to demonstrate a behaviour of ``wheat washing" (p.~13). Initially, when humans scattered wheat across the sand, the monkeys would painstakingly pick up each grain one by one. ``Wheat washing," on the other hand, involves gathering up the sand with the grains and tossing them into water, which allows the sand to drift to the bottom while the wheat floats on top (p.~12). This behaviour also spread throughout the colony, though not quite as pervasively (p.~12). In analogy with the specialized human chef who might design new recipes and share them, the originator of these behaviours might have a specialized interest in food preparation, to the benefit of their community.

\paragraph{The need for directedness towards inostensible referents:} Coherent long-term learning requires directedness towards inostensible referents. An inostensible concept, supported by the properties that the learner already knows will be true of the inostensible referent, is the form taken by the next---metacognitively most appropriate (\citealp[]{wade2019role})---learning opportunity to coherently build on existing knowledge. The only sensible activity to experience curiosity-satisfying observations is to take a systematic sequence of actions to obtain the specific information that will make their inostensible concept ostensible. Given that the learner will never have a perfect model of the world including the inostensible referent (it wouldn't be inostensible, in that case!), the learner must make a best guess and adapt their plan as they proceed. 

\paragraph{The usefulness of cessation when satisfied:} Cessation when satisfied creates efficiency by taking advantage of the following idea: what makes an appropriate answer depends on the question. For some inostensible referents, repetitive behaviour might be appropriate: just think back to the example with the peculiar-sounding floorboard. A reasonable way to acquire sufficient evidence to decide if your weight transfer caused the noise is indeed to try repeating that weight transfer several times---once might be a fluke, but three or four times seems sufficient to suggest you're causing the noise. 

Our formulation of cessation when satisfied was directly inspired by the behaviour generated by intrinsic-reward methods and how it contrasts with specific curiosity. The reactive nature of intrinsic rewards motivate a learner to re-experience a state multiple times. Specific curiosity, on the other hand, doesn't require this kind of repetition for all inostensible concepts. For many questions, only single experiences of each curiosity-satisfying observation is required. After all, you don't need to re-read `whodunnit' out of curiosity---once you've read that part once, your curiosity for that particular inostensible referent can end.

In most cases, there are multiple possible forms of evidence that we would accept as curiosity-satisfying. One of the seemingly most important for humans is testimony from others \citep{harris2012trusting}. This kind of evidence rarely requires repetitive behaviour (unless the person you're asking isn't listening). If anything, it may require probes into how reliable the source of information is, or seeking a second opinion via a different mode of behaviour. Not only does the kind of evidence required vary depending on the inostensible referent, the reliability required of an the answer varies even further. How important is it that we have the right answer, versus just a working theory?

In this way, humans demonstrate extreme flexibility when it comes to specifying what makes an acceptable curiosity-satisfying situation. While our next prototypes of curious machines may not have such beautifully tailored recognition systems for sufficient evidence for their curiosity to be satisfied, it is time to move away from simple repetition as a proxy for the satisfaction of curiosity.

\paragraph{The importance of transience:} A close relative of cessation when satisfied, transience is necessary for functional curiosity in biological learners. After all, humans and animals can only (physically) be in one place at one time, and their attention is thought to be a similarly limited resource \cite[p.~2]{lloyd2018interrupting}. Constantly reorienting those limited bodily and attentional resources is impractical, and so committing to a single goal for a period of time benefits the learner \cite[p.~2]{lloyd2018interrupting}. Specific curiosity is one example of this kind of goal-directed behaviour. As detailed by \citet{lloyd2018interrupting}, goal-directed behaviour will be more effective in an uncertain environment if the behaviour of the agent can be interrupted by time-sensitive demands, like attending to a loud noise that might indicate danger, pangs of hunger \citep[p.~35]{simon1967motivational}, or even the recognition that, in the past, you regretted a decision made in a similar situation  \citep[p.~498]{hoch1991time}. 
    
In this sense, transience also has a strong relationship with stay-switch decisions observed in animal decision making, wherein an animal constantly balances its near-term reward with its expectations of long-term average reward, thereby governing the persistence of its current behaviour (c.f., human patch foraging and the marginal value theorem; \citealp{constantino2015learning}). 
    
Even more critically, transience resolves some of the trouble that `un-realizable' inostensible concepts could cause. When we say that some inostensible concepts are un-realizable, we are noting that the very nature of inostensible concepts is that, in some cases, they can't be made ostensible. Not everything that could be dreamt up by a learner is necessarily a thing that the learner could find, especially if the lifetime of the learner is limited. While I could find myself curious about the location of the nearest Earth-orbiting teapot, I would struggle to find out whether such a teapot exists, never mind its location. When asking about unknowns, it is necessary that a learner might sometimes ask the wrong questions, and so needs to be able to stop chasing curiosity-satisfying situations that don't exist.

The condition of specific curiosity is a concerted effort to make an inostensible concept ostensible. Directedness towards inostensible It requires adaptive planning, which is likely resource-heavy, and, in biological learners, active movement of the body towards perceiving curiosity-satisfying observations. Transience helps the learner manage an all-or-nothing effort to satisfy their curiosity, because it means that behaviour and use of attentional resources can be fully reallocated to other matters as needed.

\paragraph{Voluntary exposure over curiosity by chance:}
Accepting the premise that curiosity will be valuable to our machine agents, we certainly don't want our agents to avoid curiosity. But do we really want voluntary exposure, or would it be sufficient for the agent to stumble across curiosity-inducing observations without increased preference for them?
    
Before we provide our answer to that question, we would like to note some subtlety to the voluntary exposure that humans exhibit. Humans have been observed to voluntarily expose themselves to some observations that they are aware will be curiosity-inducing \cite[pp.~75-76]{loewenstein1994psychology}, like a puzzle or the latest bingeable TV show, but there are other curiosity-inducing observations that humans will not choose to expose themselves to. 
    
\citet{ruan2018teasing} presented the results of some experiments where humans exhibited specific curiosity, but not voluntary exposure. Their experiments centred on what they called an ``uncertainty creation--resolution process" (p.~556). In their experiments, this process consisted of the learner being ``first teased with some missing information" (e.g. presented with a trivia question) ``and then given that information" (p.~556). In four experiments (see the discussions of \textit{Choice} for Studies 1 through 4, pp.~561--565), they found that, given a choice between experiencing an `uncertainty creation--resolution process' or not, most of their participants chose not, suggesting that they did not exhibit voluntary exposure.
    
The authors offered two hypotheses about why their participants failed to exhibit voluntary exposure. One hypothesis was that seeking uncertainty, or choosing to be exposed to curiosity-inducing observations, might be a trait exhibited by a minority of people \citep[p. 560]{ruan2018teasing}. The very healthy industries producing puzzles, mysteries, and cliff-hanger-laden television series that we mentioned in Section \ref{sec:voluntary} bring this hypothesis into doubt. Their other hypothesis was that, in cases where people voluntarily expose themselves to curiosity-inducing situations, they ``have control over when they receive the missing information" \cite[p.~560]{ruan2018teasing}, which merits further study. 

Based on our computational case study, we suggest a novel hypothesis that voluntary exposure might be learned via multiple experiences of curiosity being induced in similar situations. It is possible that while these people have learned to predict the positive experience associated with their favourite forms of curiosity-inducing situations, be they crossword puzzles, mystery novels, or mathematical problems, the experimental setup might be too unfamiliar to lead to voluntary exposure. In this way, considering the value of voluntary exposure brings us back to coherent long-term learning. Tying voluntary exposure to individual interest enhances learner specialization, a key benefit of coherent long-term learning as we argued above.

Whatever domains we specialize our voluntary exposure towards, specific curiosity tends to drive us into a solving process. Whether racking our brains for the right word for a crossword or picking out the right clues to solve a murder mystery, curiosity helps us build and solidify our knowledge. In particular, human learning benefits from retrieval practice, and curiosity helps us when we're in danger of forgetting something we have already been exposed to, and if that something is coming up again, it is likely a somewhat consistent part of the context we interact in day-to-day. Learners have to practice to develop skills, so if we don't have to attend to a more pressing matter like food or sleep or whatever, practicing these kinds of solving processes, especially within an area of individual interest, so as to build up knowledge in a specialized, individual way, is a really good idea.

Most learners are thought to juggle many competing interests. Which of an learner's needs should be prioritized over another is probably situational and difficult to answer, but we argue that all else being equal, intelligent agents imbued with curiosity should choose to expose themselves to curiosity-inducing situations. With the right implementation, artificial curiosity should direct the agent towards fruitful learning opportunities, much as biological curiosity is thought to \citep[p.~1382]{wade2019role}. Assuming that our design of machine curiosity manages to do the same, we want our machine agents to seek curiosity, which starts with a preference for curiosity-inducing situations---that is, voluntary exposure.

\section{Conclusion}

Curiosity is central to biological intelligence, and machine curiosity is an area of emerging activity for machine intelligence researchers in their pursuit of learning agents that can engage in complex, information-rich environments like the natural world. Throughout this work, we have directly connected insight and empirical evidence from the study of human and animal curiosity to advances in machine intelligence. In particular, we have for the first time translated the idea of specific curiosity to the domain of machine intelligence and shown how it can lead a reinforcement learning machine to exhibit key behaviours associated with curiosity. As a first major contribution of this work, we presented a comprehensive, multidisciplinary survey of animal and machine curiosity. We then used that body of evidence to synthesize and define what we consider to be five of the most important properties of specific curiosity: 

\begin{enumerate}
    \item directedness towards inostensible referents;
    \item cessation when satisfied;
    \item voluntary exposure;
    \item transience;
    \item coherent long-term learning. 
\end{enumerate}

As a second main contribution of this work, we constructed a proof-of-concept reinforcement learning agent interleaving the most salient and immediate properties of specific curiosity. We then conducted empirical sweeps and ablations to probe the role that these integrated properties have on the agent's curious behaviour (and how the removal of individual properties substantially impacts this behaviour). Our computational specific curiosity agent was found to exhibit short-term directed behaviour, update its long-term preferences, and adaptively seek out curiosity-inducing situations. One major insight we draw from this work is that the separation of curiosity-inducing situations from curiosity-satisfying situations is critical to understanding curious behaviour. 

We consider this study a landmark synthesis and translation of specific curiosity to the domain of machine learning and reinforcement learning. It is our hope that this exploration of computational specific curiosity will inspire a new frontier of interdisciplinary work by machine intelligence researchers, and that it will further provide new computational mechanisms to model and study the phenomenon of curiosity in the natural world.

\acks{Thank you to all the amazing people who made this work better and clearer, especially Kate Pratt. Both Brian Tanner and Niko Yasui provided valuable conversations on the contents of this paper. The authors also wish to thank their funding providers. NMA was supported by scholarships from the Natural Sciences and Engineering Research Council of Canada (NSERC), the University of Alberta, the Government of Alberta, and the Women Techmakers Scholars Program. Work by PMP was supported by grants or awards from the Canada CIFAR AI Chairs Program, Alberta Innovates, the Alberta Machine Intelligence Institute (Amii), the Government of Alberta, the University of Alberta, NSERC, and the Canada Research Chairs program.}

\vskip 0.2in
\bibliography{main.bib}
\bibliographystyle{apalike}

\end{document}